\pdfoutput=1
\documentclass{article} 
\usepackage{algpseudocode}
\usepackage{paralist}
\usepackage{authblk}
\usepackage{hyperref}
\usepackage[accepted]{icml2019}

\usepackage{amsmath,amsfonts,bm}

\def\eqref#1{equation~\ref{#1}}

\def\1{\bm{1}}

\DeclareMathAlphabet{\mathsfit}{\encodingdefault}{\sfdefault}{m}{sl}
\SetMathAlphabet{\mathsfit}{bold}{\encodingdefault}{\sfdefault}{bx}{n}

\DeclareMathOperator*{\argmax}{arg\,max}

\usepackage[utf8]{inputenc} 
\usepackage[T1]{fontenc}    
\usepackage{url}            
\usepackage{booktabs}       
\usepackage{amsfonts}       
\usepackage{nicefrac}       
\usepackage{microtype}      

\usepackage{amsmath}
\usepackage{times} 
\usepackage{amssymb,longtable,calc}
\usepackage{mathptmx}       
\usepackage{graphicx}
\usepackage{subfigure}
\usepackage[textwidth=2cm,colorinlistoftodos]{todonotes}
\usepackage{xspace}
\usepackage{latexsym}
\usepackage{amsmath,amssymb,graphicx,xspace}
\usepackage{times}   
\usepackage{multirow}
\usepackage{color,comment,rotating,subfigure,url,xspace} 
\usepackage{tikz}
\usetikzlibrary{arrows,shadows,shapes,backgrounds,decorations,snakes,fit}
\usepackage{algorithm}
\usetikzlibrary{shadows}

\usepackage{booktabs}
\usepackage{caption}

\usepackage{titlesec}
\titlespacing\section{0pt}{3pt plus 2pt minus 2pt}{5pt plus 2pt minus 2pt}
\titlespacing\subsection{0pt}{1pt plus 2pt minus 2pt}{3pt plus 2pt minus 2pt}
\titlespacing\subsubsection{0pt}{2pt plus2pt minus 2pt}{2pt plus 2pt minus 2pt}
\titlespacing\paragraph{0pt}{2pt plus 2pt minus 2pt}{2pt plus 2pt minus 2pt}

\icmltitlerunning{EDDI}
\usepackage{xcolor}

\begin{document}

\twocolumn[

\icmltitle{EDDI: Efficient Dynamic Discovery of~\\High-Value Information with Partial VAE}

\begin{icmlauthorlist}
\icmlauthor{Chao Ma}{cam}
\icmlauthor{Sebastian Tschiatschek}{msr}
\icmlauthor{Konstantina Palla}{msr}
\icmlauthor{Jos{\'e} Miguel Hern{\'a}ndez-Lobato}{cam,msr}
\icmlauthor{Sebastian Nowozin}{google}
\icmlauthor{Cheng Zhang}{msr}
\end{icmlauthorlist}

\icmlaffiliation{cam}{Department of Engineering, University of Cambridge, Cambridge, UK}
\icmlaffiliation{msr}{Microsoft Research, Cambridge, UK}
\icmlaffiliation{google}{Google AI, Berlin, Germany (contributed while being with Microsoft Research)}

\icmlcorrespondingauthor{Cheng Zhang}{Cheng.Zhang@microsoft.com}


\icmlkeywords{Machine Learning, ICML}
\vskip 0.3in
]

\newcommand{\fix}{\marginpar{FIX}}
\newcommand{\new}{\marginpar{NEW}}

\vspace{-15pt}
\printAffiliationsAndNotice{} 
\begin{abstract}

Many real-life decision making situations allow further relevant information to be acquired at a specific cost, for example, in assessing the health status of a patient we may decide to take additional measurements such as diagnostic tests or imaging scans before making a final assessment.  
Acquiring more relevant information enables better decision making, but may be costly. 
How can we trade off the desire to make good decisions by acquiring further information with the cost of performing that acquisition?
To this end, we propose a principled framework, named \emph{EDDI} (Efficient Dynamic Discovery of high-value Information), based on the theory of Bayesian experimental design.
In EDDI, we propose a novel \emph{partial variational autoencoder} (Partial VAE) to predict missing data entries problematically given any subset of the observed ones, and combine it with an acquisition function that maximizes expected information gain on a set of target variables.
We show cost reduction at the same decision quality and improved decision quality at the same cost in multiple machine learning benchmarks and two real-world health-care applications. 
\end{abstract}
\section{Introduction}
\label{sec:intro}

Imagine a person walking into a hospital with a broken arm. The first question from health-care personnel would likely be ``How did you break your arm?'' instead of ``Do you have a cold?'', because the answer reveals relevant information for this patient's treatment.
Human experts dynamically acquire information based on the current understanding of the situation.
Automating this human expertise of asking relevant questions is difficult.
In other applications such as online questionnaires for example, most existing online questionnaire systems either present exhaustive questions \citep{lewenberg2017knowing,shim2018joint} or use extremely time-consuming human labeling work to manually build a decision tree to reduce the number of questions \citep{zakim2008underutilization}.
This wastes the valuable time of experts or users (patients).
An automated solution for personalized dynamic acquisition of information has great potential to save much of this time in many real-life applications. 

What are the technical challenges to building an intelligent information acquisition system?
%
\emph{Missing data is a key issue}: taking the questionnaire scenario as an example, at any point in time we only observe a small subset of answers yet have to reason about possible answers for the remaining questions.
We thus need an accurate probabilistic model that can perform inference given a variable subset of observed answers.
\emph{Another key problem is deciding what to ask next}:
this requires assessing the value of each possible question or measurement, the exact computation of which is intractable.
However, compared to current active learning methods we select individual features, not instances; therefore, existing methods are not applicable.
In addition, these traditional methods are often not scalable to the large volume of data available in many practical cases~\citep{settles2012active, lewenberg2017knowing}.  

We propose the EDDI (Efficient Dynamic Discovery of high-value Information) framework as a scalable information acquisition system for any given task.  We assume that information acquisition is always associated with some cost. Given a task, such as estimating the customers' experience or assessing population health status, we dynamically decide which piece of information to acquire next. The framework is very general, and the information can be presented in any form such as answers to questions, or results of lab tests. Our contributions are:
	\begin{compactitem}
	\item We propose a novel efficient information acquisition framework, EDDI (Section \ref{sec:method}). To enable EDDI, we contribute technically: 
      \begin{compactenum}
      \item \textit{A new partial amortized inference method for generative modeling under partially observed data (Section \ref{sec:pVAE}).}
      We extend the variational autoencoder (VAE) \citep{kingma2014auto,Rezende2014StochasticBA}, to account for partial observations. The resulting method, which we call the Partial VAE, is inspired by the set formulation of the data \citep{qi2017pointnet,zaheer2017deep}. The Partial VAE, as a probabilistic framework in the presence of missing data, is highly scalable, and serves as the base for the EDDI framework. Note that Partial VAE itself is widely applicable and can be used on its own as a non-linear probabilistic framework for missing-data imputation. 
      \item \textit{An information theoretic acquisition function with a novel efficient approximation, yielding a novel variable-wise active learning method (Section \ref{sec:reward}).} \\
      Based on the partial VAE, we actively select the unobserved variable which contributes most to the task, such as costumer surveys and health assessments, evaluated using the mutual information. This acquisition function does not have an analytical solution, and we derive a novel efficient approximation. 
      \end{compactenum}
    \item We demonstrate the performance of EDDI in various settings, and apply it in real-life health-care scenarios (Section \ref{sec:exp}). 
    \begin{compactenum}
    \item We first show the superior performance of the Partial VAE framework on an image inpainting task (Section \ref{sec:inpaint}). 
    \item We then use 6 different datasets from the Machine Learning repository of University of Irvine (UCI) \citep{Dua:2017} to demonstrate the behavior of EDDI, comparing with multiple baseline methods (Section \ref{sec:UCI}).
    \item Finally, we evaluate EDDI on two real-life health-care applications: risk assessment in intensive care (Section \ref{sec:MIMIC}) and public health assessment using a national survey (Section \ref{sec:NHANES}), where traditional methods without amortized inference do not scale. EDDI shows clear improvements in both applications. 
    \end{compactenum}
	\end{compactitem}

\section{Related Work}
\label{sec:related}

EDDI requires a method that handles partially observed data to enable dynamic variable wise active learning. We thus review related methods for handling partial observation and performing active learning. 

\subsection{Partial Observation}

Missing data entries are common in many real-life applications, which has created a long history of research on the topic of dealing with missing data \citep{rubin1976inference, dempster1977maximum}. We describe existing methods below with the focus of probabilistic methods:

\paragraph{Traditional methods without amortization.~} Prediction based methods have shown advantages for missing value imputation \citep{scheffer2002dealing}. Efficient matrix factorization based methods have been recently applied \citep{keshavan2010matrix, jain2010guaranteed, salakhutdinov2008bayesian}, where the observations are assumed to be able to decompose as the multiplication of low dimensional matrices. In particular, many probabilistic frameworks with various distribution assumptions \citep{salakhutdinov2008bayesian, blei2003latent} have been used for missing value imputation \citep{yu2016temporal, hamesse2018simultaneous} and also recommender systems where unlabeled items are predicted \citep{matchbox09,wang2011collaborative,gopalan2014content}.  

The probabilistic matrix factorization method has been used in the active variable selection framework called the dimensionality reduction active learning model (DRAL),\citep{lewenberg2017knowing}.  These traditional methods suffer from limited model capacity since they are typically linear. Additionally, they do not scale to large volumes of data and thus are usually not applicable in real-world applications. For example, \citet{lewenberg2017knowing}  tested the performance of their method with a single user due to the heavy computational cost of traditional inference methods for probabilistic matrix factorization.

\paragraph{Utilizing Amortized Inference.~} 
Amortized inference  \citep{kingma2014auto,Rezende2014StochasticBA,zhang2017advances} has significantly improved the scalability of deep generative latent variable models. In the case of partially observed data, amortized inference is particularly of interest due to the speed requirement in many real-life applications. 
\citet{wu2018conditional} use amortized inference during training, where the training dataset is assumed to be fully observed. 
During test time, the traditional non-scalable inference is used to infer missing data entries from the partially observed dataset using the pre-trained model. This method is restrictive since it is not scalable in the test time and the fully observed training set assumption does not hold for many applications.

\citet{nazabal2018handling} use zero imputation (ZI) for amortized inference for both training and test sets with missing data entries. ZI is a generic and straightforward method that first fills the missing data with zeros, and then feeds the imputed data as input for the inference network. The drawback of ZI is that it introduces bias when the data are not missing completely at random which leads to a poorly fit model. We also observe artifacts when 
using it for the image inpainting task. 
Independent of our work, \citet{garnelo2018conditional} explore interpreting variational autoencoder (amortized inference) as stochastic processes, which also handles partial observation per se.   

\subsection{Active Learning}

\paragraph{Traditional Active Learning.~~}
Active learning, also referred to as experimental design, aims to obtain optimal performance with fewer selected data (or experiments) \citep{lindley1956measure, mackay1992information, settles2012active}. Traditional active learning aims to select the \emph{next data point} to label. Many information theoretical approaches have shown promising results in various settings with different acquisition functions \citep{ mackay1992information,mccallumzy1998employing, houlsby2011bayesian}. These methods commonly assume that the data are fully observed, and the acquisition decision is instance wise.  Little work has dealt with missing values within instances.  \citet{zheng2002active} deal with missing data values by imputing with traditional non-probabilistic methods \citep{little1987statistical} first. It is still an instance-wise active learning framework. 

Different from traditional active learning, our proposed framework performs \emph{variable-wise active learning} for \emph{each} instance. In this setting, information theoretical acquisition functions need a new design as well as  non-trivial approximations. The most closely related work is the aforementioned DRAL \citep{lewenberg2017knowing}, which deals with variable-wise active learning for each instance. 

\paragraph{Active Feartue Acquisition (AFA).~}
Active sequential feature selection is of great need, especially in cost-sensitive applications. 
Thus, many methods have also been applied and resulted in the class of methodologies called Active Feature Acquisition (AFA) \citep{melville2004active,saar2009active, thahir2012efficient, huang2018active}. For instance, \citet{melville2004active,saar2009active}  have designed objectives to select any feature from any instance to minimize the cost to achieve high accuracy. The proposed framework is very general. However, the problem setting of AFA methods is different from our active variable selection problem.AFA aims to select training set optimally that would result in the best classifier (model), while assume that the test data are fully observed. On the contrary, our framework aims to identify and acquire high value information sequentially for each teat instance.

\section{Method}
\label{sec:method}

In this section, we first formalize the active variable selection problem. Then, we present the Partial VAE to model and perform inference on partial observations.  Finally, we complete the EDDI framework by presenting our new acquisition function and estimation method.

\subsection{Problem formulation}
In this work, we focus on the following active variable selection problem. Let $\mathbf{x}=[x_1, \ldots, x_{|I|}]$ be a set of random variables with probability density $p(\mathbf{x})$. Furthermore, let a subset of the variables $\mathbf{x}_O$, $O \subset I$, be observed while the variables $\mathbf{x}_U$, $U = I \setminus O$, are unobserved. Assume that we can query the value of variables $x_i$ for $i \in U$. The goal of active variable selection is to query a sequence of variables in $U$ in order to predict a quantity of interest $f(\mathbf{x})$, as accurately as possible while simultaneously performing as few queries as possible, where $f(\cdot)$ can be any (random) function. 
This problem, in the simplified myopic setting, can be formalized as that of proposing the next variable $x_{i^*}$ to be queried by maximizing a reward function $R$ at each step:
\begin{align} \label{eq:problem}
  i^* = \argmax_{i \in U} R(i \mid \mathbf{x}_O),
\end{align} 
where $R(i \mid \mathbf{x}_O)$  quantifies the merit of our prediction of $f(\cdot)$ given $\mathbf{x}_0$ and $x_i$. Furthermore, the reward can quantify other properties important to the problem, e.g.\ the cost of acquiring $x_i$.

\subsection{Partial Amortization of Inference Queries}
\label{sec:pVAE}

We first introduce how to establish a generative probabilistic model of random variables $\mathbf{x}$, that is capable of handling unobserved (missing) variables $\mathbf{x}_U$ with variable size.  Our approach to this, named the Partial VAE, is based on the variational autoencoder (VAE), which enables amortized inference to scale to large volumes of data.

\paragraph{VAE and amortized inference.~~} A VAE defines a generative model in which the data $\mathbf{x}$ is generated from latent variables $\mathbf{z}$,  $p(\mathbf{x,z}; \mathbf{\theta}) = \prod_{i} p_{\mathbf{\theta}}(\mathbf{x}_i|\mathbf{z}) p(\mathbf{z})$. The data generation, $p_{\mathbf{\theta}}(\mathbf{x}|\mathbf{z})$, is realized by a deep neural network. To approximate the posterior of the latent variable $p_\theta(\mathbf{z} | \mathbf{x})$, VAEs use \emph{amortized} variational inference. Specifically, it uses an encoder, which is another neural network with the data $\mathbf{x}$ as input to produce a variational approximation of the posterior $q(\mathbf{z}|\mathbf{x}; \mathbf{\phi})$. As traditional variational inference, VAE is trained by maximizing an evidence lower bound (ELBO),
which is equivalent to minimizing the KL divergence between $q(\mathbf{z}|\mathbf{x}; \mathbf{\phi})$ and $p_\theta(\mathbf{z}|\mathbf{x})$.

VAEs are not directly applicable when data points have arbitrary subset of data entries missing. Consider the situation that the variables are divided into \emph{observed} variables $\mathbf{x}_O$ and \emph{unobserved} variables $\mathbf{x}_U$.
In this setting, we would like to efficiently and accurately infer $p(\mathbf{z}|\mathbf{x}_O)$ and $p(\mathbf{x}_U|\mathbf{x}_O)$.  
One main challenge is that there are many possible  partitions  $\{U,O\}$, where the size of observed variables might vary.
Therefore, classic approaches to training a VAE with the variational bound and amortized inference networks are not applicable. We propose to extend amortized inference to handle partial observations.

\paragraph{Partial VAE.~}
In a VAE, $p(\mathbf{x}|\mathbf{z})$ is factorized, i.e.
\begin{equation}
\vspace{-2pt}
p(\mathbf{x}|\mathbf{z}) = \prod_i p_i(\mathbf{x}_i|\mathbf{z}).
\end{equation}
This implies that given $\mathbf{z}$, the observed variables $\mathbf{x}_O$ are conditionally independent of $\mathbf{x}_U$.
Therefore,
\begin{equation}
p(\mathbf{x}_U|\mathbf{x}_O,\mathbf{z}) = p(\mathbf{x}_U|\mathbf{z}),
\end{equation}
and inferences about $\mathbf{x}_U$ can be reduced to inference about $\mathbf{z}$.
Hence, the key object of interest in this setting is $p(\mathbf{z}|\mathbf{x}_O)$, i.e., the posterior over the latent variables $\mathbf{z}$ given the observed variables  $\mathbf{x}_O$. Once we obtain $\mathbf{z}$, computing $\mathbf{x}_U$ is straightforward. To approximate $p(\mathbf{z}|\mathbf{x}_O)$, we introduce a variational inference network $q(\mathbf{z}|\mathbf{x}_O)$ and define a partial variational lower bound,
\begin{align} \label{eq:pVAE}
& \log p(\mathbf{x}_O) \geq \log p(\mathbf{x}_O) - D_{\textrm{KL}}(q(\mathbf{z}|\mathbf{x}_O) \| p(\mathbf{z}|\mathbf{x}_O)) \\ \nonumber
& = \mathbb{E}_{\mathbf{z} \sim q(\mathbf{z}|\mathbf{x}_O)}[  \log p(\mathbf{x}_O|\mathbf{z}) + \log p(\mathbf{z}) - \log q(\mathbf{z}|\mathbf{x}_O)]\\ \nonumber
&\equiv \mathcal{L}_{partial}.
\end{align}
This bound, $\mathcal{L}_{partial}$, depends only on the observed variables $\mathbf{x}_O$, whose dimensionality may vary among different data points.
We thus call the the inference net, $q(\mathbf{z}|\mathbf{x}_O)$, the \emph{partial inference net}.
Specifying $q(\mathbf{z}|\mathbf{x}_O)$ requires distributions for any partition $\{O, U\}$ of $I$.

\paragraph{Amortized Inference with partial observations.~ }

\begin{figure}[t]
\centering
\subfigure[Partial VAE PN setting]{
\resizebox{0.45 \textwidth}{!}{
\pgfdeclarelayer{background}
\pgfdeclarelayer{foreground}
\pgfsetlayers{background,main,foreground}

\definecolor{babyblue}{rgb}{0.54, 0.81, 0.94}
\definecolor{bisque}{rgb}{1.0, 0.89, 0.77}
\definecolor{bittersweet}{rgb}{1.0, 0.44, 0.37}

\begin{tikzpicture}

\tikzstyle{surround} = [thick,draw=black,rounded corners=1mm]
\tikzstyle{scalarnode} = [circle, draw, fill=white!11,  
    text width=1.2em, text badly centered, inner sep=2.5pt]
\tikzstyle{scalarnodenoline} = [  fill=white!11, 
    text width=1.2em, text badly centered, inner sep=2.5pt]
\tikzstyle{arrowline} = [draw,color=black, -latex]
\tikzstyle{dashedarrowcurve} = [draw,color=black, dashed, out=100,in=250, -latex]
\tikzstyle{dashedarrowline} = [draw,color=black, dashed,  -latex]

\node [] at (0.2,0.5) (ip) {\small $(|O|, M+1)$};

\node (rect) at (0,0) [draw,thick, fill=babyblue, minimum width=1.5cm, minimum height=0.5cm] (e1){$e_1$};
\node (rect) at (0,-0.5) [draw,thick,fill=babyblue,minimum width=1.5cm, minimum height=0.5cm] {$e_2$};
\node (rect) at (0,-1) [draw,thick,fill=white,minimum width=1.5cm, minimum height=0.5cm] {$...$};
\node (rect) at (0,-1.5) [draw,thick,fill=babyblue,minimum width=1.5cm, minimum height=0.5cm] {$e_{i}$};
\node (rect) at (0,-2) [draw,thick,fill=babyblue,minimum width=1.5cm, minimum height=0.5cm] (e4) {$e_{|O|}$};

\node (rect) at (1,0) [draw,thick, fill=bisque, minimum width=0.75cm, minimum height=0.5cm] (x1){$x_1$};
\node (rect) at (1,-0.5) [draw,thick,fill=bisque,minimum width=0.75cm, minimum height=0.5cm] (x2) {$x_2$};
\node (rect) at (1,-1) [draw,thick,fill=white,minimum width=0.75cm, minimum height=0.5cm] {$...$};
\node (rect) at (1,-1.5) [draw,thick,fill=bisque,minimum width=0.75cm, minimum height=0.5cm] (x3){$x_{i}$};
\node (rect) at (1,-2) [draw,thick,fill=bisque,minimum width=0.75cm, minimum height=0.5cm] (x4){$x_{{|O|}}$};

\node (rect)  at (2.5,0) [draw, rounded rectangle,thick,inner sep=0pt, fill=blue!50, minimum width=1.5cm, minimum height=0.35cm] (h1) { \small $h$};

\node (rect) at (2.5,-0.5) [draw, rounded rectangle,thick,inner sep=0pt,fill=blue!50,minimum width=1.5cm, minimum height=0.35cm] (h2) { \small$h$};
\node [scalarnodenoline] at (2.25,-1) (s1) {shared};
\node (rect) at (2.5,-1.5) [draw, rounded rectangle,thick,inner sep=0pt,fill=blue!50,minimum width=1.5cm, minimum height=0.35cm] (h3){ \small$h$};
\node (rect) at (2.5,-2) [draw, rounded rectangle,thick,inner sep=0pt,fill=blue!50,minimum width=1.5cm, minimum height=0.35cm] (h4){  \small$h$};

\path [arrowline]  (x1) to (h1);
\path [arrowline]  (x2) to (h2);
\path [arrowline]  (x3) to (h3);
\path [arrowline]  (x4) to (h4);

\node [circle] at (4,-1) [draw,thick, fill=black!50, minimum width=0.4cm, minimum height=0.4cm] (g) {g};

\path [arrowline]  (3.2,0) to (g);
\path [arrowline]  (3.2,-0.5) to (g);
\path [arrowline]  (3.2,-1.5) to (g);
\path [arrowline]  (3.2,-2) to (g);

\node [] at (5,-0.5) (ip) {\small $(1,K)$};

\node (rect) at ( 5,-1) [draw,thick, fill=bittersweet, minimum width=0.8cm, minimum height=0.4cm] (c1) {$ c$};

\path [arrowline]  (g) to (c1);

\node (rect) at (2.4,-0.9) [draw,thick,dashed,minimum width=6.5cm, minimum height=3.3cm] {};


\node (rect) at ( 6,-1) [draw,thick, rotate=90, fill=black!50, minimum width=0.4cm, minimum height=0.4cm] (conc) {ENCODER};
\path [arrowline]  (c1) to (conc);

\end{tikzpicture}}
\label{fig:tikz_PN}
}
\subfigure[Parial VAE PNP setting]{
\resizebox{0.45 \textwidth}{!}{
\pgfdeclarelayer{background}
\pgfdeclarelayer{foreground}
\pgfsetlayers{background,main,foreground}

\definecolor{babyblue}{rgb}{0.54, 0.81, 0.94}
\definecolor{bisque}{rgb}{1.0, 0.89, 0.77}
\definecolor{bittersweet}{rgb}{1.0, 0.44, 0.37}

\begin{tikzpicture}

\tikzstyle{surround} = [thick,draw=black,rounded corners=1mm]
\tikzstyle{scalarnode} = [circle, draw, fill=white!11,  
    text width=1.2em, text badly centered, inner sep=2.5pt]
\tikzstyle{scalarnodenoline} = [  fill=white!11, 
    text width=1.2em, text badly centered, inner sep=2.5pt]
\tikzstyle{arrowline} = [draw,color=black, -latex]
\tikzstyle{dashedarrowcurve} = [draw,color=black, dashed, out=100,in=250, -latex]
\tikzstyle{dashedarrowline} = [draw,color=black, dashed,  -latex]

\node [] at (0.2+8,0.5) (ip2) {\small $(|O|, M)$};

\node (rect) at (0+8,0) [draw,thick, fill=babyblue, minimum width=1.5cm, minimum height=0.5cm] (e1){$e_1$};

\node [] at (0+8,-1) {$...$};

\node (rect) at (0+8,-1.5) [draw,thick,fill=babyblue,minimum width=1.5cm, minimum height=0.5cm] (e4) {$e_{|O|}$};

\node (rect) at (0.7+8,-0.7) [draw,thick, fill=bisque, minimum width=0.75cm, minimum height=0.5cm] (x1){$x_1$};

\node (rect) at (0.7+8,-2.2) [draw,thick,fill=bisque,minimum width=0.75cm, minimum height=0.5cm] (x4){$x_{|O|}$};

\node [circle] at (1.2+8+0.2,-0.3) [draw,thick, fill=black!50, minimum width=0.3cm, minimum height=0.3cm,inner sep=0pt] (m1){$\times$};

\node [circle] at (1.2+8+0.2,-1.8) [draw,thick,fill=black!50,minimum width=0.3cm, minimum height=0.3cm, inner sep=0pt] (m4){$\times$};

\node (rect)  at (2.5+8+0.2,-0.5) [draw, rounded rectangle,thick,inner sep=0pt, fill=blue!50, minimum width=1.5cm, minimum height=0.35cm] (h1) { \small $h$};

\node [scalarnodenoline] at (2.25+8+0.2,-1) (s1) {shared};

\node (rect) at (2.5+8+0.2,-1.6) [draw, rounded rectangle,thick,inner sep=0pt,fill=blue!50,minimum width=1.5cm, minimum height=0.35cm] (h4){  \small$h$};

\node (circle) at (4+8,-1) [draw,thick, fill=black!50, minimum width=0.4cm, minimum height=0.4cm] (g2) {g};
\path [arrowline]  (x1) to (m1);
\path [arrowline]  (x4) to (m4);
\path [arrowline]  (e1) to (m1);
\path [arrowline]  (e4) to (m4);
\path [arrowline]  (m1) to (h1);
\path [arrowline]  (m4) to (h4);
\path [arrowline]  (h1) to (g2);
\path [arrowline]  (h4) to (g2);

\node [] at (5+8,-0.5) (ip) {\small $(1,K)$};

\node (rect) at ( 5+8,-1) [draw,thick, fill=bittersweet, minimum width=0.8cm, minimum height=0.4cm] (c2) {$ c$};

\path [arrowline]  (g2) to (c2);

\node (rect) at ( 6+8,-1) [draw,thick, rotate=90, fill=black!50, minimum width=0.4cm, minimum height=0.4cm] (conc2) {ENCODER};

\path [arrowline]  (c2) to (conc2);

\node (rect) at (2.4+8,-0.9) [draw,thick,dashed,minimum width=6.5cm, minimum height=3.3cm] {};

\end{tikzpicture}}
\label{fig:tikz_PNP}
}
\vspace{-5pt}
\caption{Illustration of Partial VAE encoder architecture.} \vspace{-15pt} 
\label{fig:tikz_PN_all}
\end{figure}
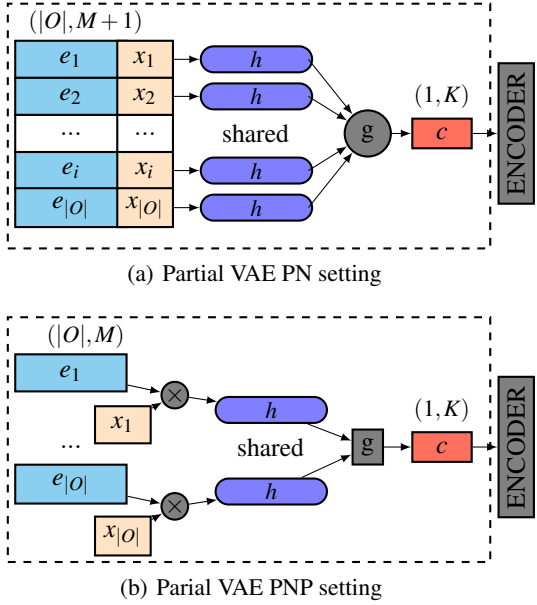

Inference under partial observations requires the inference net of VAE to be capable to handle arbitrary set of observed data, and sharing parameters across these different sized sets of observations for amortization.

Inspired by the \emph{Point Net (PN)} approach  for point cloud classification \citep{qi2017pointnet, zaheer2017deep}, we specify the approximate distribution $q(\mathbf{z}|\mathbf{x}_O)$ by a \emph{permutation invariant set function encoding}, given by:
\begin{equation}
\mathbf{c}(\mathbf{x}_O) := g(h(\mathbf{s}_1),h(\mathbf{s}_2),...,h(\mathbf{s}_{|O|})),
\end{equation}
where  $\mathbf{s}_d$ carries the information of the input of the $d$-th observed variable, and $|O|$ is the number of observed variables. In particular, $\mathbf{s}_d$ contains the information about the  identity of the input $\mathbf{e}_d$ and the corresponding input value $x_d$.  There are many ways to define the identity variable, $\mathbf{e}_d$. Naively, it could be the coordinates of observed pixels for images, and one-hot embedding of the number of questions in a questionnaire. With different problem settings, it can be beneficial to learn $\mathbf{e}$ as an embedding of the identity of the variable, either with or without an naive encoding as input.  In this work, we treat $\mathbf{e}$ as an unknown embedding, to be optimized during training. 

There are also different ways to construct $\mathbf{s}_d$. A common choice is concatenation, $\mathbf{s}_d = [\mathbf{e}_d, x_d]$, which is often used in computer vision applications \citep{qi2017pointnet}. Such architecture is illustrated in Figure \ref{fig:tikz_PN}. We refer to this setting as the \emph{Pointnet (PN)} specification of Partial VAE. However, the construction of $\mathbf{s}_d$ can be more flexible.  We propose to construct $\mathbf{s}_d = \mathbf{e}_d*x_d$ using element-wise multiplication as an alternative, shown in Figure \ref{fig:tikz_PNP}. We show that this formulation generalizes naive Zero Imputation (ZI) VAE \citep{nazabal2018handling} (cf. Appendix \ref{sec:ZIasPN}). We refer to the multiplication setting as the \emph{Pointnet Plus (PNP)} specification of Partial VAE. 
 
We can then use a neural network $h(\cdot)$ to map the input $\mathbf{s}_d$ to $\mathbb{R}^K$, where and $K$ is the latent space size.  The key to the PNP/PN structure is the permutation invariant aggregation operation $g(\cdot)$, such as max-pooling or summation.  In this way,  the mapping $\mathbf{c}(\mathbf{x}_O)$ is invariant to the permutations of elements of $\mathbf{x}_O$, and $\mathbf{x}_O$ can have arbitrary length. Finally, the fixed-size code $\mathbf{c}(\mathbf{x}_O)$ is fed into an ordinary neural network, that transforms the code into the statistics of a multivariate Gaussian distribution to approximate $p(\mathbf{z}|\mathbf{x}_O)$. The procedure is illustrated in Figure~\ref{fig:tikz_PN_all}. As discussed before, given  $p(\mathbf{z}|\mathbf{x}_O)$, we can estimate $p(\mathbf{x}_U|\mathbf{z})$.

\subsection{Efficient Dynamic Discovery of High-value Information}
\label{sec:reward}

We now cast the active variable selection problem (\ref{eq:problem}) as an adaptive Bayesian experimental design problem, utilizing $p(\mathbf{x}_U | \mathbf{x}_O)$ inferred by the Partial VAE. Algorithm \ref{alg:EDDI} summarizes the EDDI framework.

\paragraph{Information Reward.~}
We designed a variable selection acquisition function in an information theoretic way following Bayesian experimental design \citep{lindley1956measure, bernardo1979expected}.
\citet{lindley1956measure} provides a generic formulation of Bayesian experimental design by maximizing the expected Shannon information. \citet{bernardo1979expected} generalizes it by considering the decision task context. 

For a given task, we are interested in  statistics of some variables $\mathbf{x}_{\phi}$, where $\mathbf{x}_{\phi} \subset \mathbf{x}_U$. Given a new instance (user), assume that we have observed $\mathbf{x}_O$ so far for this instance, and we need to select the next variable $x_i$ (an element of $\mathbf{x}_{U \setminus \phi}$) to observe.
 Following \citet{bernardo1979expected}, we select $x_i$ by maximizing:
\begin{equation} \label{eq:IR}
R(i,\mathbf{x}_O) = \mathbb{E}_{\mathbf{x}_i \sim p(\mathbf{x}_i|\mathbf{x}_O)}
	D_{\textrm{KL}}\left[p(\mathbf{x}_\phi | \mathbf{x}_i,\mathbf{x}_O) \,\|\, p(\mathbf{x}_\phi | \mathbf{x}_O)
\right].
\end{equation}
In our paper, we mainly consider the case that a subset of interesting observations represents the statistics of interest $\mathbf{x}_{\phi}$.  Sampling $\mathbf{x}_i \sim p(\mathbf{x}_i|\mathbf{x}_o)$ is approximated by $\mathbf{x}_i \sim \hat{p}(\mathbf{x}_i|\mathbf{x}_o)$, where $\hat{p}(\mathbf{x}_i|\mathbf{x}_o)$ can be obtained by using the Partial VAE. It is  implemented by first sampling $\mathbf{z} \sim q(\mathbf{z}|\mathbf{x}_o)$, and then $\mathbf{x}_i \sim p(\mathbf{x}_i|\mathbf{z})$. The same applies for  $p(\mathbf{x}_i, \mathbf{x}_\phi|\mathbf{x}_o)$ which appears in Equation (\ref{eq:CHAIN}).

{
\begin{algorithm}[t]
\begin{algorithmic}[1]
 \Require {Training dataset $\mathbf{X}$, which is partially observed; Test dataset $\mathbf{X}^*$ with no observations collected yet; Indices $\phi$ of target variables. }
 
 \State \textbf{Train Partial VAE} by optimizing partial variational bound with  $\mathbf{X}$ (cf.\ Section \ref{sec:pVAE})
 
 \State \textbf{Actively acquire feature value} $x_i$ to estimate $\mathbf{x}^*_\phi$ for each test point (cf.\ Section \ref{sec:reward}) 
\end{algorithmic}
\setlength\labelsep{8em}
\begin{minipage}{\textwidth}
\begin{algorithmic} 
 \For{each test instance}
   \State $\mathbf{x}_O \leftarrow \emptyset$ (no variable value has been observed for \\
   any test point)
	\Repeat
      \State Choose variable $x_i$ from $U \setminus \phi$ to maximize the \\
      information reward (Equation~(\ref{eq:CHAIN2}))
      \State $\mathbf{x}_O \leftarrow x_i \cup \mathbf{x}_O$
    \Until Stopping criterion reached (e.g. the time budget) 
 \EndFor
\end{algorithmic}
\end{minipage}
\caption{EDDI: Algorithm Overview}
\label{alg:EDDI}
\end{algorithm}
 }

\paragraph{Efficient approximation of the Information reward.~}
The Partial VAE allows us to sample $\mathbf{x}_i \sim p(\mathbf{x}_i|\mathbf{x}_o)$. However, the KL term in Equation (\ref{eq:IR}), 
\begin{align}
& D_{KL}\left[ p(\mathbf{x}_{\phi}|\mathbf{x}_i,\mathbf{x}_o)||p(\mathbf{x}_{\phi}|\mathbf{x}_o) \right] \\ \nonumber
& = - \int_{\mathbf{x}_{\phi}} p(\mathbf{x}_{\phi}|\mathbf{x}_i,\mathbf{x}_o) \log \frac{p(\mathbf{x}_{\phi}|\mathbf{x}_o) }{p(\mathbf{x}_{\phi}|\mathbf{x}_i,\mathbf{x}_o)}, 
\end{align}
is intractable since both $p(\mathbf{x}_{\phi}|\mathbf{x}_i,\mathbf{x}_o)$ and $p(\mathbf{x}_{\phi}|\mathbf{x}_o)$ are intractable. For high dimensional $\mathbf{x}_{\phi}$, entropy estimation could be difficult.  The entropy term $\int_{\mathbf{x}_{\phi}} p(\mathbf{x}_{\phi}|\mathbf{x}_i,\mathbf{x}_o) \log p(\mathbf{x}_{\phi}|\mathbf{x}_i,\mathbf{x}_o)$ depends on $i$ hence cannot be ignored.  In the following, we show how to approximate this expression. 

Note that analytic solutions of KL-divergences are available under specific variational distribution families of $q(\mathbf{z}|\mathbf{x}_{O})$ (such as the Gaussian distribution commonly used in VAEs). Instead of calculating the information  reward in $\mathbf{x}$ space, we have shown that one can compute in the $\mathbf{z}$ space (cf. Appendix \ref{sec:app_chain}):
\begin{align} \label{eq:CHAIN}
R(i,\mathbf{x}_o)  = & \mathbb{E} _{ \mathbf{x}_i \sim p(\mathbf{x}_i|\mathbf{x}_o)} D_{KL}\left[ p(\mathbf{z}|\mathbf{x}_i,\mathbf{x}_o)||p( \mathbf{z}|\mathbf{x}_o)\right] - \\ \nonumber
&\mathbb{E} _{\mathbf{x}_{\phi},\mathbf{x}_i \sim p(\mathbf{x}_{\phi}, \mathbf{x}_i|\mathbf{x}_o)}  D_{KL}\left[ p(\mathbf{z}|\mathbf{x}_{\phi},\mathbf{x}_i,\mathbf{x}_o)||p( \mathbf{z}|\mathbf{x}_{\phi}, \mathbf{x}_o) \right].
\end{align}
Note that Equation (\ref{eq:CHAIN}) is exact. Additionally, we use the partial VAE approximation 
  $
  p(\mathbf{z}|\mathbf{x}_{\phi},\mathbf{x}_i,\mathbf{x}_o) \approx q(\mathbf{z}|\mathbf{x}_{\phi},\mathbf{x}_i,\mathbf{x}_o)
$, $p(\mathbf{z}|\mathbf{x}_o) \approx q(\mathbf{z}_i|\mathbf{x}_o)$ and $p(\mathbf{z}|\mathbf{x}_i,\mathbf{x}_o) \approx q(\mathbf{z}_i|\mathbf{x}_i, \mathbf{x}_o)$. 
This leads to the final approximation of the information reward:
\begin{align} \label{eq:CHAIN2}
\hat{R}(i,\mathbf{x}_o) =& \mathbb{E} _{ \mathbf{x}_i \sim \hat{p}(\mathbf{x}_i|\mathbf{x}_o)} D_{KL}\left[ q(\mathbf{z}|\mathbf{x}_i,\mathbf{x}_o)||q( \mathbf{z}|\mathbf{x}_o)\right] -\\ \nonumber
&\mathbb{E} _{\mathbf{x}_{\phi},\mathbf{x}_i \sim \hat{p}(\mathbf{x}_{\phi}, \mathbf{x}_i|\mathbf{x}_o)}  D_{KL}\left[ q(\mathbf{z}|\mathbf{x}_{\phi},\mathbf{x}_i,\mathbf{x}_o)||q( \mathbf{z}|\mathbf{x}_{\phi}, \mathbf{x}_o) \right].
\end{align}
With this approximation,  the divergence between $q(\mathbf{z}|\mathbf{x}_i,\mathbf{x}_o)$ and $q( \mathbf{z}|\mathbf{x}_o)$ can often be computed analytically in the Partial VAE setting, for example, under Gaussian parameterization.
The only Monte Carlo sampling required is the one set of samples $\mathbf{x}_{\phi},\mathbf{x}_i \sim p(\mathbf{x}_{\phi}, \mathbf{x}_i|\mathbf{x}_o)$ that can be shared across different KL terms in Equation (\ref{eq:CHAIN2}). Our EDDI framework is opensource at \url{https://github.com/Microsoft/EDDI}.

\section{Experiments}
\label{sec:exp}
\begin{figure*}[t]
\subfigure[Input]{\centering
    \includegraphics[width=0.08\textwidth]{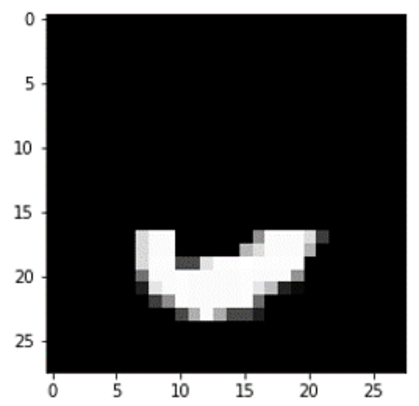}
    \label{fig:trunc}} 
    \hspace{-5pt}
\subfigure[ZI]{\centering
   \includegraphics[width=0.21\textwidth]{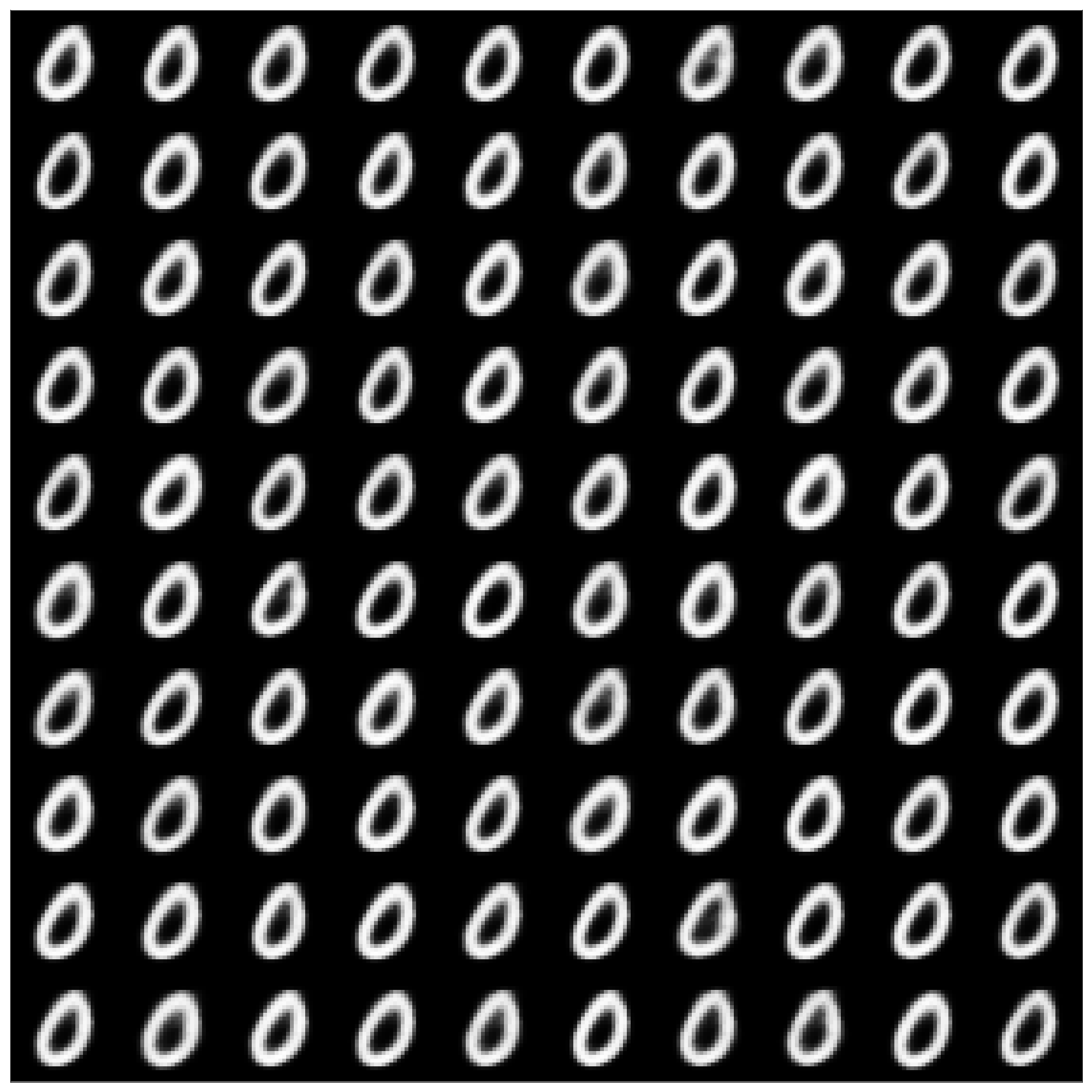}
   \label{fig:IP_ZI}}
      \hspace{-5pt}
    \subfigure[ZI-m]{\centering
   \includegraphics[width=0.21\textwidth]{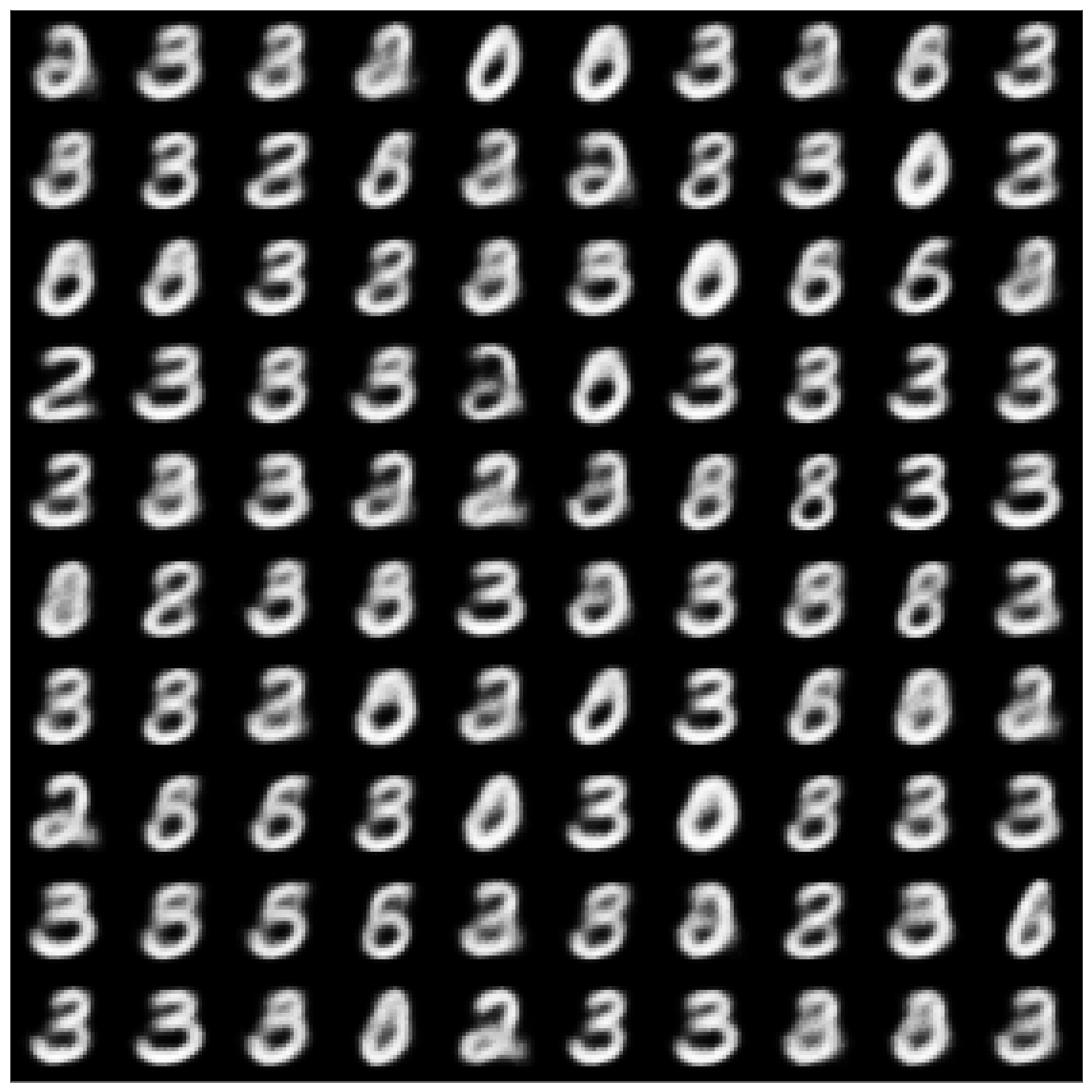}
   \label{fig:IP_ZIM}}
      \hspace{-5pt}
      \subfigure[PN]{
   \includegraphics[width=0.21\textwidth]{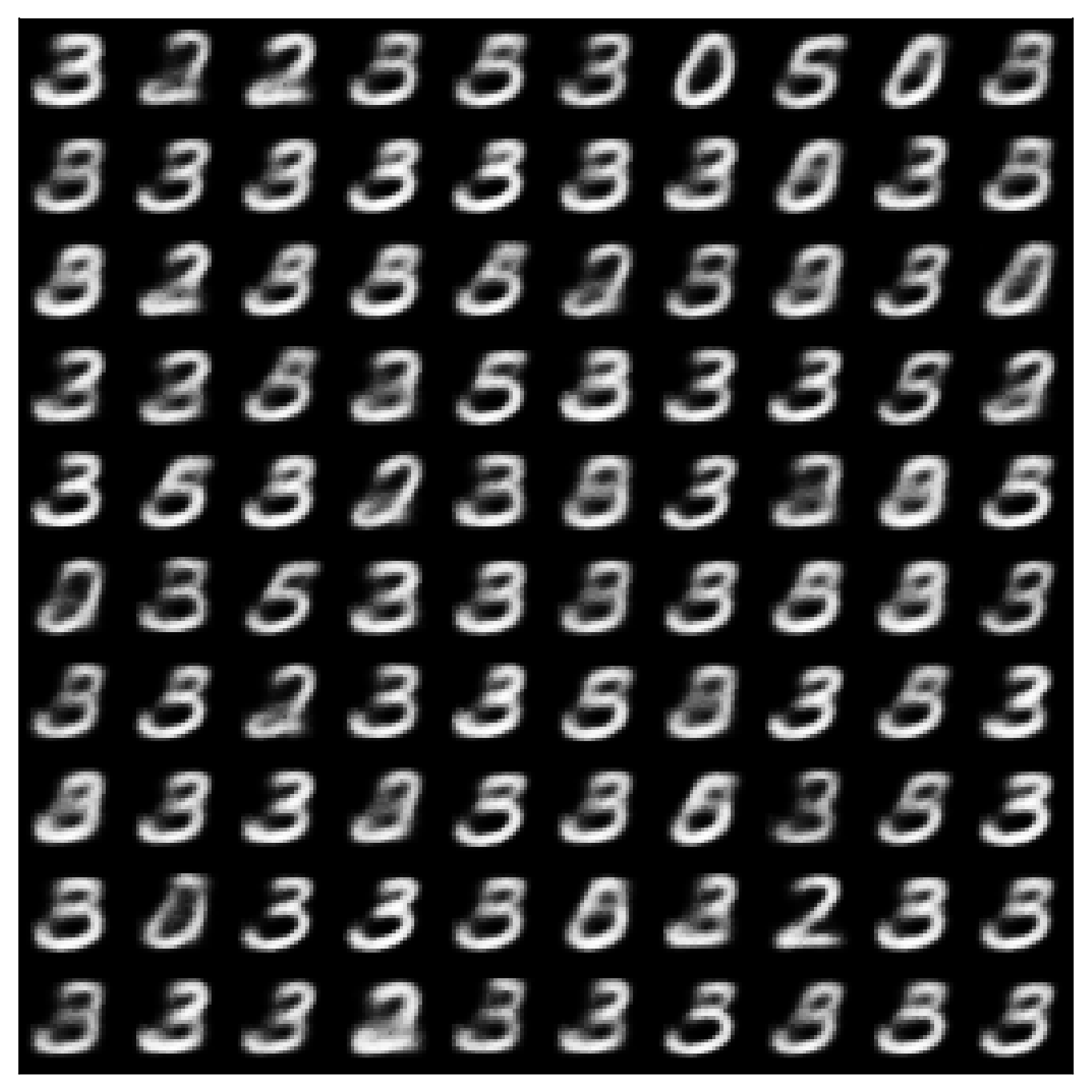}
   \label{fig:IP_PN}}
      \hspace{-5pt}
   \subfigure[PNP]{\centering
   \includegraphics[width=0.21\textwidth]{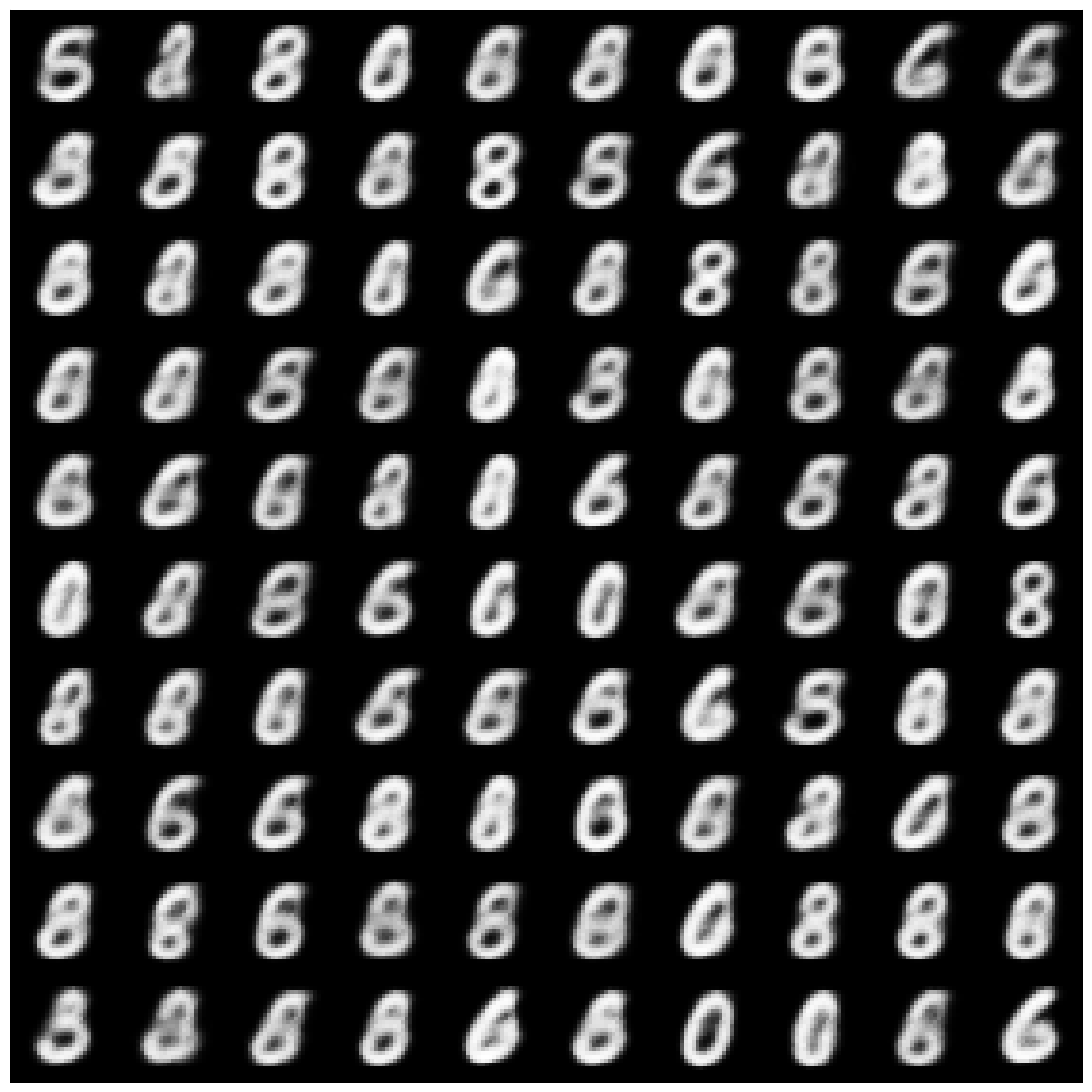}
   \label{fig:IP_PNP}}
   \vspace{-5pt}
   \caption{Image inpainting example with MNIST dataset using Partial VAE with four settings. 
   }
   \label{fig:mnist_completion}
\end{figure*}

\begin{figure*}[t]
\centering
\subfigure[Boston Hosing]{\centering
   \includegraphics[width=0.32\textwidth]{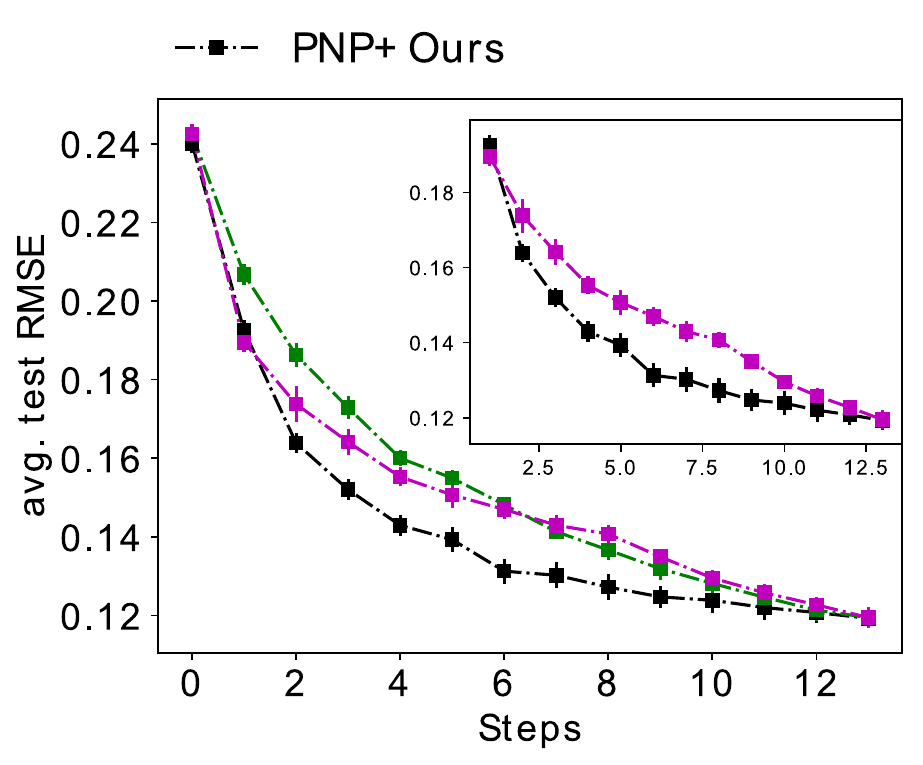}}
\subfigure[Energy]{\centering
   \includegraphics[width=0.32\textwidth]{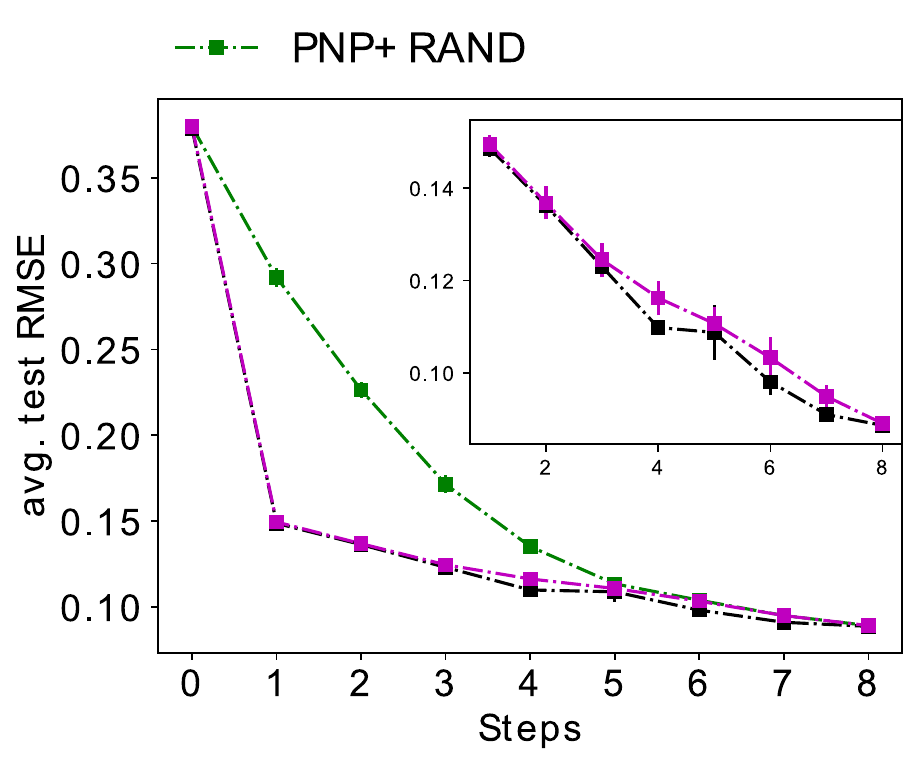}}
  \subfigure[Wine]{
   \includegraphics[width=0.32\textwidth]{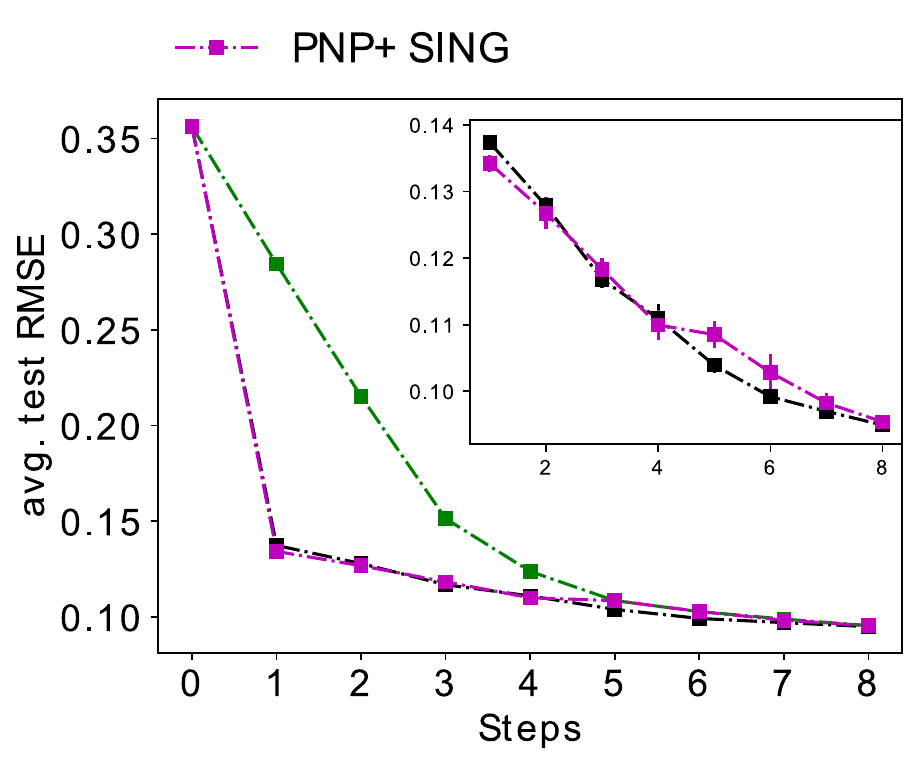}}
   \vspace{-5pt}
   \caption{Information curves of active variable selection, demonstrated on three UCI datasets (based on PNP parameterization of Partial VAE). This displays negative test RMSE (y axis, the lower the better) during the course of active selection (x-axis). Error bars represent standard errors over 10 runs.
   \vspace{-8pt}}. 
   \label{fig:uci_pnp}
\end{figure*}
Here we evaluate the proposed EDDI framework. We first assess the Partial VAE component of EDDI alone on an image inpainting task both qualitatively and quantitatively (Section \ref{sec:inpaint}). We compare our proposed two PN-based Partial VAE with the zero-imputing (ZI) VAE \citep{nazabal2018handling}. 
Additionally, we modify the ZI VAE to use the mask matrix indicating which variables are currently observed as input.  We name this method ZI-m VAE.
We then demonstrate the performance of the entire EDDI framework on datasets from the UCI repository (Section \ref{sec:UCI} ), as well as in two real-life application scenarios: Risk assessment in intensive care (Section \ref{sec:MIMIC}) and public health assessment with national health survey (Section \ref{sec:NHANES}).  We compare the performance of EDDI, using four different Partial VAE settings, with three baseline information acquisition strategies. The first baseline is the \emph{random active feature selection strategy (denoted as RAND)} which randomly picks the next variable to observe. RAND reflects the strategy used in many real-world applications, such as online surveys. The second baseline method is the \emph{single best strategy (denoted as SING)} which finds a single fixed global optimal order of selecting variables. This order is then applied to all data points.  SING uses the objective function as in Equation (\ref{eq:CHAIN2}) to find the optimal ordering by averaging over all the test data.

\subsection{Image inpainting with Partial VAE}
\label{sec:inpaint}
We evaluate the performance of Partial VAE with the image inpainting task, which is to fill in the removed pixels. We perform the evaluation in two different settings: in the first setting, pixels are randomly removed, and in the second setting, a continuous patch of pixels are removed.
\begin{table}[h]
\setlength{\tabcolsep}{4pt}
\centering
\caption{Comparing models trained on partially observed MNIST. VAE-full is an ideal reference. }
\resizebox{\linewidth}{!}{
\begin{tabular}{p{1.in}p{0.6in}p{0.6in}p{0.6in}p{0.6in}p{0.6in}} \toprule
\textbf{Method} &VAE-full & ZI & ZI-m & PN & PNP \\ \midrule
Train ELBO & -95.05 & $ \textbf{-113.64}$&-117.29&-121.43&$ \textbf{-113.64}$ \\ 
Test ELBO (Rnd.) & -101.46 &-116.01& -118.61 & -122.20 & \textbf{-114.01} \\
Test ELBO (Reg.) & -101.46& -130.61 & -123.87 & -116.53 & \textbf{-113.19} \\ \bottomrule
\end{tabular}
}
\label{tab:mnist_llh}
\end{table}
\paragraph{Inpainting Random Missing Pixels.~} We use MNIST dataset \citep{lecun1998mnist} and remove pixels randomly for this task. 
The same settings are used for all methods (see Appendix \ref{sec:app_inpaint} for details). During training, we remove a random portion (uniformly sampled between 0\% and 70\%) of pixels. We then impute missing pixels on a partially observed test set (constructed by removing 70\% of the pixels uniform randomly). The performance of pixel imputation is evaluated by test ELBOs on missing pixels. The first two rows in Table \ref{tab:mnist_llh} show training and test ELBOs for all algorithms using this partially observed dataset. Additionally, we show ordinary VAE (VAE-full) trained on the fully observed dataset as an ideal reference. Among all Partial VAE methods, the PNP approach performs best. 

\begin{figure*}[t]
\centering
\begin{minipage}[t]{0.35 \textwidth}
\vspace{0pt}
\centering
\includegraphics[width=4.5cm]{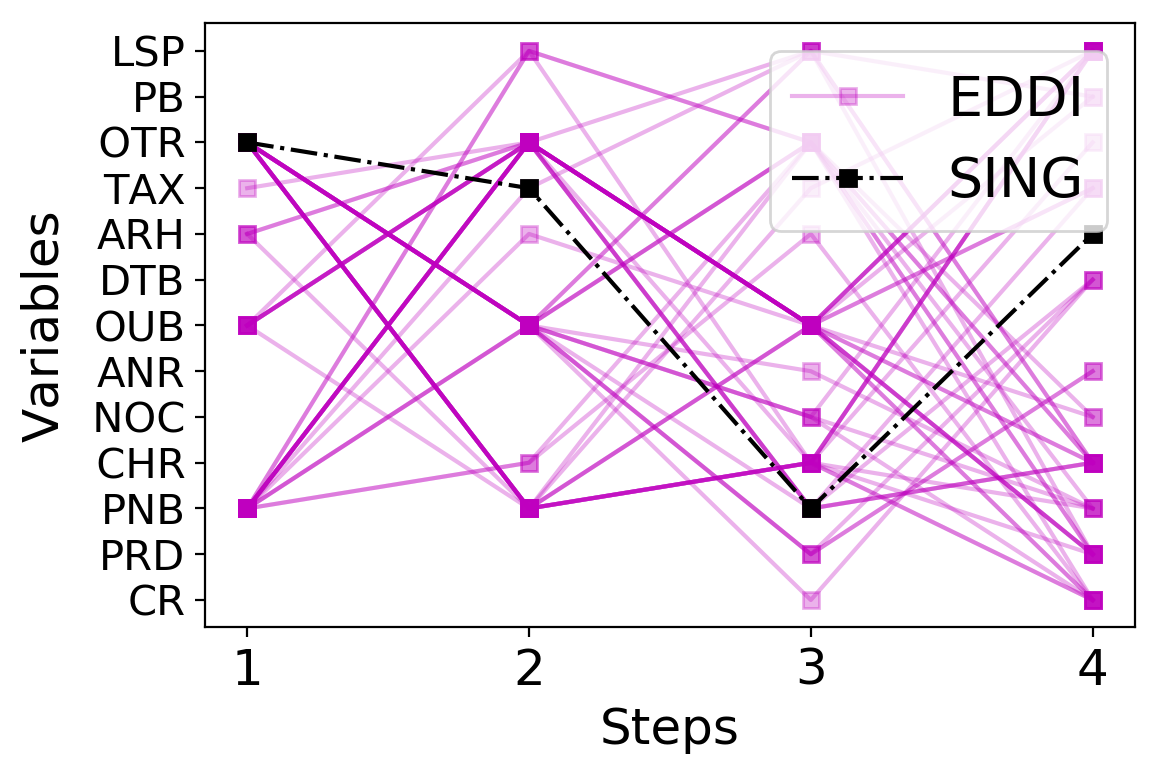}
\vspace{-4pt}
\captionsetup{font = small}
\caption{First four decision steps on Boston Housing test data. EDDI is ``personalized'' comparing SING. Full names of the variables are listed in the Appendix \ref{sec:app_UCI}.}
\label{fig:bh_dec}
\end{minipage}
\hspace{5pt}
\begin{minipage}[t]{0.35 \textwidth}
\vspace{0pt}
\includegraphics[width=4.5cm]{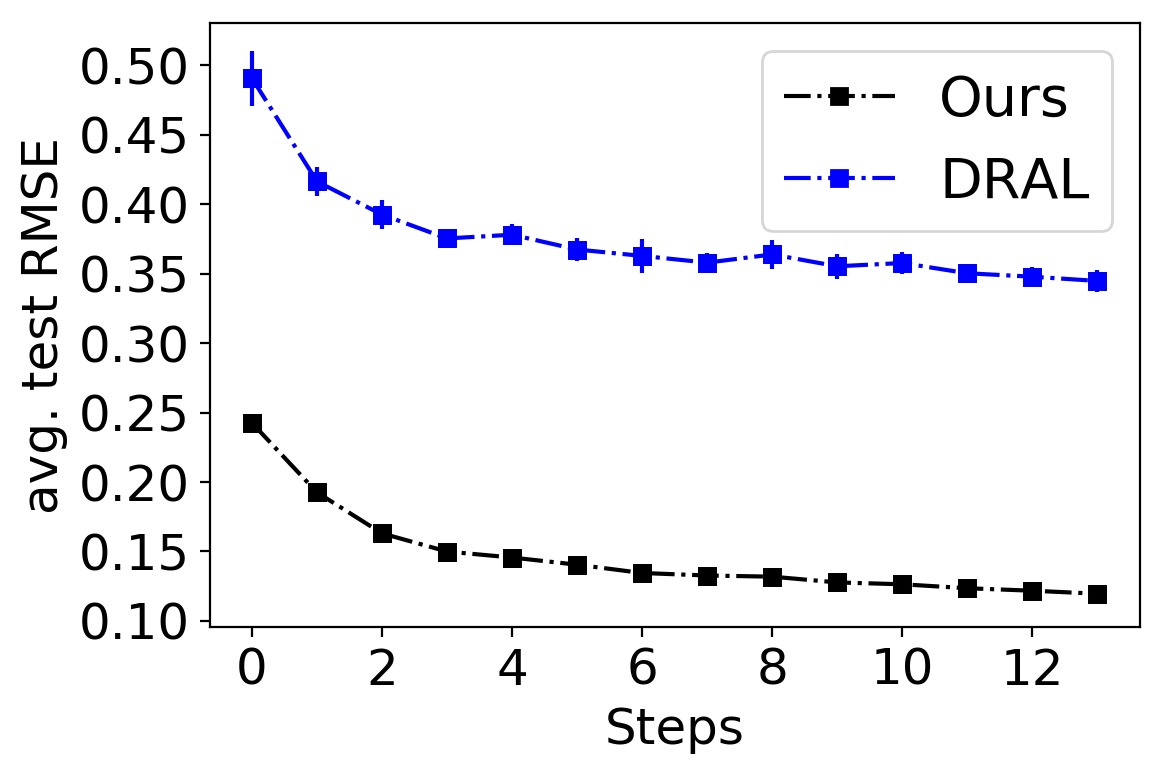}
\vspace{-6pt}
\captionsetup{font = small}
\caption{Comparison of DRAL \citep{lewenberg2017knowing} and EDDI on Boston Housing dataset. EDDI out performs DRAL significantly regarding test RMSE in every step.  }
\label{fig:matchbox_vs_eddi}
\end{minipage}
\hspace{5pt}
\begin{minipage}[t]{0.22 \textwidth}
\vspace{15pt}
\centering
\scalebox{0.8}{
\begin{tabular}{c c}
\toprule
Method & Time 
\\ \midrule
DRAL & 2747.16 \\
EDDI & \textbf{2.64} \\
\end{tabular}
}
\captionsetup{font = small}
\captionof{table}{Test CPU time (in seconds) per test point for active variable selection using EDDI and DRAL. EDDI is $10^3$ times more efficient than DRAL \citep{lewenberg2017knowing} computationally.   }
\label{tab:Time}
\end{minipage}
\end{figure*}

\paragraph{Inpainting Regions.~}  We then consider inpainting large contiguous regions of images. It aims to evaluate the capability of the Partial VAEs to produce all possible outcomes with better uncertainty estimates.
With the same trained model as before, we remove the region of the upper $60\%$ pixels of the image in the test set.
We then evaluate the average likelihoods of the models.  The last row of Table \ref{tab:mnist_llh} shows the results of the test ELBO in this case. PNP based Partial VAE performs better than other settings.  Note that given only the lower half of a digit,  the number cannot be identified uniquely.
ZI (Figure \ref{fig:IP_ZI})  fails to cover the different possible modes due to its limitation in posterior inference. ZI-m (Figure \ref{fig:IP_ZIM}) is capable of producing multiple modes. However, some of the generated samples are not consistent with the given part (i.e., some digits of 2 are generated). Our proposed PN (Figure \ref{fig:IP_PN}) and PNP (Figure \ref{fig:IP_PNP}) 
are capable of recovering different modes, and are consistent with observations.

\subsection{EDDI on UCI datasets}
\label{sec:UCI}

\begin{table}[h]
\centering
\caption{Average ranking of AUIC over 6 UCI datasets.  }
\resizebox{\linewidth}{!}{
\begin{tabular}{p{1.in}p{0.8in}p{0.8in}p{0.8in}p{0.8in}} \toprule
\textbf{Method} &ZI & ZI-m & PNP & PN \\ \midrule
EDDI & 5.72 (0.03
) & 5.54 (0.02
) & \textbf{5.08 (0.02
)} & 5.25 (0.02)\\ 
Random & 8.03 (0.03
) & 8.10 (0.03
) & 7.77 (0.03
)& 7.79 (0.03
)\\ 
Single best & 8.68 (0.03
)& 5.50 (0.02
) & 5.20 (0.02
)& 5.28 (0.02
)\\ 
\bottomrule
\end{tabular}
}
\label{tab:uci_auic}
\end{table}

Given the effectiveness of our proposed Partial VAE, we now demonstrate the performance of our proposed EDDI framework in comparison with random selection (RAND) and single optimal ordering  (SING). 
We first apply EDDI on 6 different UCI datasets (cf. Appendix \ref{sec:app_UCI}) \citep{Dua:2017}. We report the results of EDDI with all these four different specifications of Partial VAE (ZI, ZI-m, PN, PNP). 

All Partial VAE  are first trained on partially observed UCI datasets where a random portion of variables is removed. We actively select variable for each test point starting with empty observation $\mathbf{x}_o = \emptyset$. 
In all UCI datasets, we randomly sample $10\%$ of the data as the test set. All experiments are repeated for ten times.

Taking PNP based setting as an example, Figure \ref{fig:uci_pnp}  shows the test RMSE on $\mathbf{x}_{\phi}$ for each variable selection step with three different datasets, where $\mathbf{x}_{\phi}$ is defined by the UCI task. We call this curve the \emph{information curve (IC)}.
We see that EDDI can obtain information efficiently. It archives the same test RMSE with less than half of the variables.  Single optimal ordering also improves upon random ordering. However, it is less efficient compared with EDDI, since EDDI perform active learning for each data instance which is ``personalized''. Figure \ref{fig:bh_dec} shows an example of the decision processes using EDDI and SING. The first step of EDDI overlaps largely with SING. From the second step, EDDI makes ``personalized'' decisions.

We also present the average performance among all datasets with different settings. The area under the information curve (AUIC),
can then be used to compare the performance across models and strategies. Smaller AUIC value indicates better performance. 
However, due to different datasets have different scales of RMSEs and different numbers of variables (indicated by steps), it is not fair to average the AUIC across datasets to compare overall performances. We thus define average  \emph{ranking} of AUIC that compares 12 methods (indexed by $i$) averaging these datasets as: $r_{i} = \frac{1}{\sum_j N_j}\sum_{j=1}^{6} \sum_{k=1}^{N_j} r_{ijk}, \ \ i = 1,..,12$. These 12 methods are cross combinations of four Partial VAE models with three variable selection strategies. $r_{i}$ is the final ranking of $i$th combination, $r_{ijk}$ is the ranking of the $i$th combination (based on AUIC value) regarding the $k$th test data point in the $j$th UCI dataset, and $N_j$ is the size of the $j$th UCI dataset.  This gives us $ 6 \sum_j N_j$ different rankings. Finally, we compute the mean and standard error statistics based on these rankings. Table \ref{tab:uci_auic} summarize the average ranking results. We provide additional statistical significance test (Wilcoxcon signed-rank test for paired data) in Appendix \ref{sec:wilcox}. Based on these experimental results, we see that EDDI outperforms other variable selection order in all different Partial VAE settings. Among different partial VAE settings, PNP/PN-based settings perform better than ZI-based settings.

\paragraph{Comparison with non-amortized method.~}Additionally, we compare EDDI to DRAL 
\citep{lewenberg2017knowing} which is the state-of-the-art method for the same problem setting. 
As discussed in Section \ref{sec:related}, DRAL is linear and requires high computational cost. 
 The DRAL paper only tested their method on a single test data point due to its limitation on computational efficiency. We compare DRAL with EDDI on Boston Housing dataset with ten randomly selected test points here. Results are shown in Figure \ref{fig:matchbox_vs_eddi}, where EDDI significantly outperforms DARL thanks to more flexible Partial VAE model. Additionally, EDDI is 1000 times more efficient than DARL as shown in Table \ref{tab:Time}.

\subsection{Risk assessment with MIMIC-III}
\label{sec:MIMIC}
We now apply EDDI to risk assessment tasks using the Medical Information Mart for Intensive Care (MIMIC III) database \citep{johnson2016mimic}. 
MIMIC III is the most extensive publicly available clinical database, containing real-world records from over 40,000 critical care patients with 60,000 ICU stays. The risk assessment task is to predict the final mortality. We preprocess the data for this task following \citet{harutyunyan2017multitask}   \footnote{\href{https://github.com/yerevann/mimic3-benchmarks}{https://github.com/yerevann/mimic3-benchmarks}}. This results in a dataset of 21139 patients.  
We treat the final mortality of a patient as a Bernoulli variable. 
For our task, we focus on variable selection, which corresponds to medical instrument selection.  We thus further process the time series variables into static variables based on temporal averaging.   

\begin{figure}[t]
\centering
\centering
   \includegraphics[width=4.8cm]{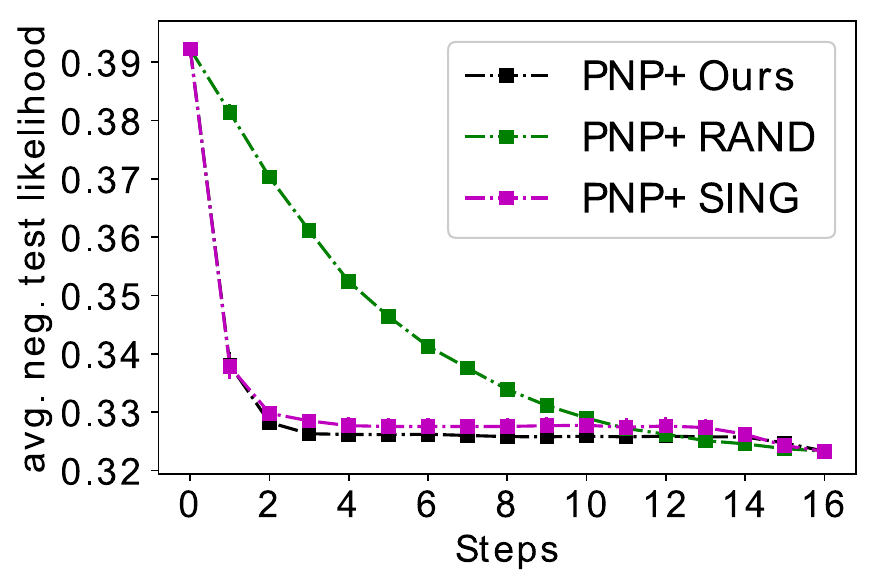}
   \captionsetup{font = small}
   \caption{Information curves (based on Bernoulli negative log likelihood) of active variable selection on risk assessment task on MIMIC III  with PNP setting. \vspace{-8pt}  }
   \label{fig:mimic}
\end{figure}

Figure \ref{fig:mimic} shows the information curve (based on Bernoulli likelihoods) of different strategies, using PNP based Partial VAE as an example (more results in Appendix \ref{sec:app_MIMIC}). Table \ref{tab:mimic} shows the average ranking of AUIC with different settings. In this application, EDDI significantly outperforms other variable selection strategies in all different settings of Partial VAE, and PNP based setting performs best.

\begin{figure}[t]
\centering
   \includegraphics[width=5cm]{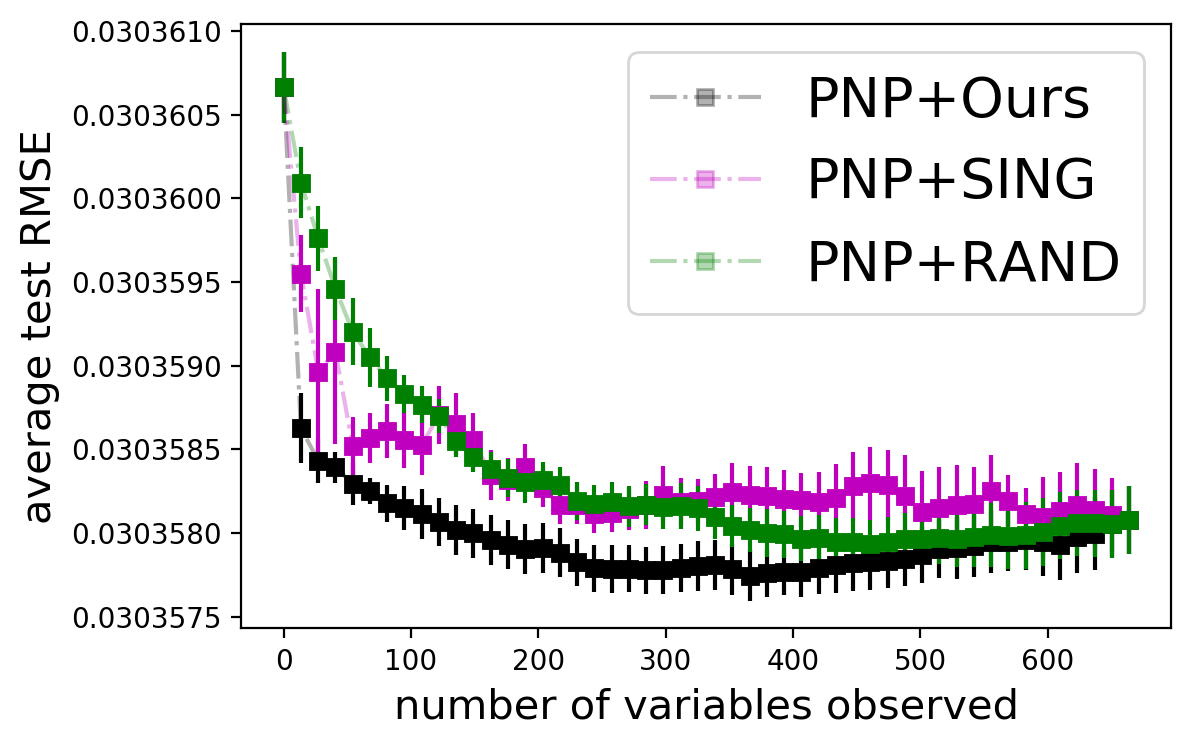}
   \vspace{-5pt}
   \captionsetup{font = small}
   \caption{Information curves of active (grouped) variable selection on risk assessment task on NHANES with PNP setting.  \vspace{-5pt}}
   \label{fig:nhs}
\end{figure}

\begin{table}[t]
\centering
\scalebox{0.8}{
\begin{tabular}{c c c c} \toprule
\textbf{Method} & EDDI  &  Random & Single best \\ \midrule
ZI & 8.83 (0.01) & 7.97 (0.02) & 9.83 (0.01)   \\
ZI-m &4.91 (0.01) & 7.00 (0.01) & 5.91 (0.01)  \\
PN  & 4.96 (0.01) & 6.62 (0.01) & 5.96 (0.01) \\
PNP & \textbf{4.39 (0.01)} & 6.18 (0.01) & 5.39 (0.01) \\ 
\bottomrule
\end{tabular}
}
\captionsetup{font = small}
\captionof{table}{ Average ranking on AUIC of MIMIC III}
\label{tab:mimic}
\end{table}

\begin{table}[t]
\centering
\scalebox{0.8}{
\begin{tabular}{c c c c} \toprule
\textbf{Method} & EDDI &  Random & Single best  \\ \midrule
ZI & 6.00 (0.10) & 8.45 (0.09) & 6.51 (0.09)   \\
ZI-m & 8.06 (0.09) & 8.67 (0.09) & 8.68 (0.07)  \\
PN & 5.28 (0.10) & 5.57 (0.10) & 5.46 (0.09) 
\\
PNP& \textbf{4.80} (0.10) & 5.30 (0.10) & 5.17 (0.10)\\
\bottomrule
\end{tabular}
}
\captionsetup{font = small}
\captionof{table}{Average ranking on AUIC of NHANES 
}
\label{tab:nhs}
\end{table}

\subsection{Public Health Assessment with NHANES}
\label{sec:NHANES}

Finally, we apply our methods to public health assessment using NHANES  2015-2016  data \cite{cdc2005national}.  NHANES is a program with adaptable components of measurements, to assess the health and nutritional status of adults and children in the United States. Every year, thousands individuals of all ages are interviewed and examined in their homes.
This 2015-2016 NHANES data contains three major sections, the questionnaire interview, examinations and lab tests for 9971 subjects in the publicly available version of this cycle.  In our setting, we consider the whole set of lab test results (139 variables) as the target variable of interest $\mathbf{x}_{\phi}$ since they are expensive and reflects the subject's health status, and we active select the questions from the extensive questionnaire (665 variables).

The questionnaire of NHANES is divided into 73 different groups. In practice,  questions in the same group are often examined together. Therefore, we perform active variable selection on the group level: at each step, the algorithm selects one  group to observe. This is more challenging than the experiments in previous sections since it requires the generative model to simulate a group of unobserved data in Equation (\ref{eq:CHAIN2}) at the same time.  When evaluating test RMSE on the target variable of interest, we treat variables in each group equally. For a fair comparison, the calculation of the area under the information curve (AUIC) is weighted by the size of the group chosen by the algorithms. Specifically, AUIC is calculated after spline interpolation. The information curve plots in Figure \ref{fig:nhs}, together with Table \ref{tab:nhs} of AUIC statistics show that our EDDI outperforms other baselines. In addition, this experiment shows that EDDI is capable of performing active selection on a large pool of grouped variables to estimate a high dimensional target.

\section{Conclusion}
\label{sec:discussion}
In this paper, we present EDDI, a novel and efficient framework for dynamic active variable selection for each instance. Within the EDDI framework, we propose Partial VAE which performs amortized inference to handle missing data. Partial VAE alone can be used as a non-linear computational efficient probabilistic imputation method. Based on it, we design a variable wise acquisition function for EDDI and derive corresponding approximation method. 
EDDI has demonstrated its effectiveness on active variable selection tasks across multiple real-world applications. In the future, we would extend the EDDI framework to handle more complicated scenarios, such as data missing not at random, time-series, and the cold-start situation. 
\bibliography{ref}
\bibliographystyle{icml2019.bst}
\clearpage
\appendix
\newpage

\onecolumn
\section{Additional Derivations}

\subsection{Information reward approximation}
\label{sec:app_chain}
In our paper, given the VAE model $p(\mathbf{x} | z)$ and a partial inference network $q(\mathbf{z}|\mathbf{x}_o)$, the experimental design problem is formulated as maximization of the information reward:
$$ R(i,\mathbf{x}_o) = \mathbb{E}_{\mathbf{x}_i \sim p(\mathbf{x}_i|\mathbf{x}_o)} [D_{KL}( {\color{blue}p(\mathbf{x}_{\phi}|\mathbf{x}_i,\mathbf{x}_o)}||{\color{red}p(\mathbf{x}_{\phi}|\mathbf{x}_o) }) ] $$
Where $p(\mathbf{x}_{\phi}|\mathbf{x}_i,\mathbf{x}_o) = \int_{\mathbf{z}} p(\mathbf{x}_{\phi}|\mathbf{z}) q(\mathbf{z}|\mathbf{x}_i,\mathbf{x}_o)$, $p(\mathbf{x}_{\phi}|\mathbf{x}_o) = \int_{\mathbf{z}} p(\mathbf{x}_{\phi}|\mathbf{z}) q(\mathbf{z}|\mathbf{x}_o)$ and $q(\mathbf{z}|\mathbf{x}_o)$ are approximate condition distributions given by partial VAE models. Now we consider the problem of directly approximating $R(i,\mathbf{x}_o)$. 

Applying the chain rule of KL-divergence, we have:
\begin{align*}
&D_{KL}( {\color{blue}p(\mathbf{x}_{\phi}|\mathbf{x}_i,\mathbf{x}_o)}||{\color{red}p(\mathbf{x}_{\phi}|\mathbf{x}_o) })  \\
&= D_{KL}( p(\mathbf{x}_{\phi},\mathbf{z}|\mathbf{x}_i,\mathbf{x}_o)||p(\mathbf{x}_{\phi}, \mathbf{z}|\mathbf{x}_o)) \\
& - \mathbb{E}_{\mathbf{x}_{\phi} \sim p(\mathbf{x}_\phi|\mathbf{x}_i,\mathbf{x}_o)} \left[ D_{KL}( p(\mathbf{z}|\mathbf{x}_{\phi},\mathbf{x}_i,\mathbf{x}_o)||p( \mathbf{z}|\mathbf{x}_{\phi}, \mathbf{x}_o)) \right],
\end{align*}

Using again the KL-divergence chain rule on $D_{KL}( p(\mathbf{x}_{\phi},\mathbf{z}|\mathbf{x}_i,\mathbf{x}_o)||p(\mathbf{x}_{\phi}, \mathbf{z}|\mathbf{x}_o))$, we have:
\begin{align*}
&D_{KL}( p(\mathbf{x}_{\phi},\mathbf{z}|\mathbf{x}_i,\mathbf{x}_o)||p(\mathbf{x}_{\phi}, \mathbf{z}|\mathbf{x}_o)) \\
&= D_{KL}( p(\mathbf{z}|\mathbf{x}_i,\mathbf{x}_o)||p( \mathbf{z}|\mathbf{x}_o))  + D_{KL}( p(\mathbf{x}_{\phi}|\mathbf{z},\mathbf{x}_i,\mathbf{x}_o)||p(\mathbf{x}_{\phi}| \mathbf{z}, \mathbf{x}_o)) \\ 
&= D_{KL}( p(\mathbf{z}|\mathbf{x}_i,\mathbf{x}_o)||p( \mathbf{z}|\mathbf{x}_o))  + D_{KL}( p(\mathbf{x}_{\phi}|\mathbf{z})||p(\mathbf{x}_{\phi}| \mathbf{z})) \\ 
& = D_{KL}( p(\mathbf{z}|\mathbf{x}_i,\mathbf{x}_o)||p( \mathbf{z}|\mathbf{x}_o)).
\end{align*}
The KL-divergence term in the reward formula is now rewritten as follows,
\begin{align*}
&D_{KL}( {\color{blue}p(\mathbf{x}_{\phi}|\mathbf{x}_i,\mathbf{x}_o)}||{\color{red}p(\mathbf{x}_{\phi}|\mathbf{x}_o) })  \\
& =  D_{KL}( p(\mathbf{z}|\mathbf{x}_i,\mathbf{x}_o)||p( \mathbf{z}|\mathbf{x}_o)) \\
& - \mathbb{E}_{\mathbf{x}_{\phi} \sim p(\mathbf{x}_\phi|\mathbf{x}_i,\mathbf{x}_o)} \left[ D_{KL}( p(\mathbf{z}|\mathbf{x}_{\phi},\mathbf{x}_i,\mathbf{x}_o)||p( \mathbf{z}|\mathbf{x}_{\phi}, \mathbf{x}_o)) \right].
\end{align*}

One can then plug in the partial VAE inference approximation: 

\begin{align*}
&p(\mathbf{z}|\mathbf{x}_{\phi},\mathbf{x}_i,\mathbf{x}_o)  \approx q(\mathbf{z}|\mathbf{x}_{\phi},\mathbf{x}_i,\mathbf{x}_o), \\ &p(\mathbf{z}|\mathbf{x}_i,\mathbf{x}_o)  \approx q(\mathbf{z}|\mathbf{x}_i,\mathbf{x}_o), \ \ p(\mathbf{z}|\mathbf{x}_o)  \approx q(\mathbf{z}|\mathbf{x}_o)
\end{align*}
Finally, the information reward is now approximated as:
\begin{align*}
&R(i,\mathbf{x}_o) \\
&\approx \mathbb{E}_{\mathbf{x}_i \sim p(\mathbf{x}_i|\mathbf{x}_o)} \left[ D_{KL}( q(\mathbf{z}|\mathbf{x}_i,\mathbf{x}_o)||q( \mathbf{z}|\mathbf{x}_o)) \right] \\
&-  \mathbb{E}_{\mathbf{x}_i \sim p(\mathbf{x}_i|\mathbf{x}_o)}\mathbb{E}_{\mathbf{x}_{\phi} \sim p(\mathbf{x}_\phi|\mathbf{x}_i,\mathbf{x}_o)} \left[ D_{KL}( q(\mathbf{z}|\mathbf{x}_{\phi},\mathbf{x}_i,\mathbf{x}_o)||q( \mathbf{z}|\mathbf{x}_{\phi}, \mathbf{x}_o)) \right] \\
& = \mathbb{E}_{{\color{green}\mathbf{x}_i \sim p(\mathbf{x}_i|\mathbf{x}_o)}} \left[ D_{KL}( q(\mathbf{z}|\mathbf{x}_i,\mathbf{x}_o)||q( \mathbf{z}|\mathbf{x}_o)) \right] \\
&-  \mathbb{E}_{{\color{green}\mathbf{x}_{\phi},\mathbf{x}_i \sim p(\mathbf{x}_{\phi}, \mathbf{x}_i|\mathbf{x}_o)}} \left[ D_{KL}( q(\mathbf{z}|\mathbf{x}_{\phi},\mathbf{x}_i,\mathbf{x}_o)||q( \mathbf{z}|\mathbf{x}_{\phi}, \mathbf{x}_o)) \right] = \hat{R}(i,\mathbf{x}_o).
\end{align*}
This new objective tries to maximize the shift of belief on latent variables $\mathbf{z}$ by introducing $\mathbf{x}_i$, while penalizing the information that cannot be absorbed by $\mathbf{x}_\phi$ (by the penalty term $D_{KL}( q(\mathbf{z}|\mathbf{x}_{\phi},\mathbf{x}_i,\mathbf{x}_o)||q( \mathbf{z}|\mathbf{x}_{\phi}, \mathbf{x}_o))$). Moreover, it is more computationally efficient since one set of samples ${\color{green}\mathbf{x}_{\phi},\mathbf{x}_i \sim p(\mathbf{x}_{\phi}, \mathbf{x}_i|\mathbf{x}_o)}$ can be shared across different terms, and the KL-divergence between common parameterizations of encoder (such as Gaussians and normalizing flows) can be computed exactly without the need for approximate integrals. Note also that under approximation $$p(\mathbf{z}|\mathbf{x}_{\phi},\mathbf{x}_i,\mathbf{x}_o)  \approx q(\mathbf{z}|\mathbf{x}_{\phi},\mathbf{x}_i,\mathbf{x}_o), \ \ p(\mathbf{z}|\mathbf{x}_i,\mathbf{x}_o)  \approx q(\mathbf{z}|\mathbf{x}_i,\mathbf{x}_o), \ \ p(\mathbf{z}|\mathbf{x}_o)  \approx q(\mathbf{z}|\mathbf{x}_o)$$, sampling $\mathbf{x}_i \sim p(\mathbf{x}_i|\mathbf{x}_o)$ is approximated by $\mathbf{x}_i \sim \hat{p}(\mathbf{x}_i|\mathbf{x}_o)$, where $\hat{p}(\mathbf{x}_i|\mathbf{x}_o)$ is defined by the following process in Partial VAE. It is  implemented by first sampling $\mathbf{z} \sim q(\mathbf{z}|\mathbf{x}_o)$, and then $\mathbf{x}_i \sim p(\mathbf{x}_i|\mathbf{z})$. The same applies for  $p(\mathbf{x}_i, \mathbf{x}_\phi|\mathbf{z})$.
\section{Additional Experimental Results}

\subsection{Image inpainting}
\label{sec:app_inpaint}

\subsubsection{Preprocessing and model details}
\label{sec:mnist_pre}
For our MNIST experiment, we randomly draw 10\% of the whole data to be our test set. Partial VAE models (ZI, ZI-m, PNP and PNs) share the same size of architecture with 20 dimensional diagonal Gaussian latent variables: the generator (decoder) is a 20-200-500-500 fully connected neural network with ReLU activations (where D is the data dimension, $D= 784$). The inference nets (encoder) share the same structure of D-500-500-200-40 that maps the observed data into distributional parameters of the latent space. For the PN-based parameterizations, we use a 500 dimensional feature mapping $h$ parameterized by a single layer neural network, and 20 dimensional ID vectors $\mathbf{e}_i$ (see Section \ref{sec:pVAE}) for each variable. We choose the symmetric operator $g$ to be the basic summation operator. 

During training, we apply Adam optimization \citep{adam2015} with default hyperparameter setting, learning rate of 0.001 and a batch size of 100. We generate partially observed MNIST dataset by adding artificially missingness at random in the training dataset during training. We first draw a missing rate parameter from a uniform distribution $\mathcal{U}(0, 0.7)$ and randomly choose variables as unobserved. This step is repeated at each iteration. We train our models for 3K iterations. 

\subsubsection{Image generation of partial VAEs}
\begin{figure}[ht]
\centering
 \subfigure[]{
   \includegraphics[width=0.2\textwidth]{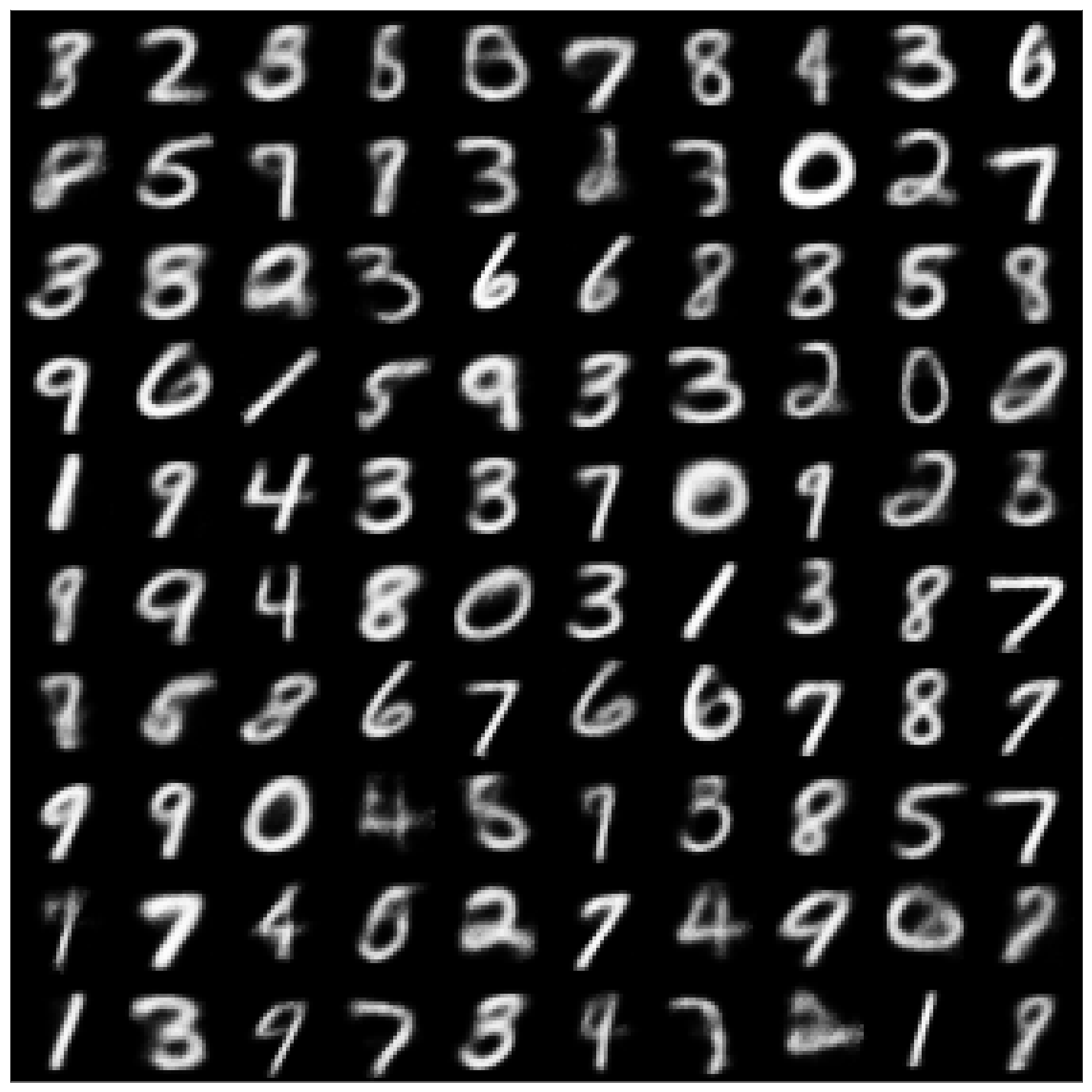}}
    \subfigure[]{\centering
   \includegraphics[width=0.2\textwidth]{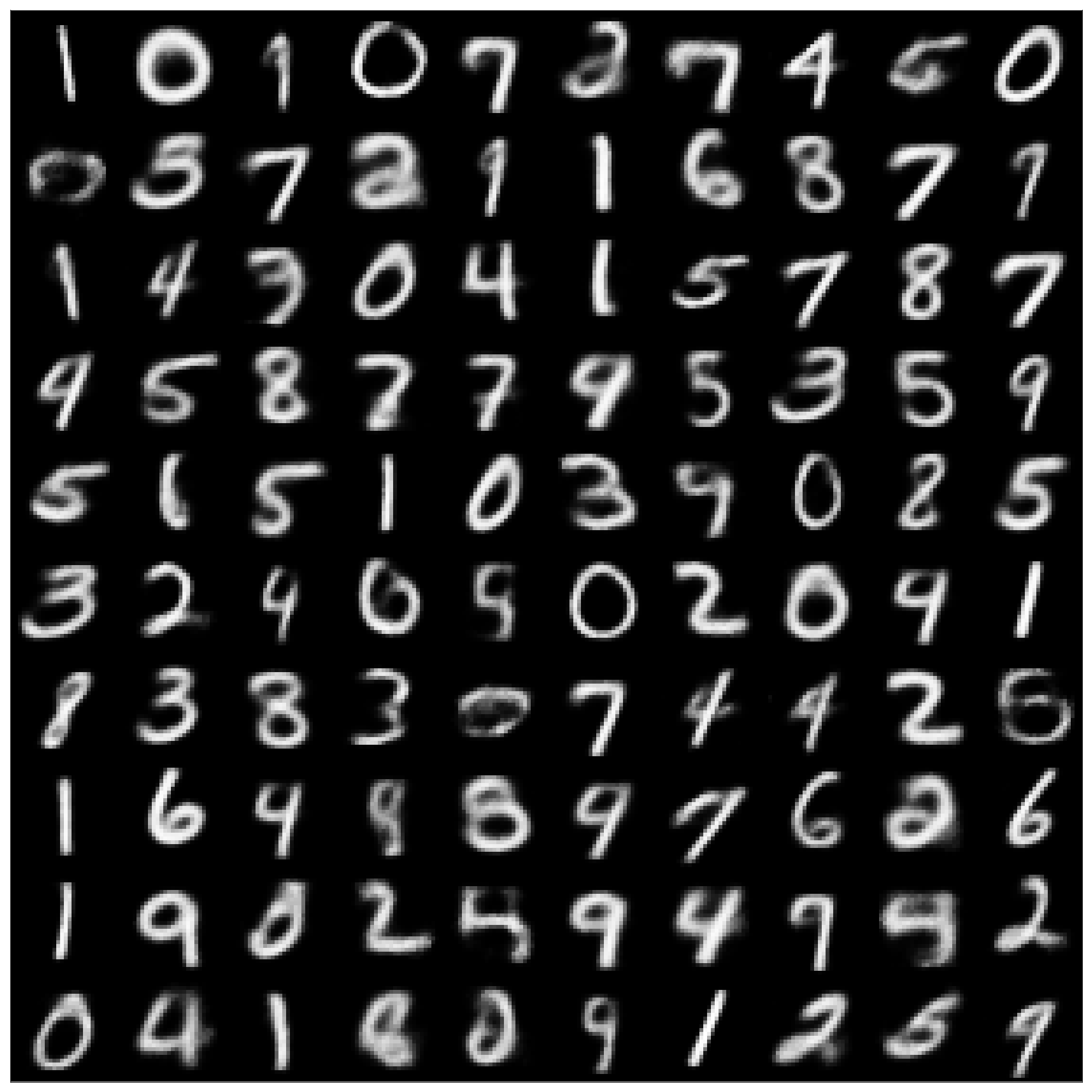}}
   \subfigure[]{
   \includegraphics[width=0.2\textwidth]{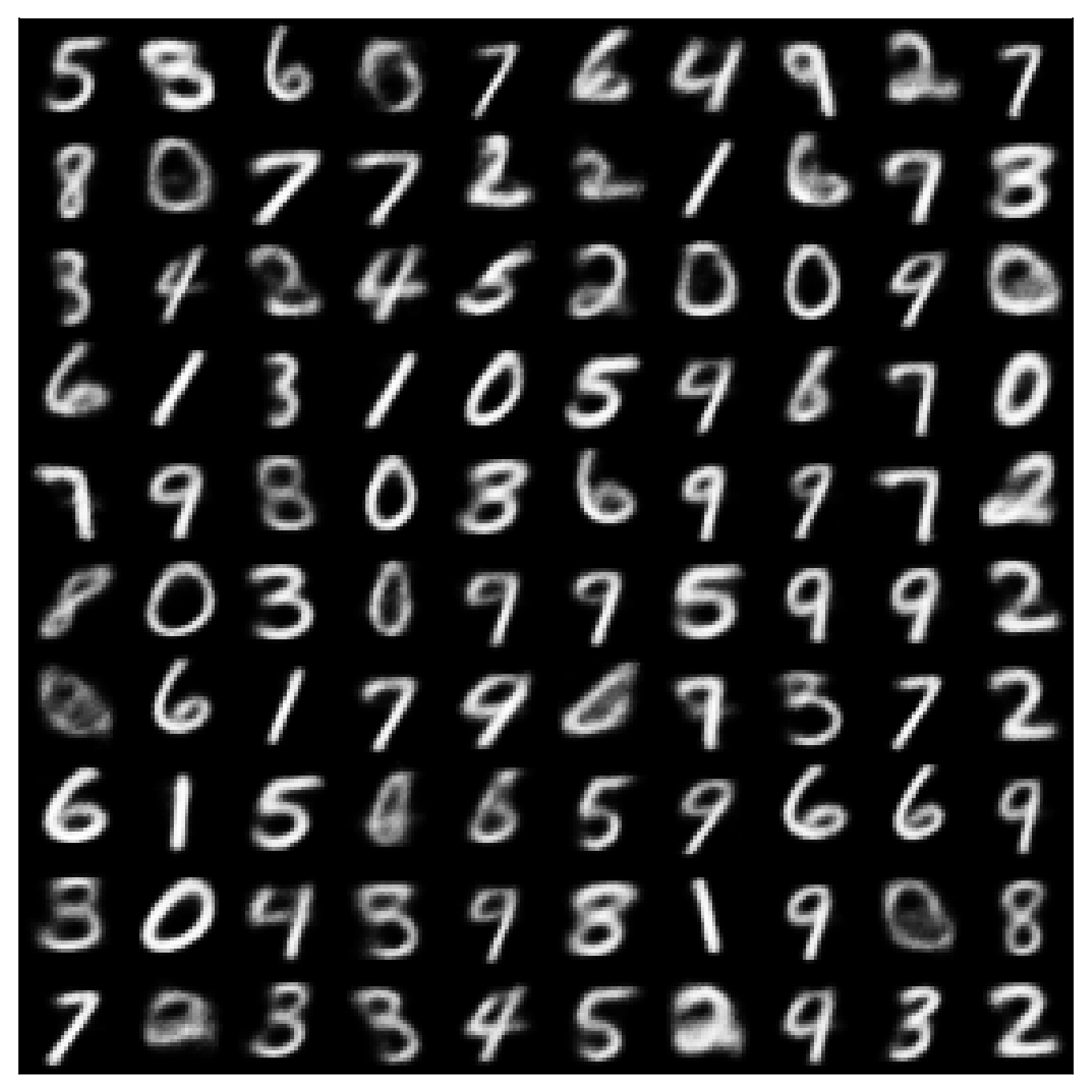}}
   \subfigure[]{\centering
   \includegraphics[width=0.2\textwidth]{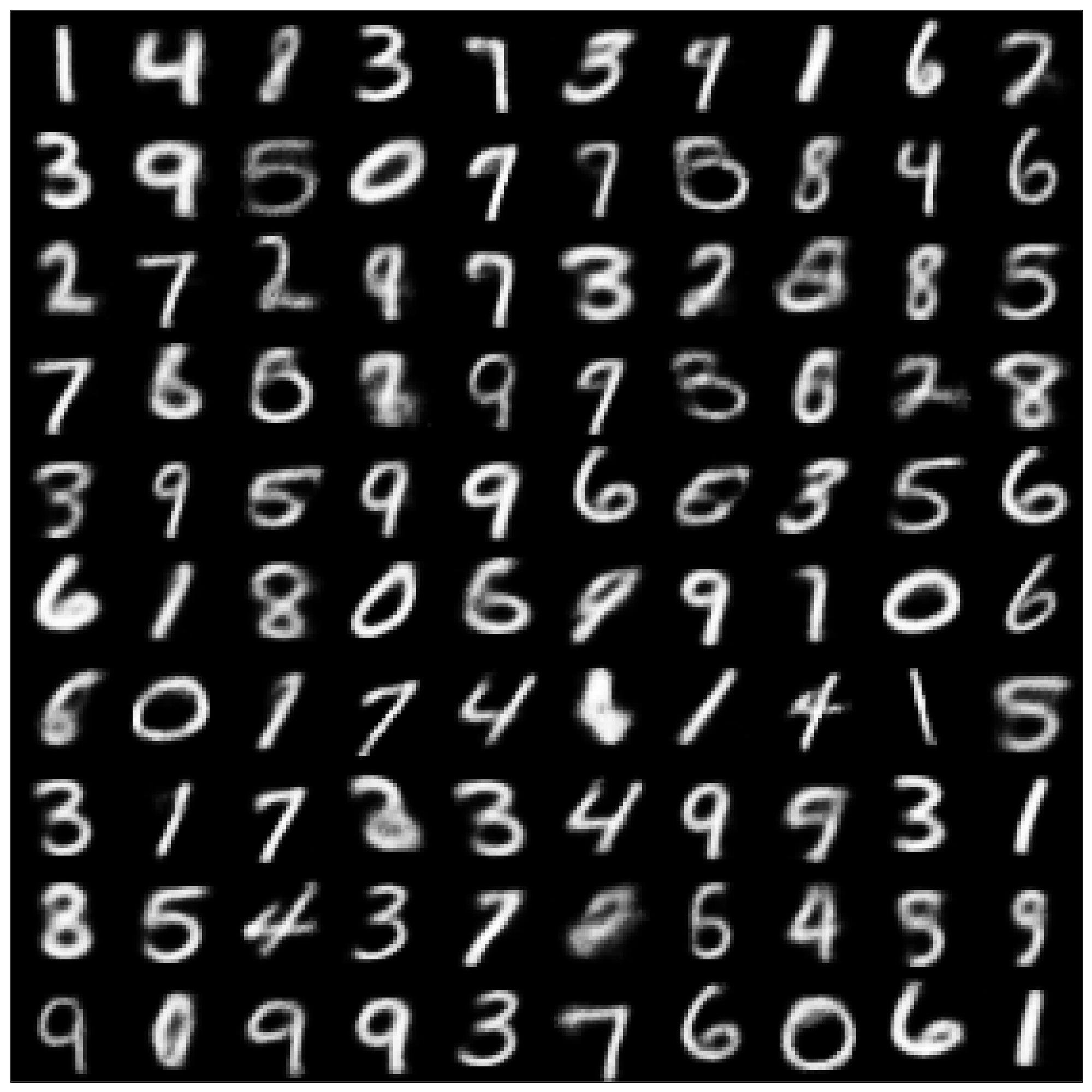}}
   \caption{Random images generated using \textbf{(a)} naive zero imputing, \textbf{(b)} zero imputing with mask, \textbf{(c)} PN and \textbf{(d)} PNP, respectively. }
   \label{fig:mnist_gen}
\end{figure}

\subsection{UCI datasets}
\label{sec:app_UCI}

We applied EDDI on 6 UCI datasets; Boston Housing, Concrete compressive strength, energy efficiency, wine quality, Kin8nm, and Yacht Hydrodynamics. The variables of interest $\mathbf{x}_{\phi}$ are chosen to be the target variables of each UCI dataset in the experiment.

\subsubsection{Preprocessing and model details}
\label{sec:uci_pre}
All data are normalized and then scaled between 0 and 1. For each of the 10 - in total- repetitions, we randomly draw 10\% of the data to be our test set. Partial VAE models (ZI, ZI-m, PNP and PNs) share the same size of architecture with 10 dimensional diagonal Gaussian latent variables: the generator (decoder) is a 10-50-100-D neural network with ReLU activations (where D is the data dimensions). The inference nets (encoder) share the same structure D-100-50-20 that maps the observed data into distributional parameters of the latent space. For the PN-based parameterizations, we further use a 20 dimensional feature mapping $h$ parameterized by a single layer neural network and 10 dimensional ID vectors $\mathbf{e}_i$ (please refer to section \ref{sec:pVAE}) for each variable. We choose the symmetric operator $g$ to be the basic summation operator. 

As in the image inpainting experiment, we apply Adam optimization during training with default hyperparameter setting, and a batch size of 100 and ingest random missingness as before. We trained our models for 3K iterations. 

During active learning, we draw 50 samples in order to estimate the expectation under $\mathbf{x}_{\phi},\mathbf{x}_i \sim p(\mathbf{x}_{\phi}, \mathbf{x}_i|\mathbf{x}_o)$ in Equation (\ref{eq:CHAIN}). Other than information curves based on test RMSEs, we will also provide information curves based on test negative log likelihoods. This will be provided in Appendix \ref{sec:add_plots}. Note that this test nllh of the target variable is also estimated using 50 samples of $\mathbf{x}_{\phi} \sim p(\mathbf{x}_{\phi}|\mathbf{x}_o)$. Then, we approximately compute the (expected) log predictive likelihood through $\log p(\mathbf{x}_{\phi}|\mathbf{x}_o) \approx  \log \frac{1}{M} \sum_{m=1}^M p(\mathbf{x}_{\phi}|\mathbf{z}_m)$, where $\mathbf{z}_m \sim q(\mathbf{z}|\mathbf{x}_o)$.

\subsubsection{Statistical signifcant test results}
\label{sec:wilcox}

In this section, we perform Wilcoxcon signed-rank significance test on the AUIC (RMSE-based) performance of different methods, to support our result in Table \ref{tab:uci_auic}. Since Table \ref{tab:uci_auic} suggests that EDDI-PNP-Partial VAE is the best algorithm overall, we set EDDI-PNP-Partial VAE as default and perform Wilcoxcon test between EDDI-PNP-Partial VAE and all other 15 different settings, to see whether the improvement is significant. Table \ref{tab:wilcox} displays the corresponding p-value for each test. It is obvious that in all 15 tests, the EDDI-PNP-Partial VAE results are significant (compared with the standard $\alpha = 0.05$ cutoff). This provides strong evidence that confirms our results in Table \ref{tab:uci_auic}.

\begin{table}[H]
\centering
\caption{p- values of Wilcoxon signed-rank test of EDDI-PNP vs. 11 other settings, on 6 UCI datasets, using AUIC (RMSE-based) as evaluation metric.  }
\scalebox{0.8}{
\begin{tabular}{p{1.in}p{0.8in}p{0.8in}p{0.8in}p{0.8in}} \toprule
\textbf{Method} &ZI & ZI-m & PNP & PN \\ \midrule
EDDI & $< 10^{-48}$ & $<10^{-23}$ & N/A & $<10^{-2}$\\ 
Random & $0$ & $0$ & $0$ & $0$\\ 
Single best & $0$ & $<10^{-13}$ & $<10^{-2}$& $<10^{-4}$\\ 
\bottomrule
\end{tabular}}
\label{tab:wilcox}
\vspace{-10pt}
\end{table}

\subsubsection{Additional plots of PN, ZI and ZI-m on UCI datasets}
\begin{figure}[H]
  \begin{tabular}{c c c} 
  \hline Boston Housing & Energy & Wine \\
      \hline
       \includegraphics[width=0.32\textwidth]{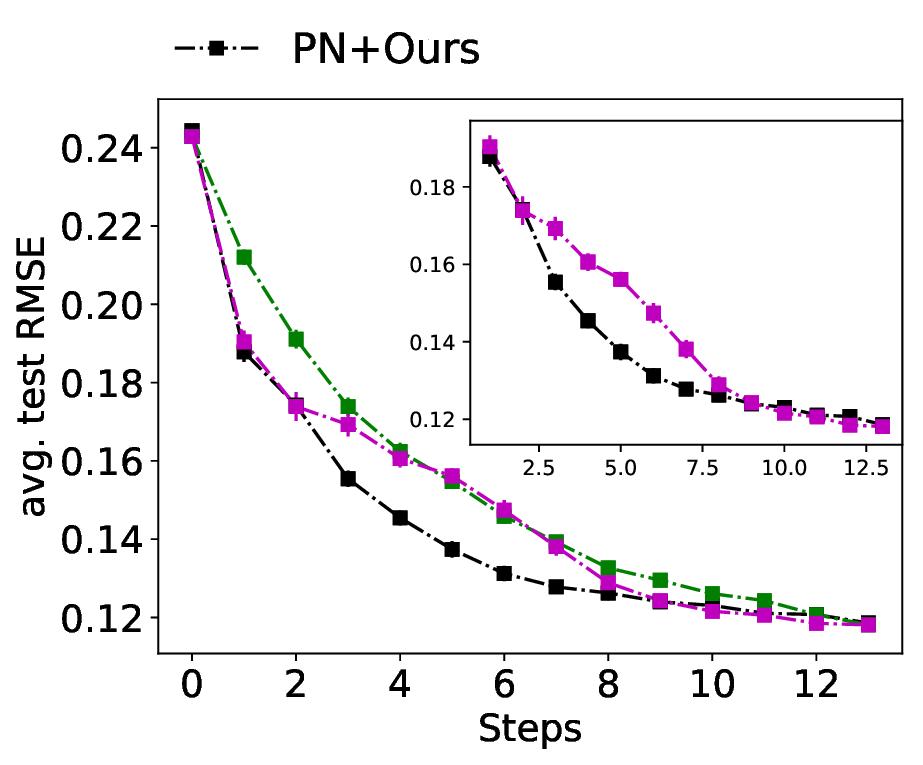} &
        \includegraphics[width=0.32\textwidth]{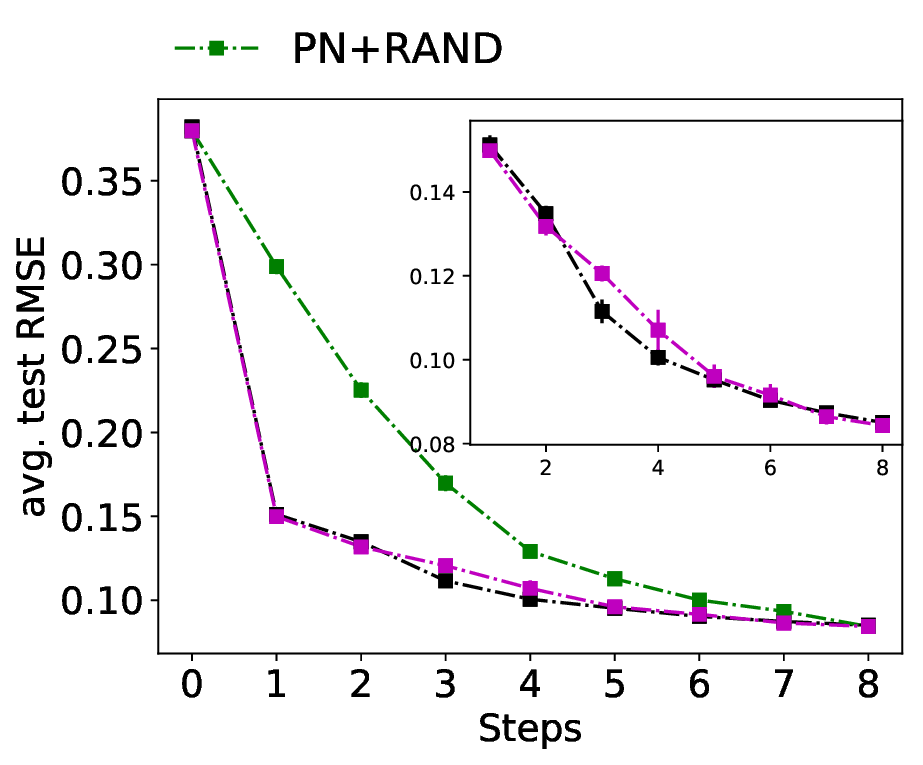}& 
        \includegraphics[width=0.32\textwidth]{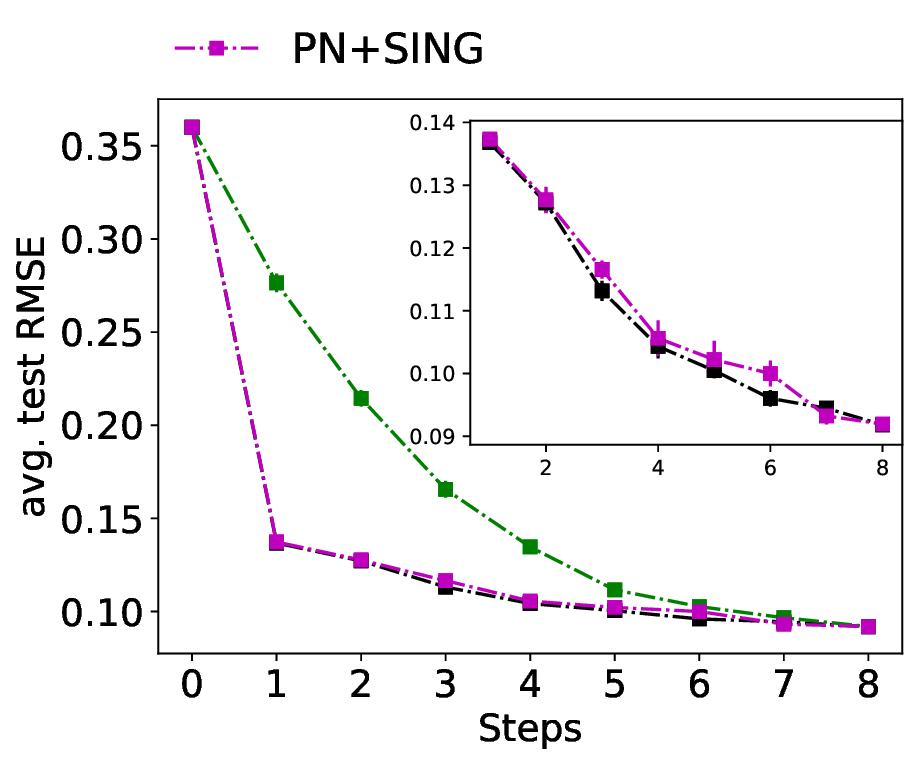}\\
          \includegraphics[width=0.32\textwidth]{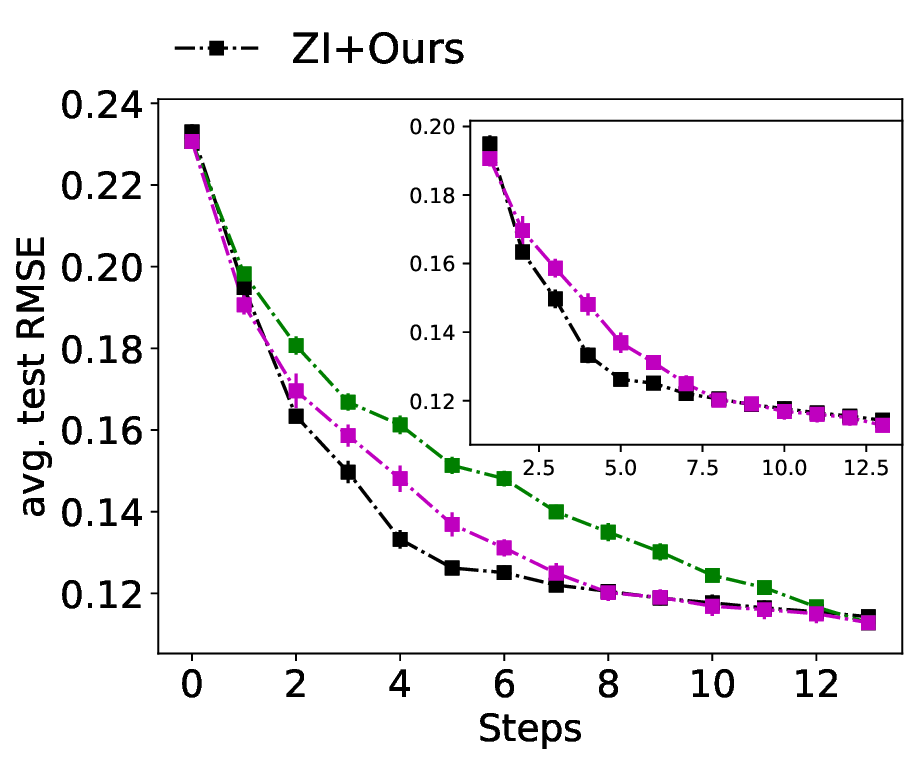} &
          \includegraphics[width=0.32\textwidth]{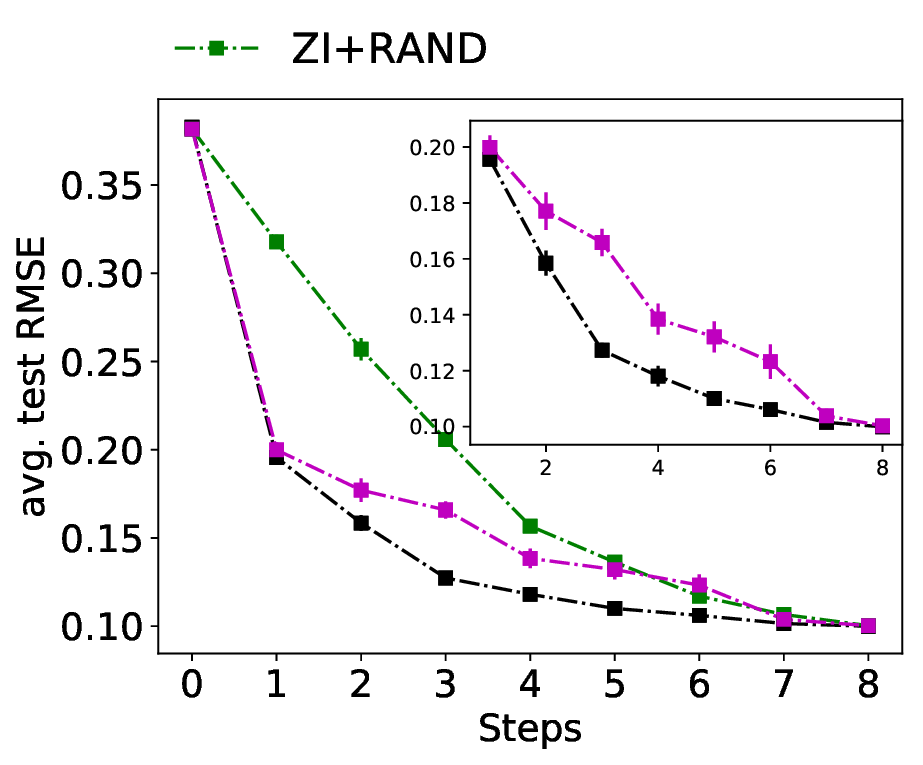} &
           \includegraphics[width=0.32\textwidth]{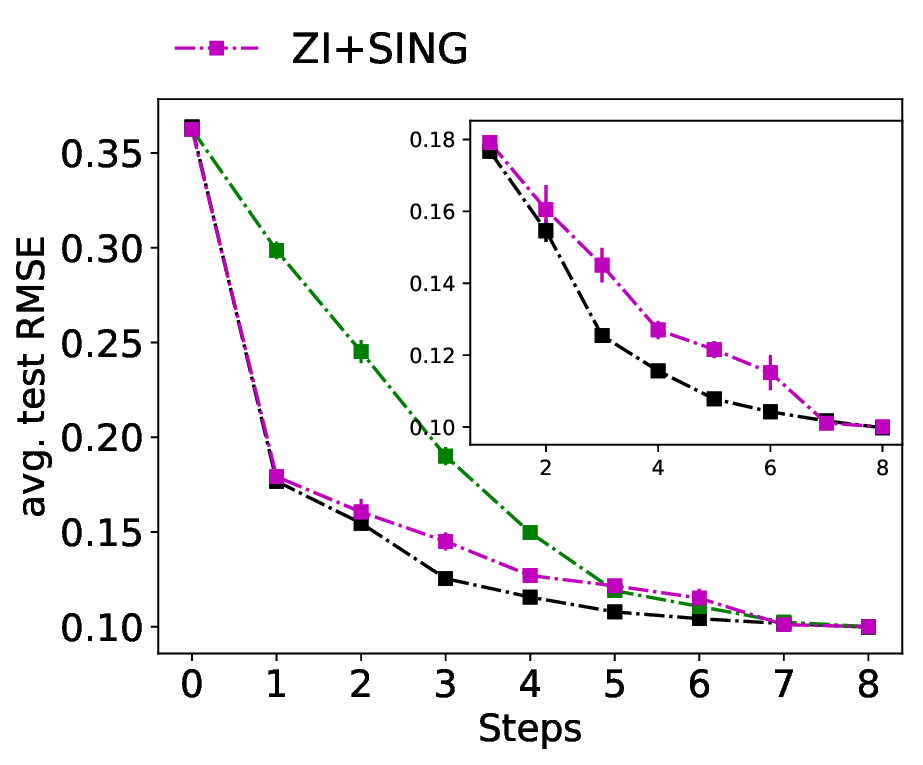}\\
           \includegraphics[width=0.32\textwidth]{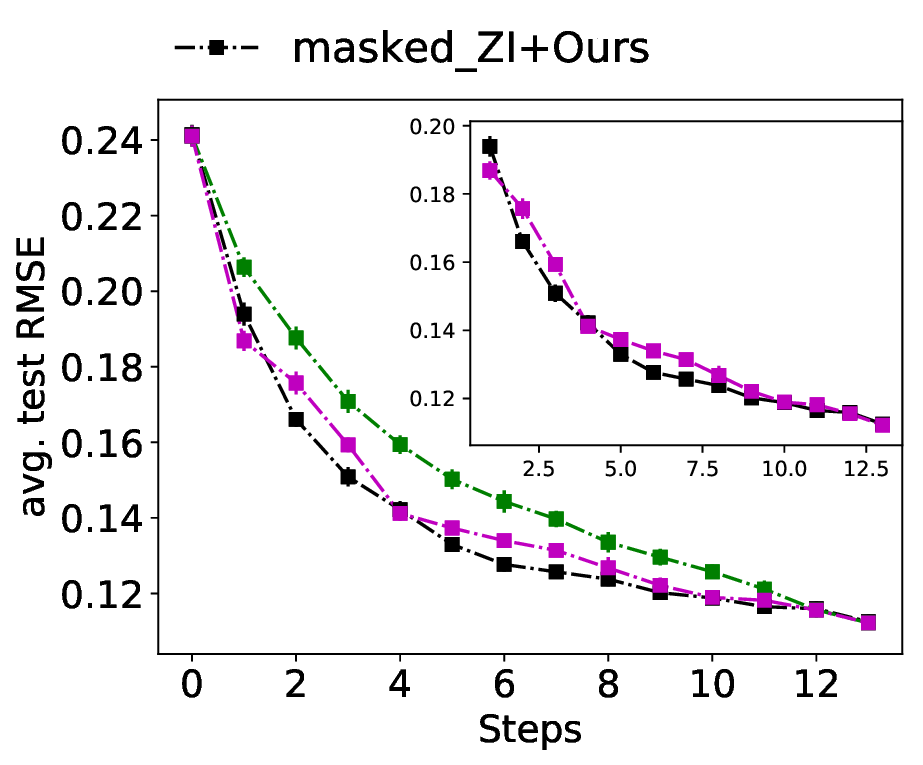} &
           \includegraphics[width=0.32\textwidth]{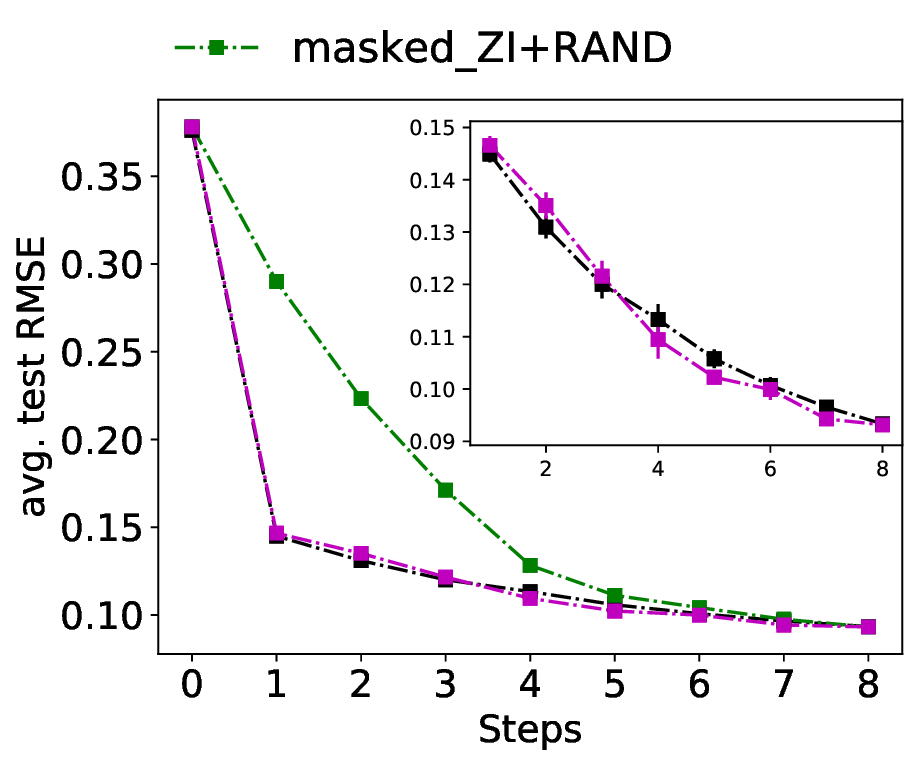} &
           \includegraphics[width=0.32\textwidth]{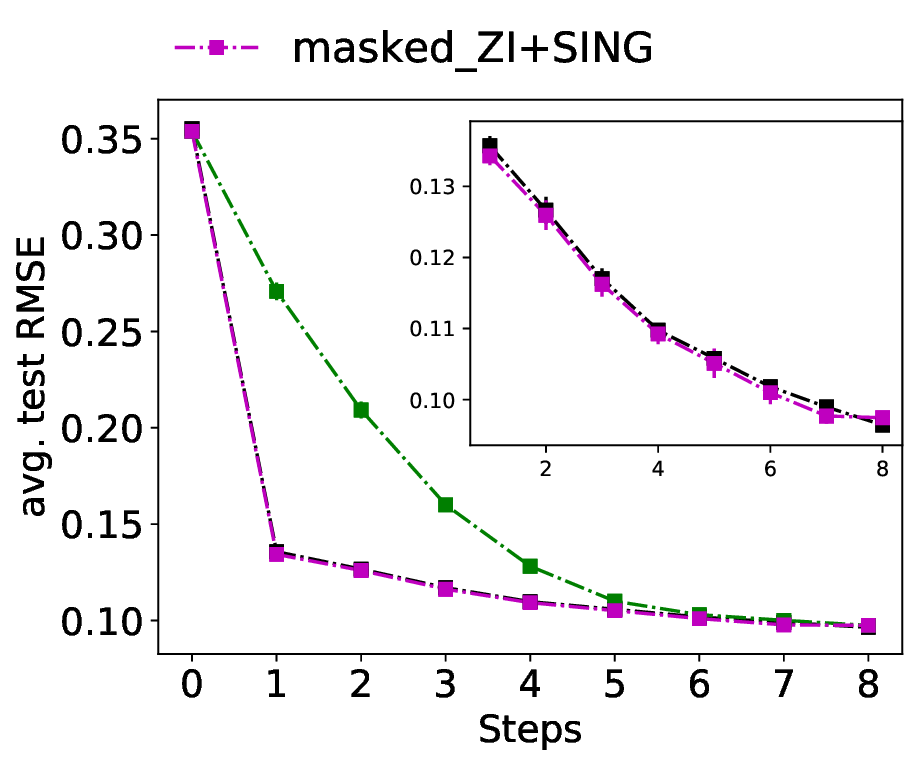} \\
           \hline                 
   \end{tabular}
    \caption{information curves (based on RMSE) of active variable selection for the three UCI datasets and the three approaches, i.e. \textbf{(First row)} PointNet (PN), \textbf{(Second row)} Zero Imputing (ZI), and \textbf{(Third row)} Zero Imputing with mask (ZI-m). \textbf{Green}: random strategy; \textbf{Black}: EDDI; \textbf{Pink}: Single best ordering. This displays RMSE (y axis, the lower the better) during the course of active selection (x-axis).} 
    \label{fig:uci_rmse}
\end{figure}

Here we present additional plots of the RMSE information curves during active learning. Figure \ref{fig:uci_rmse} presents the results for the Boston Housing, the Energy and the Wine datasets and for the three approaches, i.e. PN, ZI and masked ZI.

\subsubsection{Negative test log Likelihood plots of PN, ZI and ZI-m on UCI datasets}\label{sec:add_plots}
\begin{figure}[H]
  \begin{tabular}{c c c} 
  \hline Boston Housing & Energy & Wine \\
      \hline
       \includegraphics[width=0.32\textwidth]{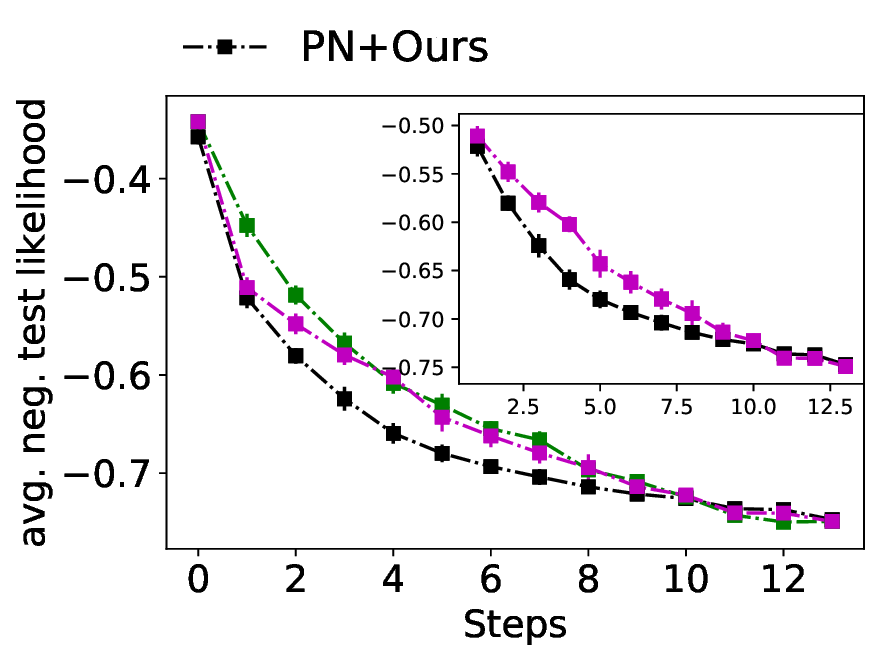} &
        \includegraphics[width=0.32\textwidth]{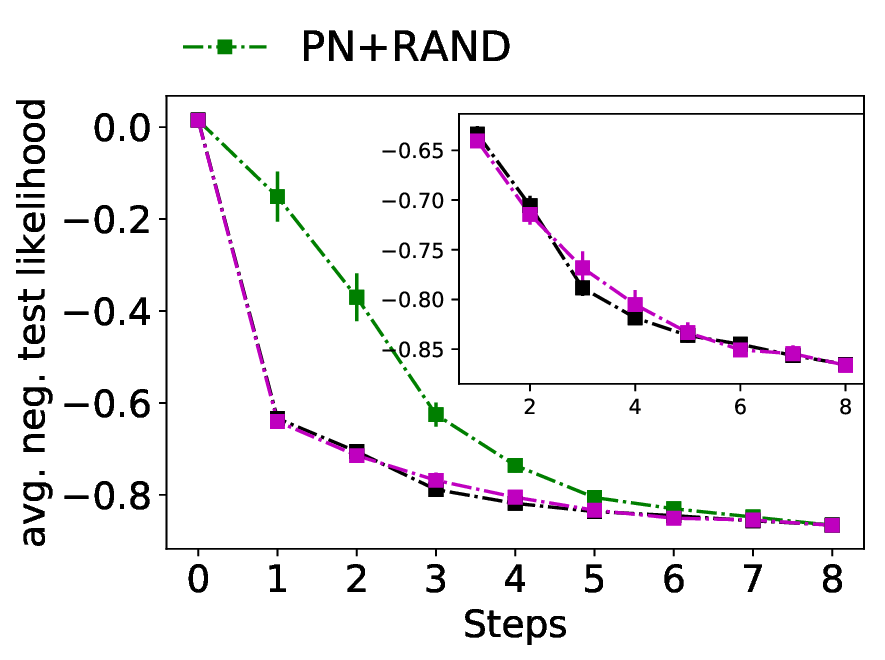}& 
        \includegraphics[width=0.32\textwidth]{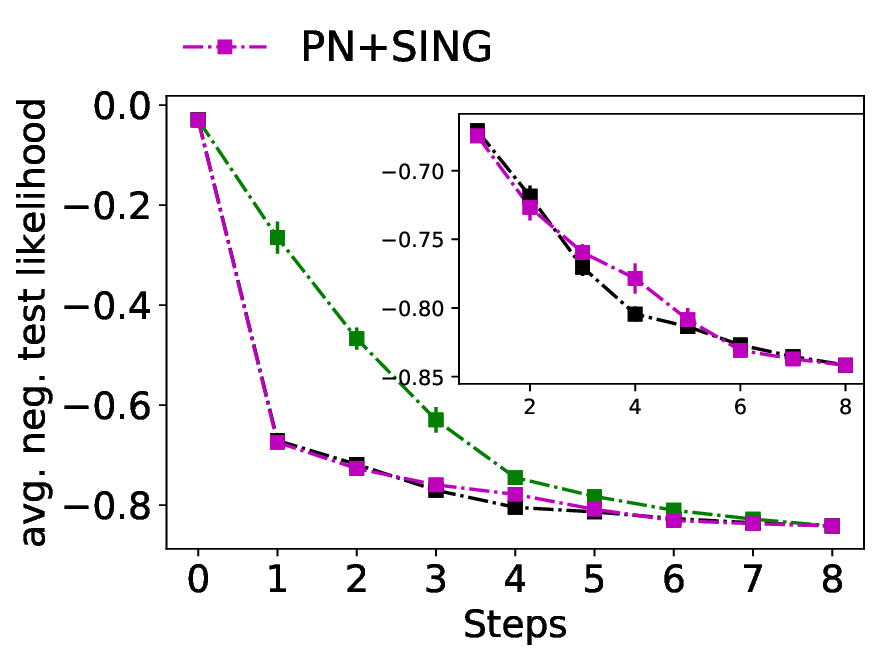}\\
          \includegraphics[width=0.32\textwidth]{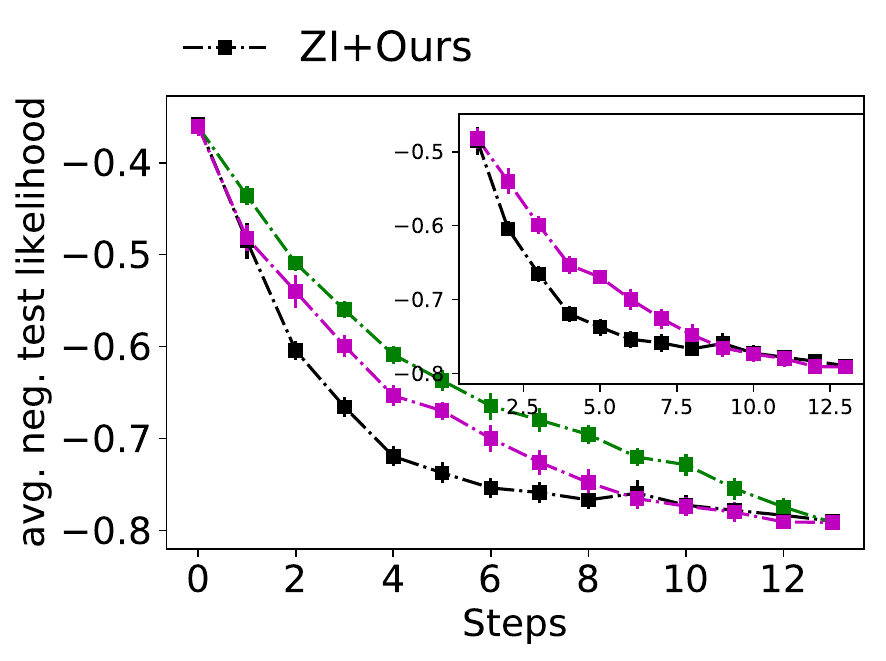} &
          \includegraphics[width=0.32\textwidth]{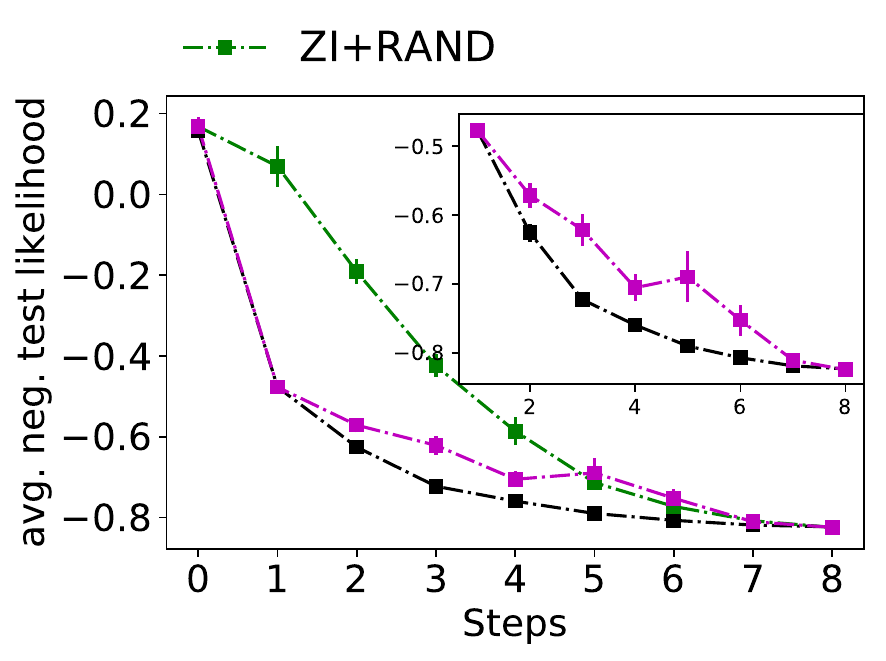} &
           \includegraphics[width=0.32\textwidth]{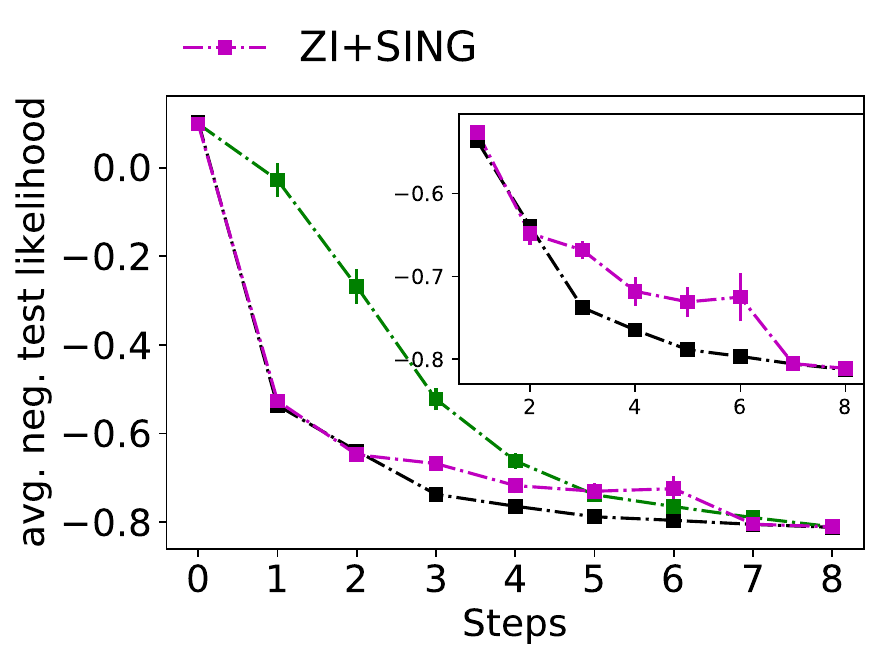}\\
           \includegraphics[width=0.32\textwidth]{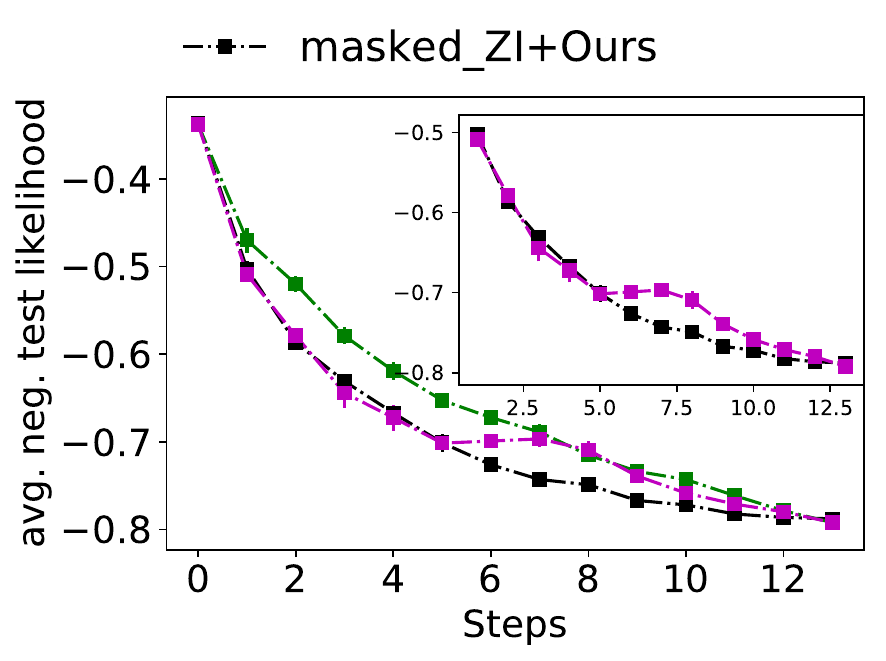} &
           \includegraphics[width=0.32\textwidth]{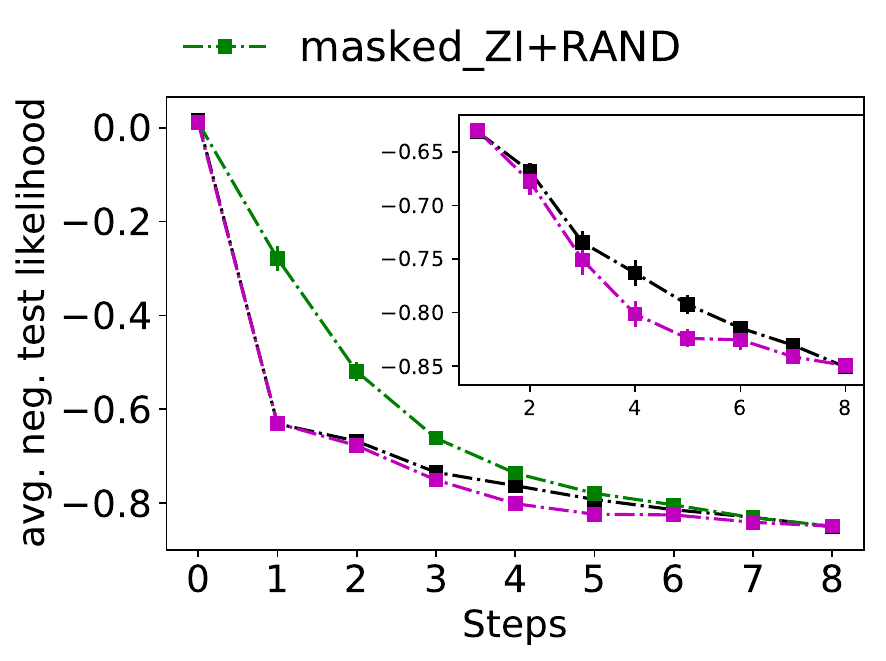} &
           \includegraphics[width=0.32\textwidth]{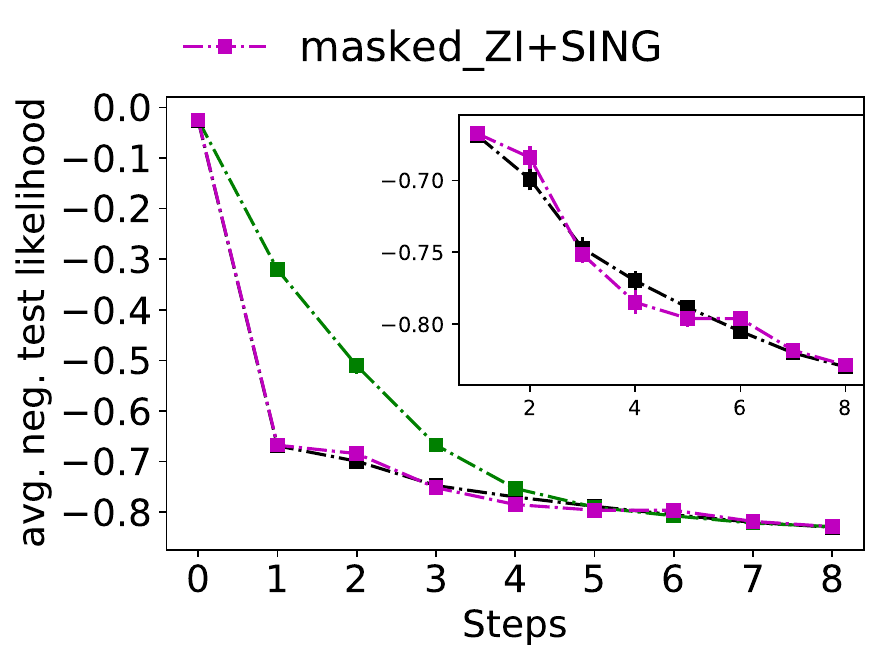} \\
           \hline                 
   \end{tabular}
    \caption{Information curves (based on test negative log-likelihood) of active variable selection for the three UCI datasets and the three approaches, i.e. \textbf{(First row)} PointNet (PN), \textbf{(Second row)} Zero Imputing (ZI), and \textbf{(Third row)} Zero Imputing with mask (ZI-m). \textbf{Green}: random strategy; \textbf{Black}: EDDI; \textbf{Pink}: Single best ordering. This displays negative test log likelihood (y axis, the lower the better) during the course of active selection (x-axis).} 
    \label{fig:uci_all}
\end{figure}

Here we present additional plots of the negative test log likelihood curves during active variable selection. Figure \ref{fig:uci_all} presents the results for the Boston Housing, the Energy and the Wine datasets and for the three approaches, i.e. PN, ZI and masked ZI.
 
\subsubsection{Comparisons between EDDI and LASSO-based method}
Here we present additional results of a new baseline, the LASSO-based feature selection. This is not presented in the main text since LASSO is designed for a different problem setting. It requires fully observed data, and only works in regression problems with one dimensional outputs. Both MIMIC III and NHANES tasks do not fulfill these requirements. Additionally, LASSO aims to select a global set of features to obtain the best performance instead of select the most informative feature given partially observed information, thus cannot be used in a sequential setting.  We thus construct  the LASSO feature selection baseline as follows for comparison: we first apply LASSO regression on training dataset which is fully observed in these UCI datasets, and select the features (denoted by $\mathcal{A}$) that correspond to non-zero coefficients. Then, during test time, LASSO strategy will observe the features one by one from $\mathcal{A}$ randomly. When all variables selected by LASSO are already picked, we stop the feature selection progress. Once LASSO has completed feature selection, we use we use the corresponding partial-VAE (ZI,ZI-m,PNP,PN) to make predictions for fairness.

\begin{figure}[H]
  \begin{tabular}{c c c} 
  \hline Boston Housing & Energy & Wine \\
      \hline
       \includegraphics[width=0.32\textwidth]{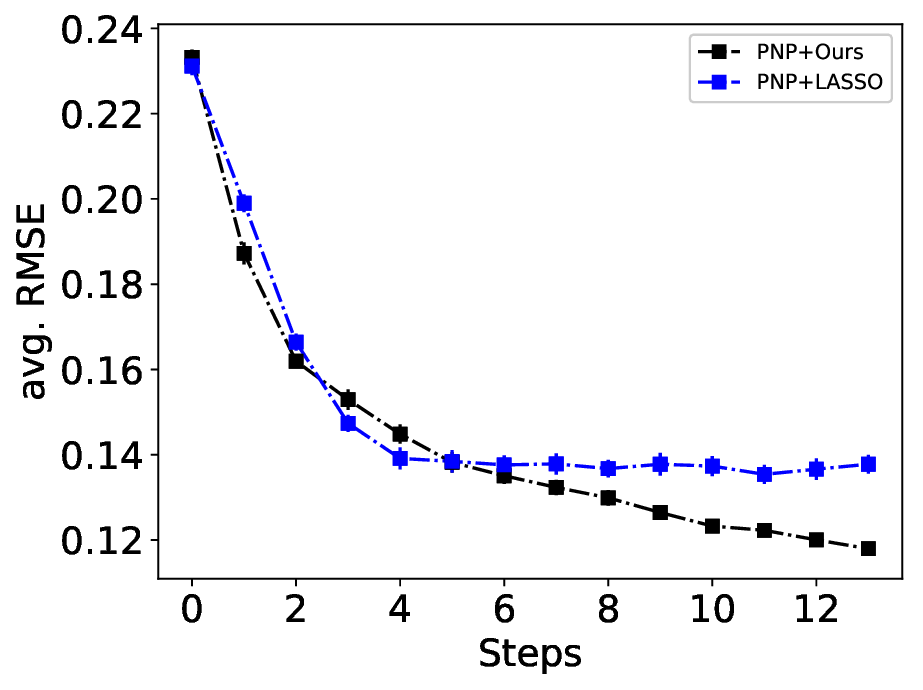} &
        \includegraphics[width=0.32\textwidth]{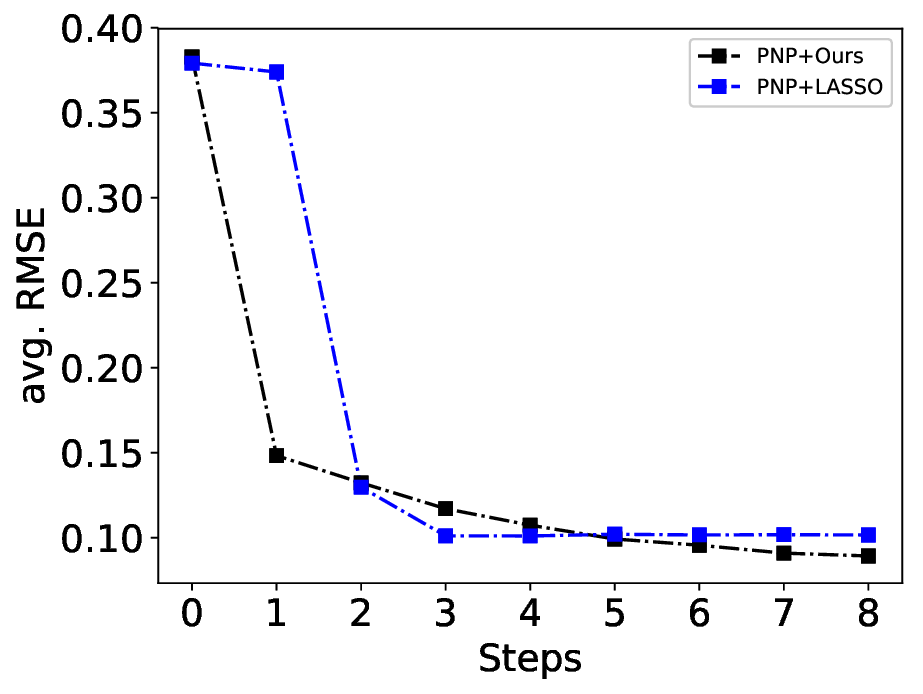}& 
        \includegraphics[width=0.32\textwidth]{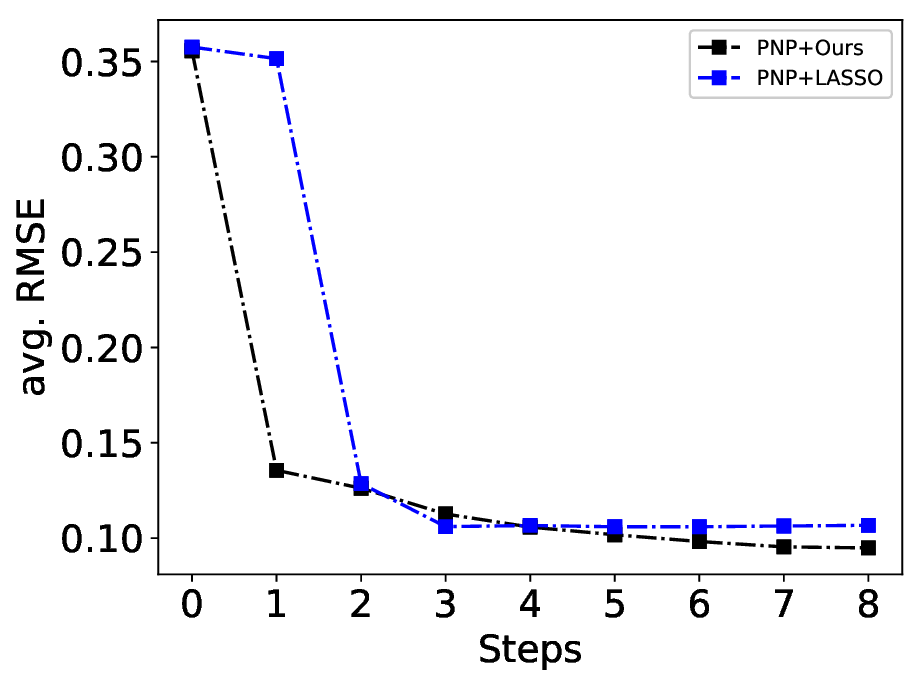}\\
           \hline                 
   \end{tabular}
    \caption{Information curves of active variable selection for the three UCI datasets and PNP-Partial VAE. \textbf{Black}: EDDI; \textbf{Blue}: Single best ordering. This displays test RMSE (y axis, the lower the better) during the course of active selection (x-axis).} 
    \label{fig:uci_lasso}
\end{figure}

Figure \ref{fig:uci_lasso} presents the results for the Boston Housing, the Energy and the Wine datasets as examples. Full results of all UCI datasets are presented in Table \ref{tab:uci_lasso}. Note that in Table \ref{tab:uci_lasso}, Wilcoxon signed-rank test is performed between EDDI and LASSO strategies for each Partial VAE models, respectively. The results indicates that EDDI significantly outperforms LASSO in all circumstances. This is despite the fact that EDDI is a greedy sequential variable selection method that built upon partially observed data, while LASSO-baseline makes use of the information from \emph{fully observed data}, and selects the set of variables in a \emph{non-greedy, global manner}, which is often unrealistic in many pratical application settings.

\begin{table}[H]
\centering
\caption{Avg. rankings of AUIC (RMSE-based), and p- values of Wilcoxon signed-rank test that EDDI outperforms LASSO (on 6 UCI datasets).  }
\scalebox{0.8}{
\begin{tabular}{p{1.in}p{0.8in}p{0.8in}p{0.8in}p{0.8in}} \toprule
\textbf{Method} &ZI & ZI-m & PNP & PN \\ \midrule
EDDI & 4.66 (0.02) & 4.53(0.02) & \textbf{4.14}(0.02) & 4.24(0.02)\\ 
LASSO & 4.86(0.02) & 4.63(0.02) & 4.41(0.02) & 4.48(0.02) \\ 
p-value & $<10^{-4}$& $<10^{-6}$ & $<10^{-24}$ & $<10^{-19}$\\ 
\bottomrule
\end{tabular}}
\label{tab:uci_lasso}
\vspace{-10pt}
\end{table}

\subsubsection{Illustration of decision process of EDDI (Boston Housing as example)}
\begin{figure}[!h] 
\centering
\subfigure[]{\centering
   \includegraphics[width=0.4\textwidth]{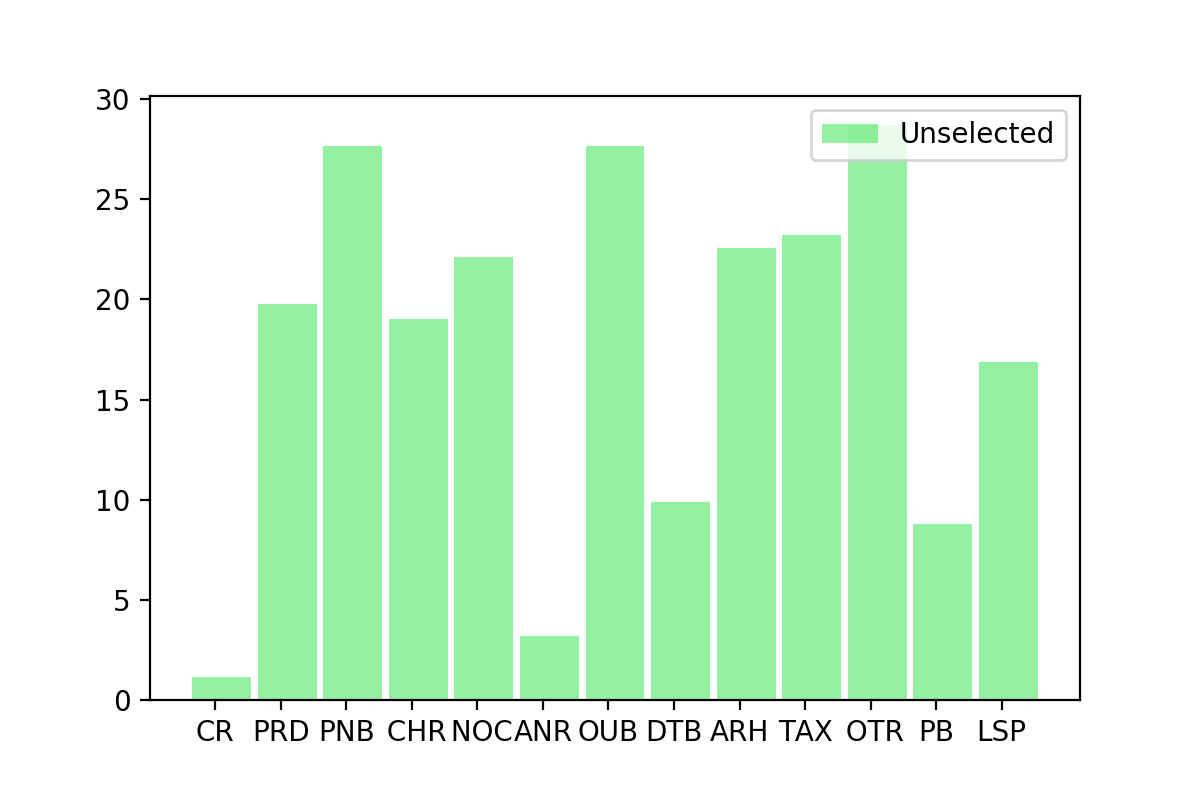}}
    \subfigure[]{\centering
   \includegraphics[width=0.4\textwidth]{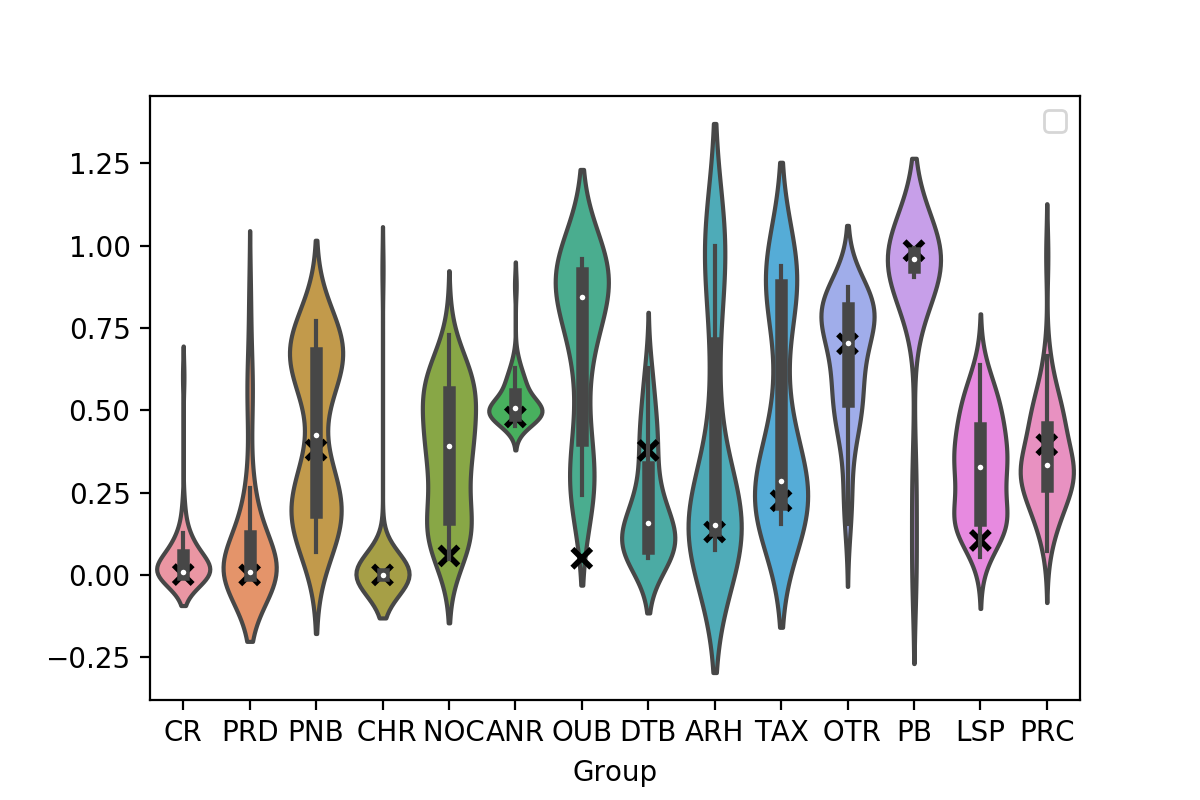}} \\
   \subfigure[]{\centering
   \includegraphics[width=0.4\textwidth]{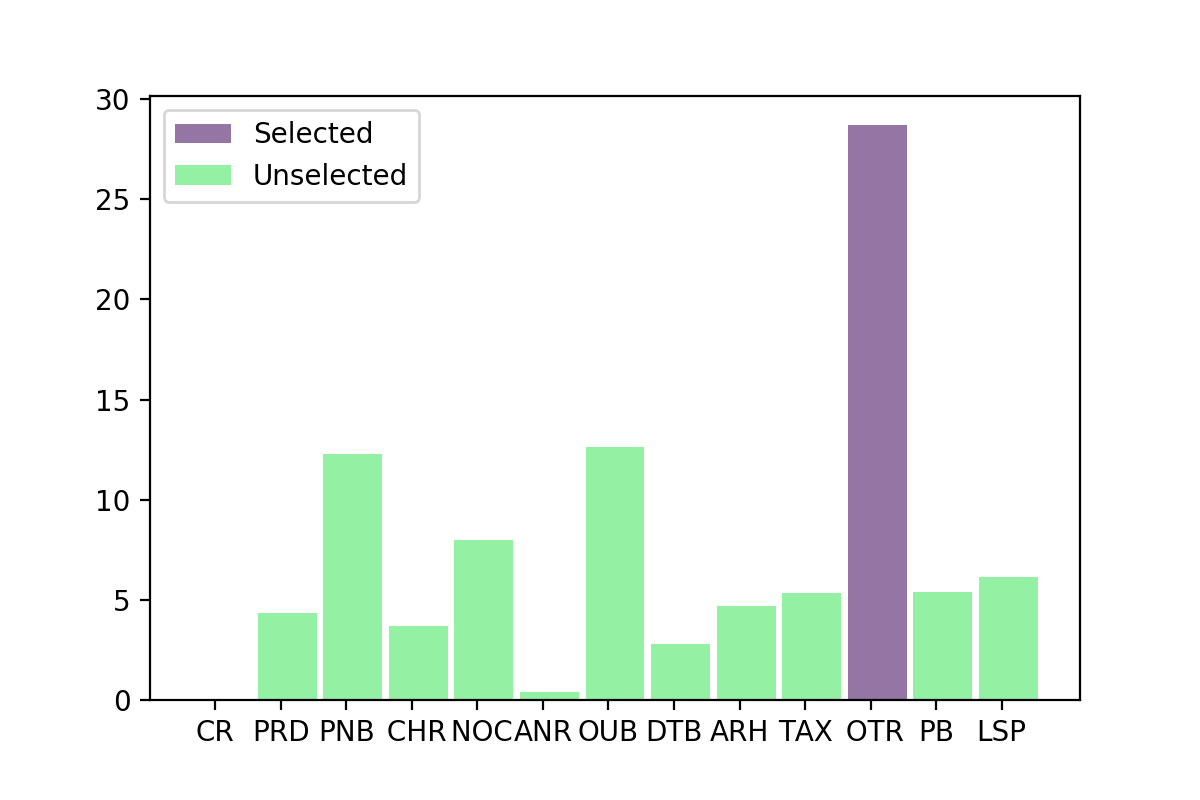}}
    \subfigure[]{\centering
   \includegraphics[width=0.4\textwidth]{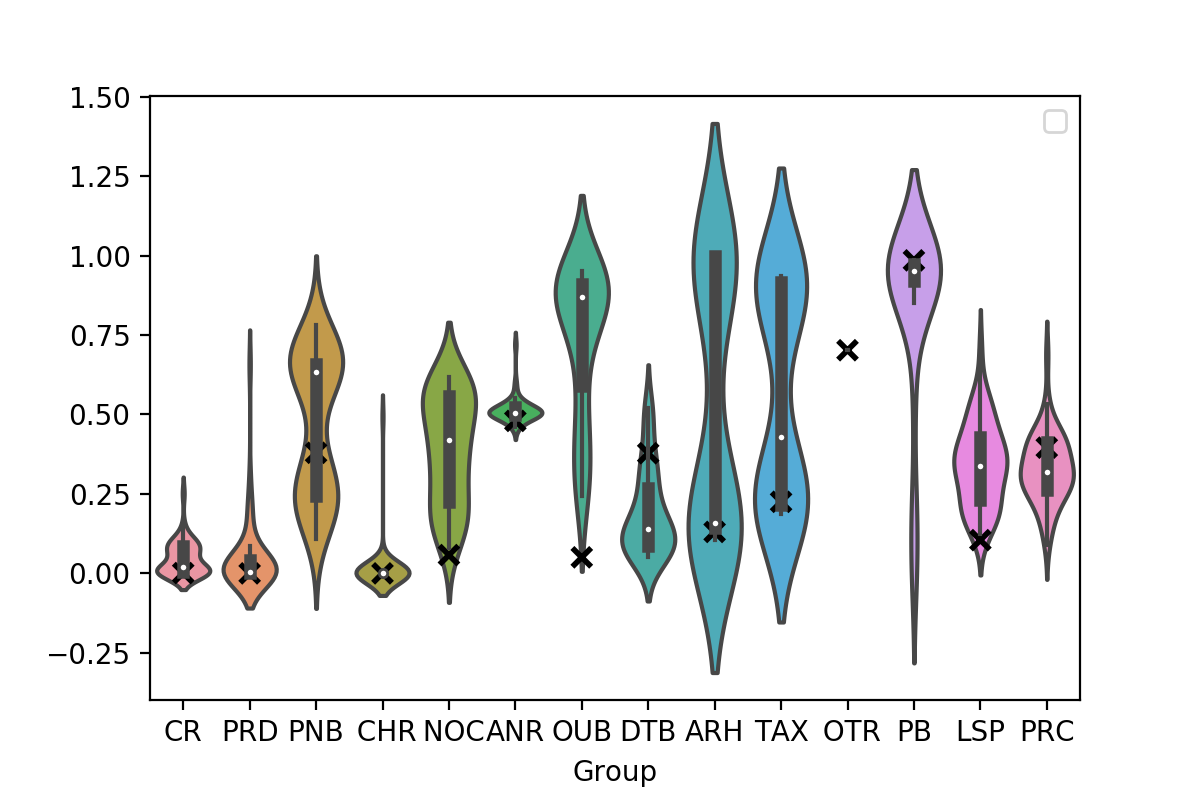}} \\
   \subfigure[]{\centering
   \includegraphics[width=0.4\textwidth]{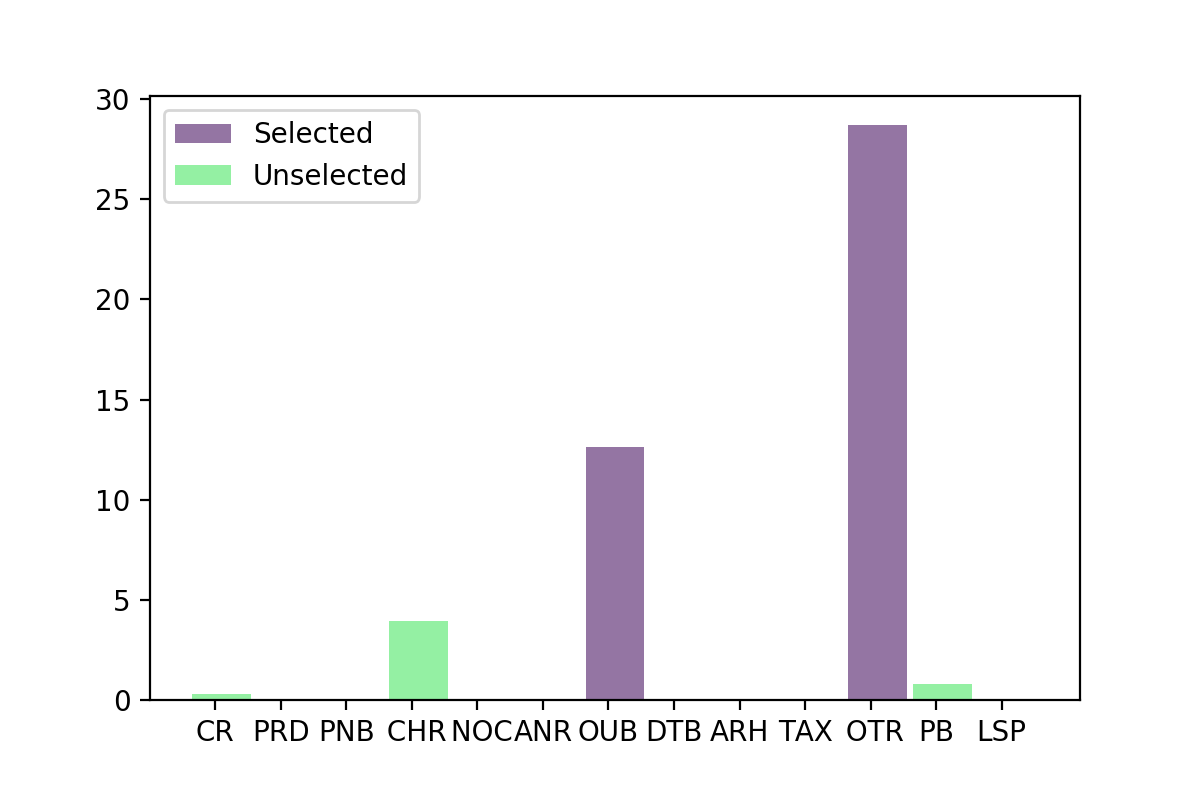}}
    \subfigure[]{\centering
   \includegraphics[width=0.4\textwidth]{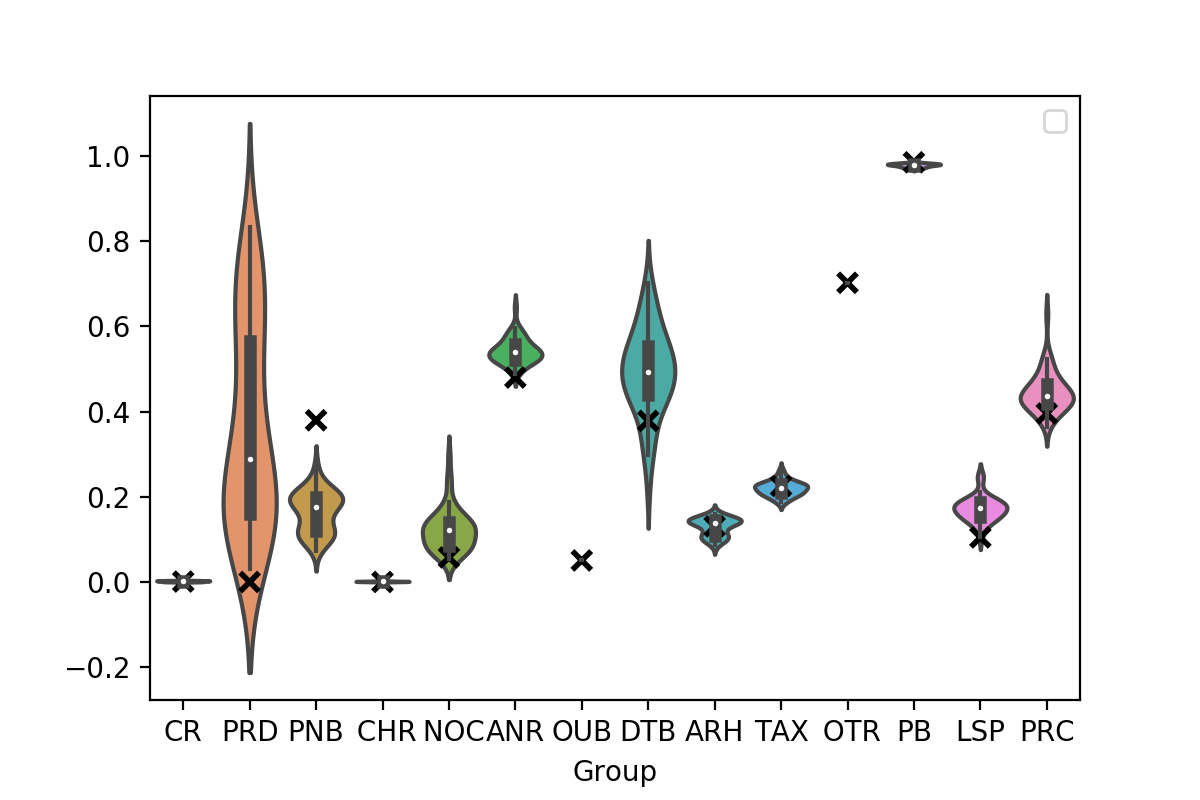}} \\
      \subfigure[]{\centering
   \includegraphics[width=0.4\textwidth]{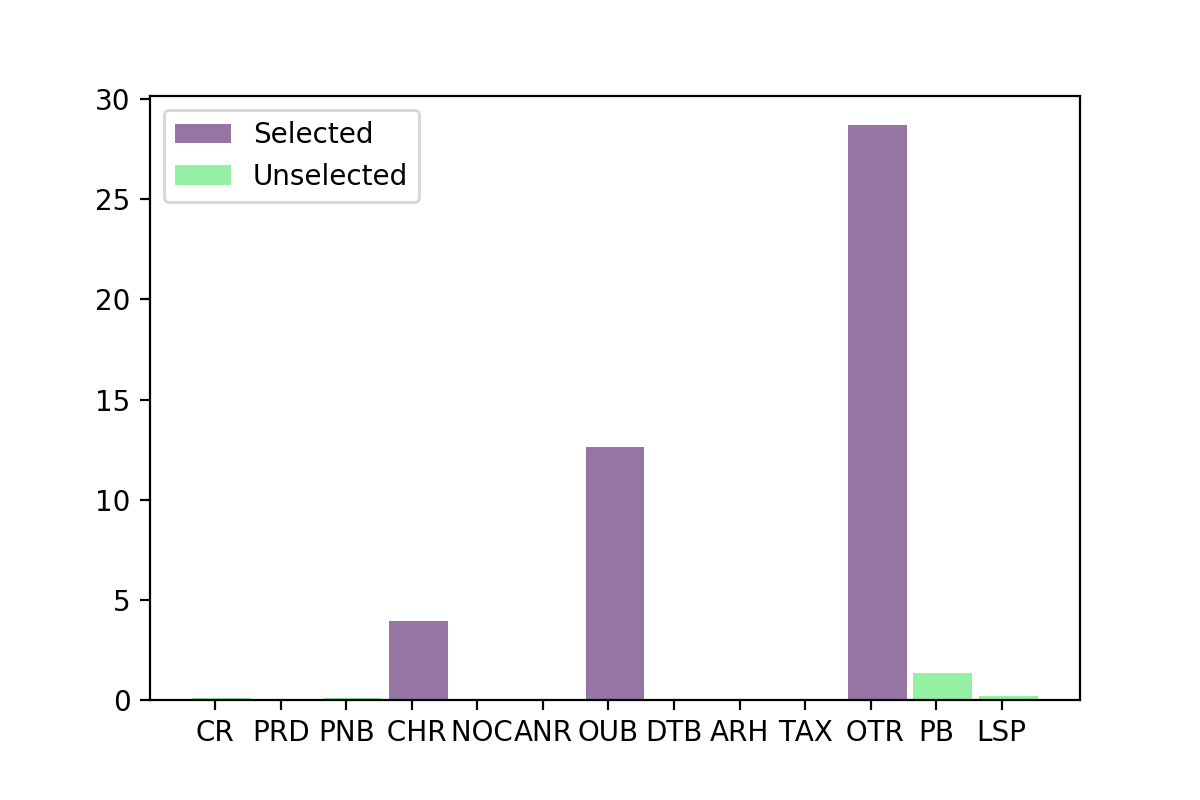}}
    \subfigure[]{\centering
   \includegraphics[width=0.4\textwidth]{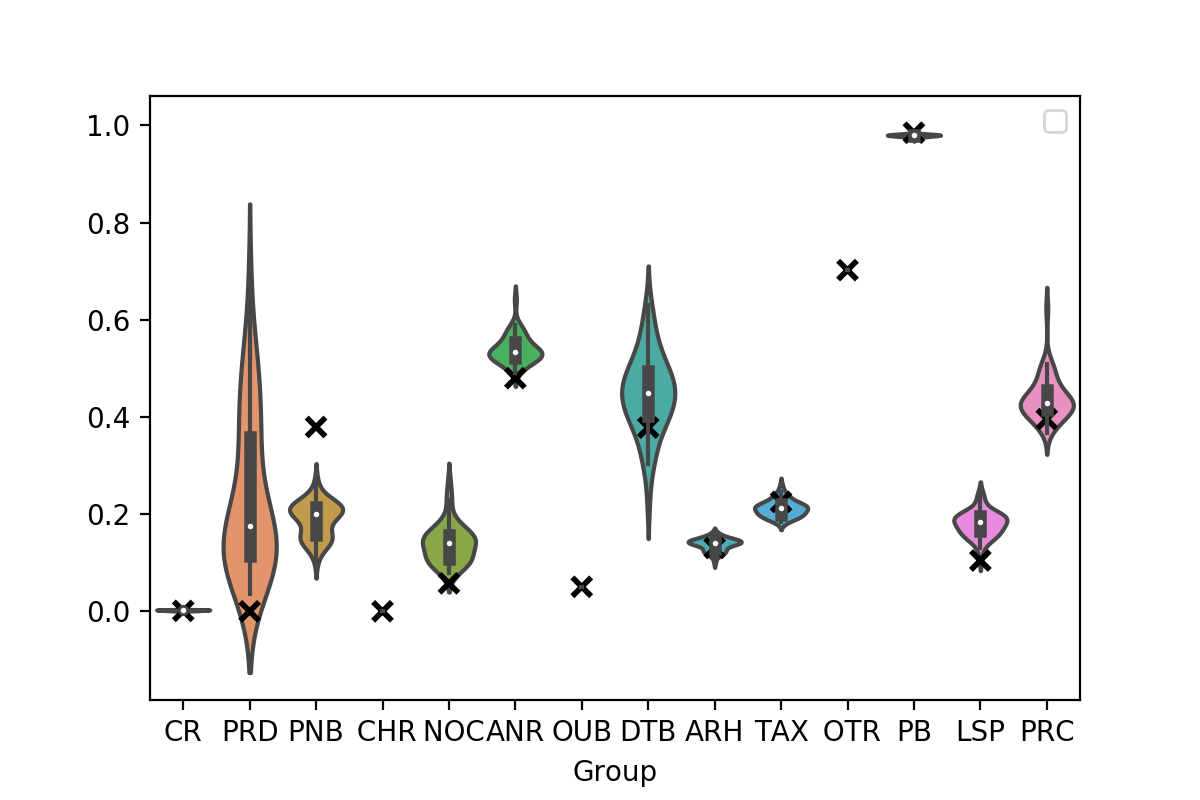}} \\
   \caption{Information reward estimated during the first 4 active variable selection steps on a randomly chosen Boston Housing test data point.  \textbf{Model}: PNP, strategy: EDDI. Each row contains two plots regarding the same time step. \textbf{Bar plots on the left} show the information reward estimation of each variable on the y-axis. All unobserved variables start with green bars, and turns purple once selected by the algorithm.  \textbf{Right}: violin plot of the posterior density estimations of remaining unobserved variables. }. 
   \label{fig:bar_chai}
\end{figure}

\begin{figure}[!h] 
\centering
\subfigure[]{\centering
   \includegraphics[width=0.4\textwidth]{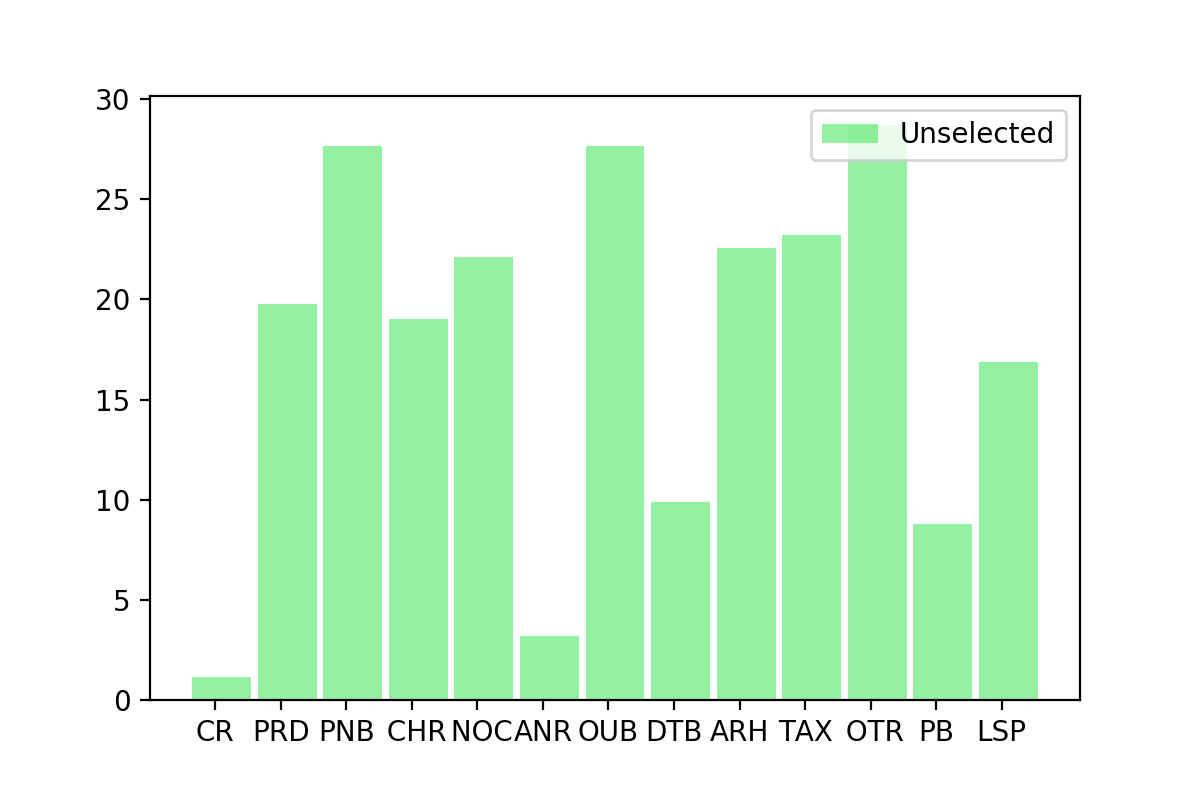}}
    \subfigure[]{\centering
   \includegraphics[width=0.4\textwidth]{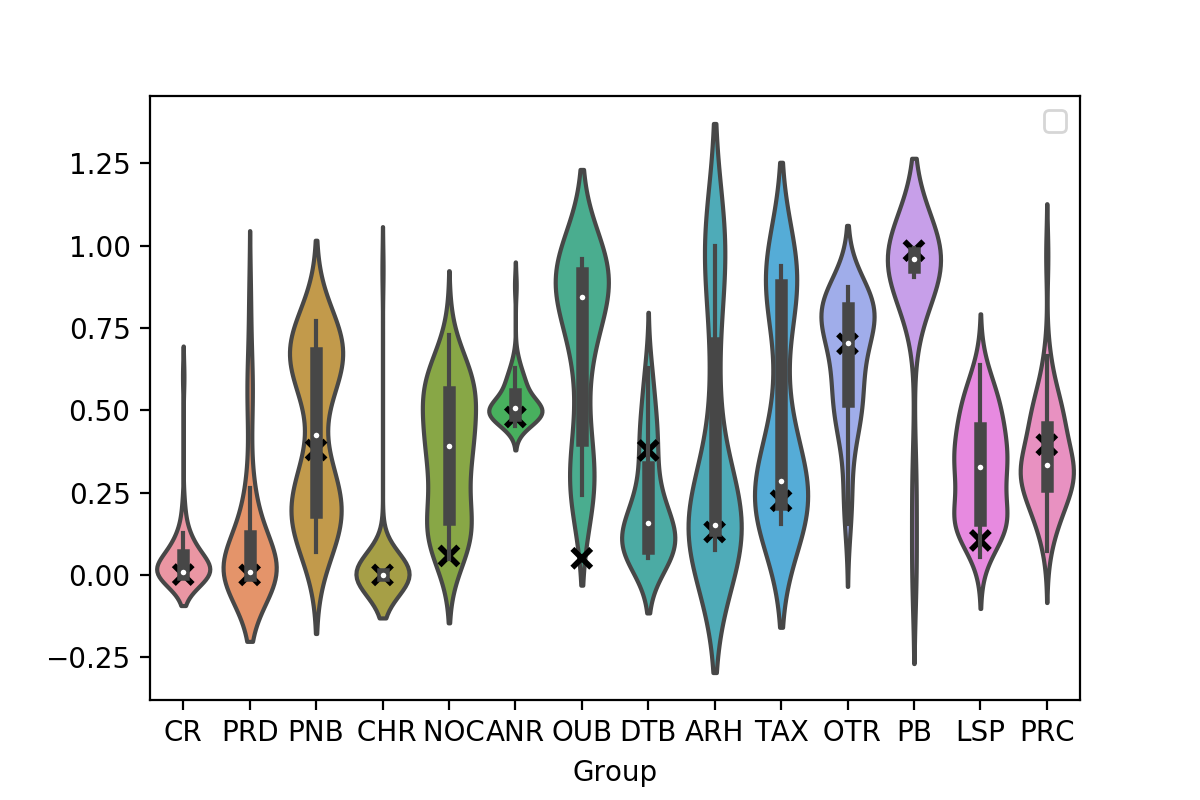}} \\
   \subfigure[]{\centering
   \includegraphics[width=0.4\textwidth]{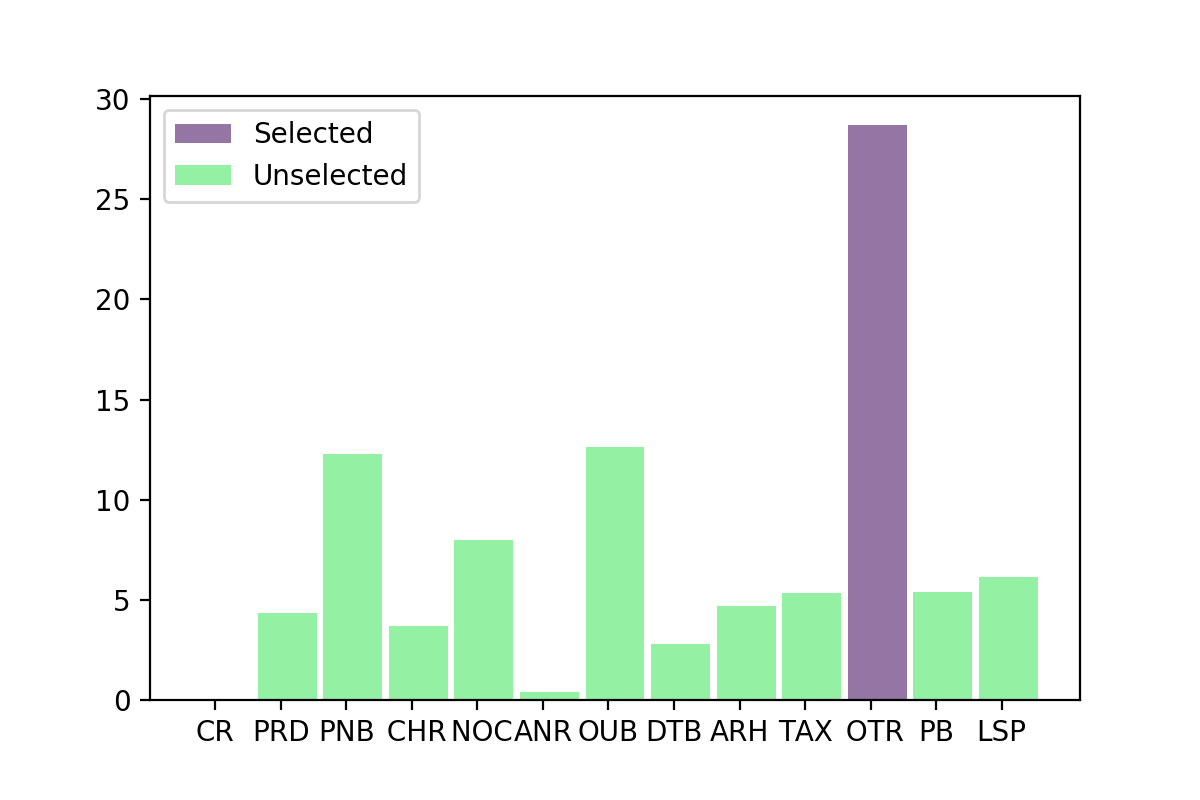}}
    \subfigure[]{\centering
   \includegraphics[width=0.4\textwidth]{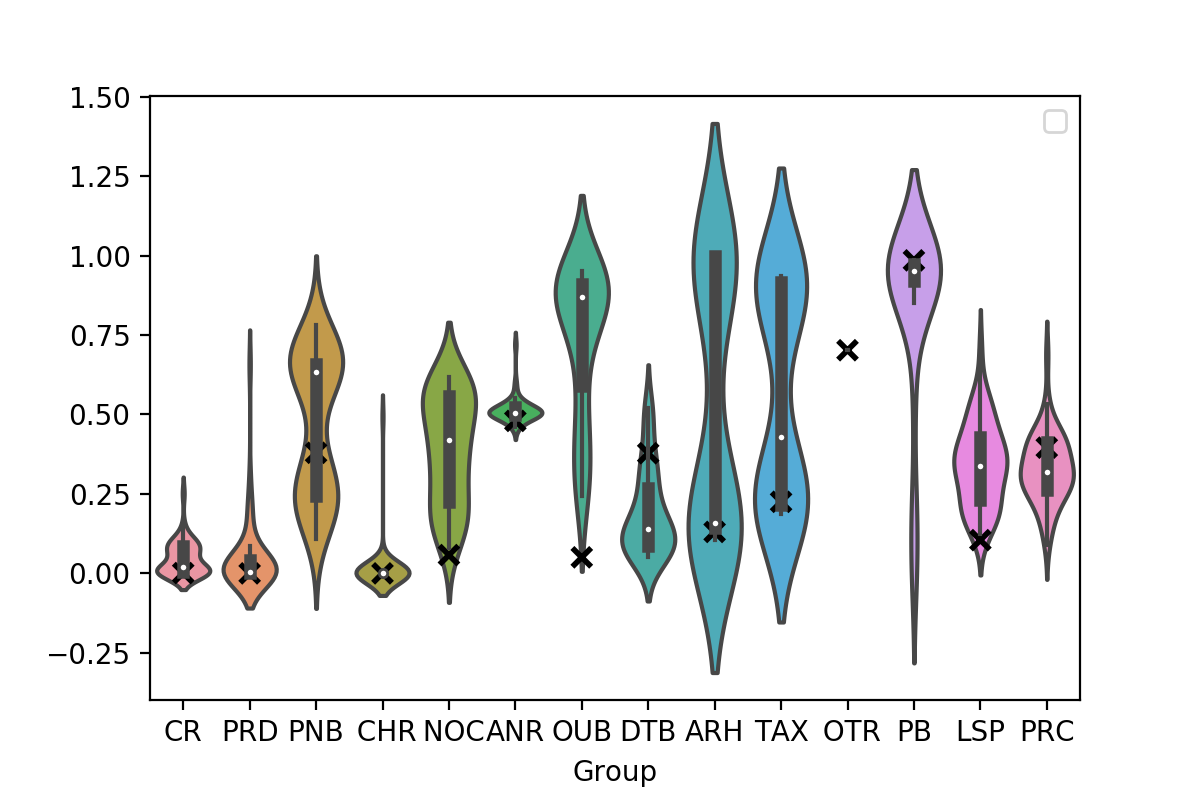}} \\
   \subfigure[]{\centering
   \includegraphics[width=0.4\textwidth]{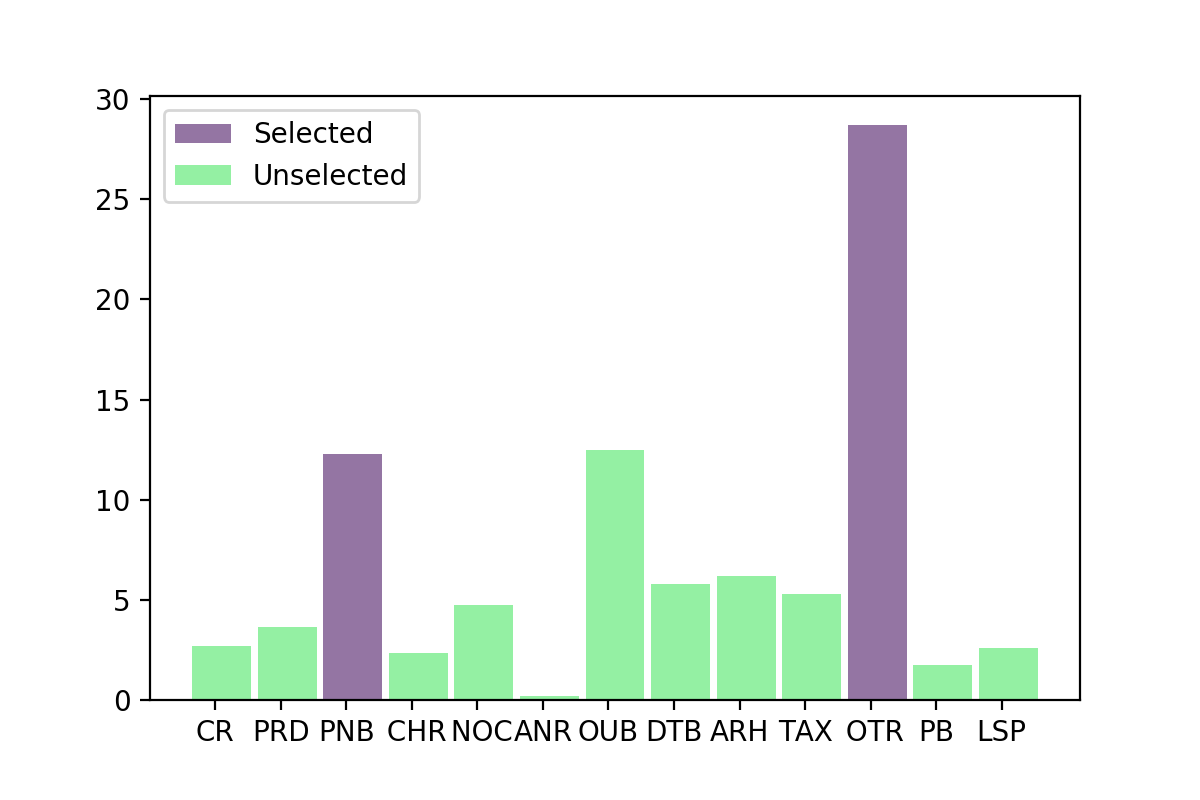}}
    \subfigure[]{\centering
   \includegraphics[width=0.4\textwidth]{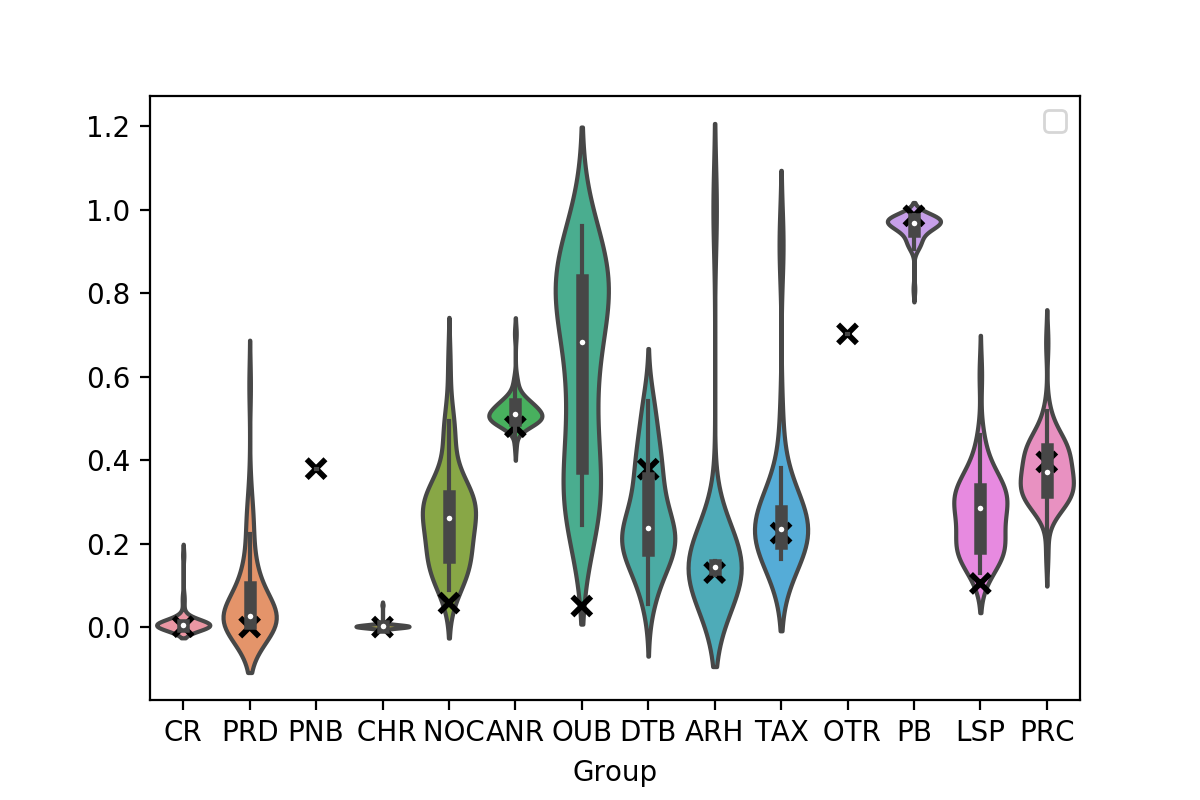}} \\
      \subfigure[]{\centering
   \includegraphics[width=0.4\textwidth]{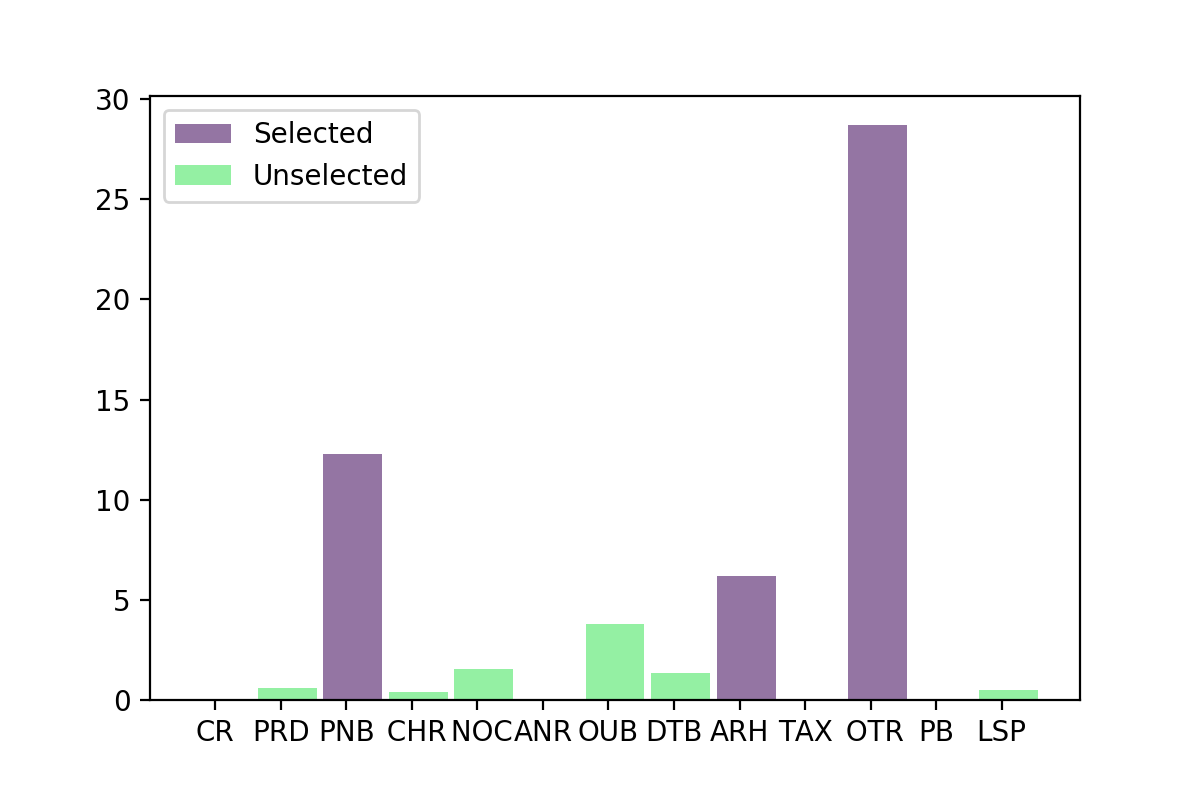}}
    \subfigure[]{\centering
   \includegraphics[width=0.4\textwidth]{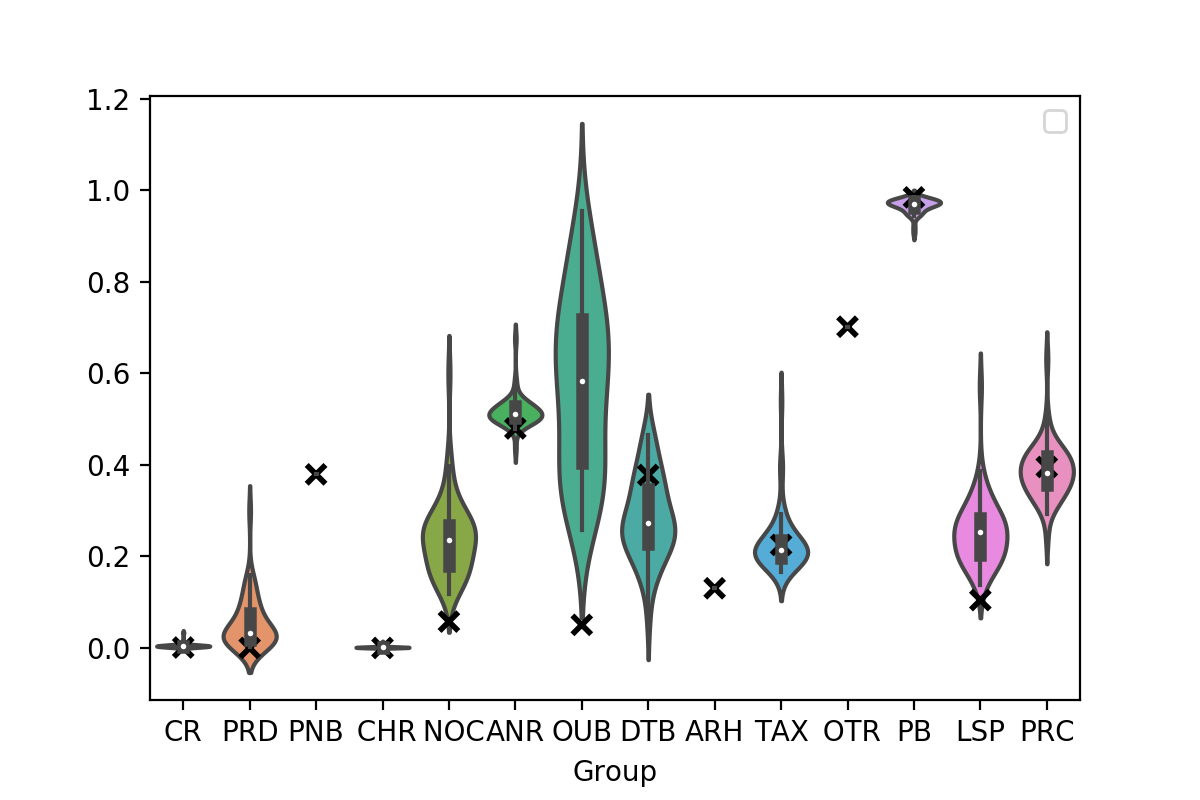}} \\
   \caption{Information reward estimated during the first 4 active variable selection steps on a randomly chosen Boston Housing test data point.  \textbf{Models}: PNP, strategy: single ordering. Each row contains two plots regarding the same time step. \textbf{Bar plots on the left} show the information reward estimation of each variable on the y-axis. All unobserved variables start with green bars, and turns purple once selected by the algorithm.  \textbf{Right}: violin plot of the posterior density estimations of remaining unobserved variables. }. 
   \label{fig:bar_sing}
\end{figure}

The decision process facilitated by the active selection of the variables (for the EDDI framework) is efficiently illustrated in Figure \ref{fig:bar_chai} and Figure \ref{fig:bar_sing} for the Boston Housing dataset and for the PNP and PNP with single best ordering approaches, respectively. 

For completeness, we provide details regarding the abbreviations of the variables used in the Boston dataset and appear both figures.
\begin{itemize}
\item[] CR - per capita crime rate by town
\item[] PRD - proportion of residential land zoned for lots over 25,000 sq.ft.
\item[] PNB - proportion of non-retail business acres per town.
\item[] CHR - Charles River dummy variable (1 if tract bounds river; 0 otherwise)
\item[] NOC - nitric oxides concentration (parts per 10 million)
\item[] ANR - average number of rooms per dwelling
\item[] AOUB - proportion of owner-occupied units built prior to 1940
\item[] DTB - weighted distances to five Boston employment centres
\item[] ARH - index of accessibility to radial highways
\item[] TAX - full-value property-tax rate per \$10,000
\item[] OTR - pupil-teacher ratio by town
\item[] PB - proportion of blacks by town
\item[] LSP - \% lower status of the population
\end{itemize}

\subsection{MIMIC-III}
\label{sec:app_MIMIC}
Here we provide additional results of our approach on the MIMIC-III dataset.

\subsubsection{Preprocessing and model details}
For our active learning experiments on MIMIC III datasets, we chose the variable of interest $\mathbf{x}_\phi$ to be the binary mortality indicator of the dataset. All data (except the binary mortality indicator) are normalized and then scaled between 0 and 1. We transformed the categorical variables into real-valued using the dictionary deduced from \citep{johnson2016mimic} that makes use of the actual medical implications of each possible values. The binary mortality indicator are treated as Bernoulli variables and Bernoulli likelihood function is applied. For each repetition (of the 5 in total), we randomly draw 10\% of the whole data to be our test set. Partial VAE models (ZI, ZI-m, PNP and PNs) share the same size of architecture with 10 dimensional diagonal Gaussian latent variables: the generator (decoder) is a 10-50-100-D neural network with ReLU activations (where D is the data dimensions). The inference nets (encoder) share the same structure of D-100-50-20 that maps the observed data into distributional parameters of the latent space. Additionally, for PN-based parameterizations, we further use a 20 dimensional feature mapping $h$ parameterized by a single layer neural network, and 10 dimensional ID vectors $\mathbf{e}_i$ (please refer to section \ref{sec:pVAE}) for each variable. We choose the symmetric operator $g$ to be the basic summation operator.

Adam optimization  and random missingness is applied as in the previous experiments. We trained our models for 3K iterations. During active learning, we draw 50 samples in order to estimate the expectation under $\mathbf{x}_{\phi},\mathbf{x}_i \sim p(\mathbf{x}_{\phi}, \mathbf{x}_i|\mathbf{x}_o)$ in Equation (\ref{eq:CHAIN}). Loss functions (RMSEs and negative log likelihoods) of the target variable is also estimated using samples of $\mathbf{x}_{\phi} \sim p(\mathbf{x}_{\phi}|\mathbf{x}_o)$ through $p(\mathbf{x}_{\phi}|\mathbf{x}_o) \approx \frac{1}{M} \sum_{m=1}^M p(\mathbf{x}_{\phi}|\mathbf{z}_m)$, where $\mathbf{z}_m \sim q(\mathbf{z}|\mathbf{x}_o)$.

\subsubsection{Additional plots of ZI, PN and ZI-m on MIMIC III}
\begin{figure}[H]
\centering
\subfigure[]{\centering
   \includegraphics[width=0.3\textwidth]{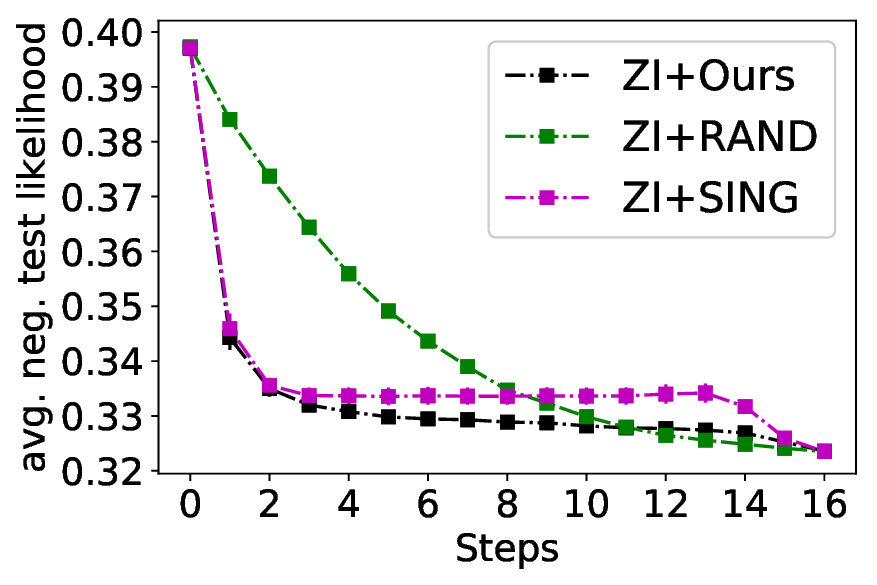}}
    \subfigure[]{\centering
   \includegraphics[width=0.3\textwidth]{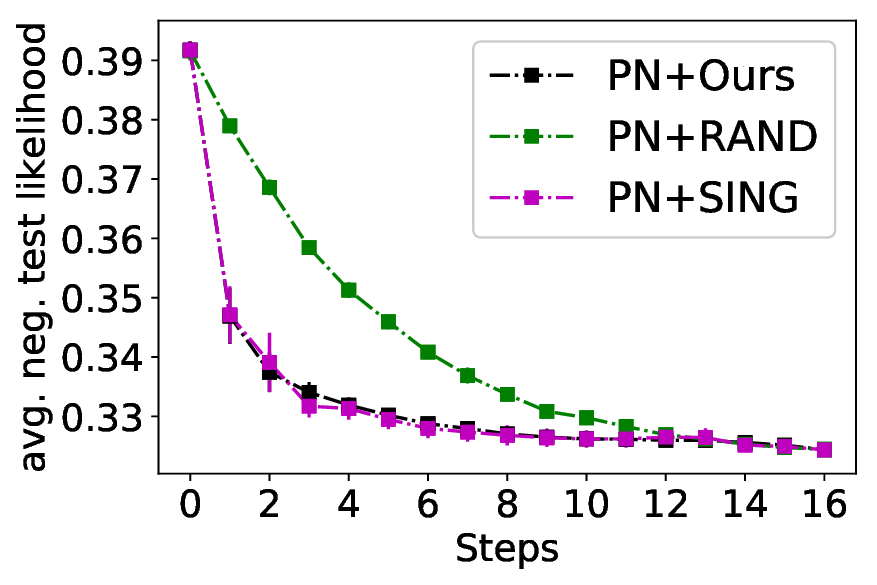}}
   \subfigure[]{
   \includegraphics[width=0.3\textwidth]{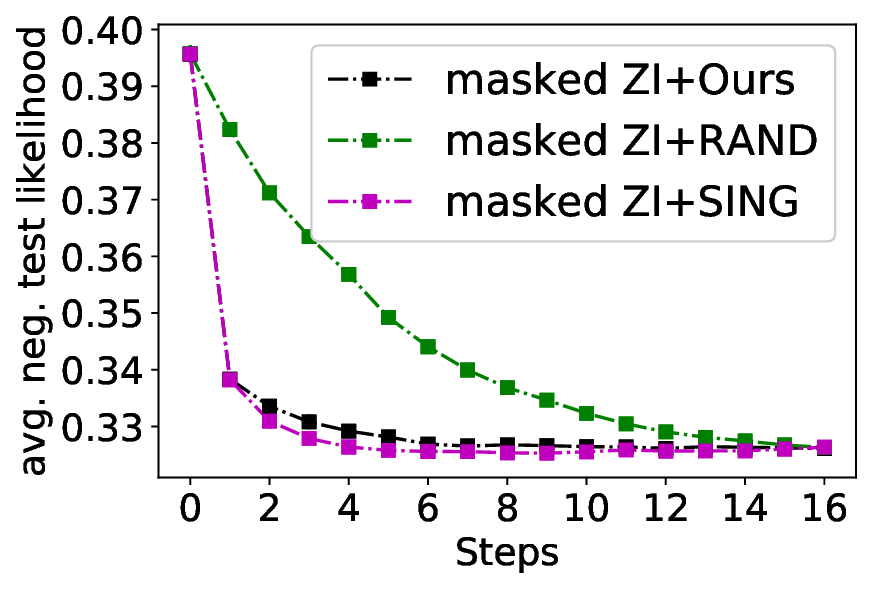}}
   \caption{Information curves of active variable selection on risk assessment task on MIMIC III, produced from: \textbf{(a)} Zero Imputing (ZI), \textbf{(b)} PointNet (PN) and \textbf{(c)} Zero Imputing with mask (ZI-m). \textbf{Green}: random strategy; \textbf{Black}: EDDI; \textbf{Pink}: Single best ordering. This displays negative test Bernoulli likelihood (y axis, the lower the better) during the course of active selection (x-axis)}. 
   \label{fig:mimic_other_three}
\end{figure}

Figure \ref{fig:mimic_other_three} shows the information curves (Bernoulli negative test likelihood-based) of active variable selection on the risk assessment task for MIMIC-III as produced by the three approaches, i.e. ZI, PN and masked ZI.

\subsection{NHANES}
\label{sec:app_NHANES}

 \subsubsection{Preprocessing and model details}
For our active learning experiments on NHANES datasets, we chose the variable of interest $\mathbf{x}_\phi$ to be the lab test result section of the dataset. All data are normalized and scaled between 0 and 1. For categorical variables, these are transformed into real-valued variables using the code that comes with the dataset, which makes use of the actual ordering of variables in questionnaire. Then, for each repetition (of the 5 repetitions in total), we randomly draw 8000 data as training set and 100 data to be test set. All partial VAE models (ZI, ZI-m, PNP and PNs) uses gaussian likelihoods, with an diagonal Gaussian inference model (encoder). Partial VAE models  share the same size of architecture with 20 dimensional diagonal Gaussian latent variables: the generator (decoder) is a 20-50-100-D neural network. The inference nets (encoder) share the same structure of D-100-50-20 that maps the observed data into distributional parameters of the latent space. Additionally, for PN-based parameterizations, we further use a 20 dimensional feature mapping $h$ parameterized by a single layer neural network, and 100 dimensional ID vectors $\mathbf{e}_i$ (please refer to section \ref{sec:pVAE}) for each variable. We choose the symmetric operator $g$ to be the basic summation operator. 

Adam optimization  and random missingness is applied as in the previous experiments. We trained all models 1K iterations. During active learning, 10 samples were drawn to estimate the expectation in Equation (\ref{eq:CHAIN2}). Losses (RMSEs) of the target variable is also estimated using 10 samples.

\section{Additional Theoretical Contributions}
\label{sec:app_theory}

\subsection{Zero imputing as a Point Net}
\label{sec:ZIasPN}

Here we present how the zero imputing (ZI) and PointNet (PN) approaches relate.
\paragraph{Zero imputation with inference net}
In ZI, the natural parameter of $\lambda$ (e.g., Gaussian parameters in variational autoencoders) is approximated using the following neural network:
$$f(\mathbf{x}) := \sum_{l=1}^L w_l^{(1)}\sigma(\mathbf{w}_l^{(0)}\mathbf{x}^T)$$,

where $L$ is the number of hidden units, $\mathbf{x}$ is the input image with $x_i$ be the value of the $i^{th}$ pixel. To deal with partially observed data $\mathbf{x} = \mathbf{x}_o \cup \mathbf{x}_u$, ZI simply sets all $\mathbf{x}_u$ to zero, and use the full inference model $f(\mathbf{x})$ to perform approximate inference.
\paragraph{PointNet parameterization}
The PN approach approximates the natural parameter $\lambda$ by a permutation invariant set function
$$g(h(\mathbf{s}_1),h(\mathbf{s}_2),...,h(\mathbf{s}_O)),$$
where $\mathbf{s}_i = [x_i,\mathbf{e}_i]$, $\mathbf{e}_i$ is the $I$ dimensional embedding/ID/location vector of the $i^{th}$ pixel, $g(\cdot)$ is a symmetric operation such as max-pooling and summation, and $h(\cdot)$ is a nonlinear feature mapping from $\mathbb{R}^{I+1}$ to $\mathbb{R}^K$ (we will always refer $h$ as \emph{feature maps} ). In the current version of the partial-VAE implementation, where Gaussian approximation is used, we set $K=2H$ with $H$ being the dimension of latent variables. We set $g$ to be the element-wise summation operator, i.e. a mapping from $\mathbb{R}^{KO}$ to $\mathbb{R}^K$ defined by:
$$g(h(\mathbf{s}_1),h(\mathbf{s}_2),...,h(\mathbf{s}_O)) = \sum_{i\in O}h(\mathbf{s}_i).$$
This parameterization corresponds to products of multiple Exp-Fam factors $\prod_{i\in O}\exp \{-\langle h(\mathbf{s}_i),\Phi\rangle\}$.
\paragraph{From PN to ZI}
To derive the PN correspondence of the above ZI network we define the following PN functions:
$$h(\mathbf{s}_i):=\mathbf{e}_i*x_i$$
$$g(h(\mathbf{s}_1),h(\mathbf{s}_2),...,h(\mathbf{s}_O)):=\sum_{k=1}^{I}\theta_k\sigma(\sum_{i\in O}h_k(\mathbf{s}_i)),$$

where $h_k(\cdot)$ is the $k^{th}$ output feature of $h(\cdot)$. The above PN parameterization is also permutation invariant; setting $L = I$, $\theta_l = w_l^{(1)}$,$(\mathbf{w}_l^{(0)})_i = (\mathbf{e}_i)_l$ the resulting PN model is equivalent to the ZI neural network.

\paragraph{Generalizing ZI from PN perspective}
In the ZI approach, the missing values are replaced with zeros. However, this ad-hoc approach does not distinguish missing values from actual observed zero values. In practice, being able to distinguish between these two is crucial for improving uncertainty estimation during partial inference. One the other hand, we have found that PN-based partial VAE experiences difficulties in training. To alleviate both issues, we proposed a generalization of the ZI approach that follows a PN perspective. One of the advantages of PN is setting the \emph{feature maps} of the unobserved variables to zero instead of the related weights. As  discussed before, these two approaches are equivalent to each other only if the factors are linear. More generally, we can parameterize the PN by:
$$h^{(1)}(\mathbf{s}_i):=\mathbf{e}_i*x_i$$
$$h^{(2)}(h^{(1)}_i):=NN_1(h^{(1)}_i)$$
$$g(h(\mathbf{s}_1),h(\mathbf{s}_2),...,h(\mathbf{s}_O)):=NN_2(\sigma(\sum_{i\in O}h_k^{(2)}(h^{(1)}_i))),$$
where $NN_1$ is a mapping from $\mathbb{R}^{I}$ to $\mathbb{R}^K$ defined by a neural network, and $NN_2$ is a mapping from $\mathbb{R}^{K}$ to $\mathbb{R}^{2H}$ defined by another neural network. 

\subsection{Approximation Difficulty of the Acquisition Function}
\label{sec:app_difficulty}
Traditional variational approximation approaches provide wrong approximation direction when applied in this case (resulting in an upper bound of the objective $R_{\phi}(i,\mathbf{x}_O)$ which we maximize). Justification issues aside, (black box) variational approximation requires sampling from approximate posterior $q(\mathbf{z}|\mathbf{x}_O)$, which leads to extra uncertainties and computations. For common proposals of approximation:
\begin{itemize}
\item Directly estimate entropy via sampling $\Rightarrow$ problematic for high dimensional target variables
\item Using reversed information reward $\mathbb{E}_{\mathbf{x}_i \sim p(\mathbf{x}_i|\mathbf{x}_o)} [D_{KL}( {\color{red}p(\mathbf{x}_{\phi}|\mathbf{x}_o)}||{\color{blue}p(\mathbf{x}_{\phi}|\mathbf{x}_o,\mathbf{x}_i) }) ]$, and then apply ELBO (KL-divergence)  $\Rightarrow$ This does not make sense mathematically, since this will result in upper bound approximation of the (reversed) information objective, this is in the wrong direction.
\item Ranganath's bound \citep{ranganath2016hierarchical} on estimating entropy$\Rightarrow$ gives upper bound of the objective, wrong direction.
\item All the above methods also needs samples from latent space (therefore second level approximation needed).
\end{itemize}

\subsection{Connection of EDDI information reward with BALD}
\label{sec:app_BALD}
We briefly discuss connection of EDDI information reward with BALD \citep{houlsby2011bayesian} and. MacKay's work \citep{mackay1992information}.
Assuming the model is correct, i.e. $q=p$, we have
\begin{align*}
R(i,\mathbf{x}_o) & =  \mathbb{E}_{\mathbf{x}_i \sim p(\mathbf{x}_i|\mathbf{x}_o)} \left[ D_{KL}( p(\mathbf{z}|\mathbf{x}_i,\mathbf{x}_o)||p( \mathbf{z}|\mathbf{x}_o)) \right] \\
&-  \mathbb{E}_{\mathbf{x}_i \sim p(\mathbf{x}_i|\mathbf{x}_o)}\mathbb{E}_{\mathbf{x}_{\phi} \sim p(\mathbf{x}_\phi|\mathbf{x}_i,\mathbf{x}_o)} \left[ D_{KL}( p(\mathbf{z}|\mathbf{x}_{\phi},\mathbf{x}_i,\mathbf{x}_o)||p( \mathbf{z}|\mathbf{x}_{\phi}, \mathbf{x}_o)) \right]. 
\end{align*}
Note that based on McKay's relationship between entropy and KL-divergence reduction, we have:
\begin{align*}
&\mathbb{E}_{\mathbf{x}_i \sim p(\mathbf{x}_i|\mathbf{x}_o)} \left[ D_{KL}( p(\mathbf{z}|\mathbf{x}_i,\mathbf{x}_o)||p( \mathbf{z}|\mathbf{x}_o)) \right] \\ 
= & \mathbb{E}_{\mathbf{x}_i \sim p(\mathbf{x}_i|\mathbf{x}_o)} \left[ H( p(\mathbf{z}|\mathbf{x}_i,\mathbf{x}_o))-H(p( \mathbf{z}|\mathbf{x}_o))] \right].
\end{align*}

Similarly, we have 
\begin{align*}
& \mathbb{E}_{\mathbf{x}_i \sim p(\mathbf{x}_i|\mathbf{x}_o)}\mathbb{E}_{\mathbf{x}_{\phi} \sim p(\mathbf{x}_\phi|\mathbf{x}_i,\mathbf{x}_o)} \left[ D_{KL}( p(\mathbf{z}|\mathbf{x}_{\phi},\mathbf{x}_i,\mathbf{x}_o)||p( \mathbf{z}|\mathbf{x}_{\phi}, \mathbf{x}_o)) \right] \\
=&  \mathbb{E}_{\mathbf{x}_{\phi} \sim p(\mathbf{x}_\phi|\mathbf{x}_o)} \mathbb{E}_{\mathbf{x}_i \sim p(\mathbf{x}_i|\mathbf{x}_\phi, \mathbf{x}_o)}\left[ D_{KL}( p(\mathbf{z}|\mathbf{x}_{\phi},\mathbf{x}_i,\mathbf{x}_o)||p( \mathbf{z}|\mathbf{x}_{\phi}, \mathbf{x}_o)) \right] \\
=& \mathbb{E}_{\mathbf{x}_{\phi} \sim p(\mathbf{x}_\phi|\mathbf{x}_o)} \mathbb{E}_{\mathbf{x}_i \sim p(\mathbf{x}_i|\mathbf{x}_\phi, \mathbf{x}_o)}\left[ H( p(\mathbf{z}|\mathbf{x}_{\phi},\mathbf{x}_i,\mathbf{x}_o))-H(p( \mathbf{z}|\mathbf{x}_{\phi}, \mathbf{x}_o)) \right] \\
=& \mathbb{E}_{\mathbf{x}_i \sim p(\mathbf{x}_i|\mathbf{x}_o)}\mathbb{E}_{\mathbf{x}_{\phi} \sim p(\mathbf{x}_\phi|\mathbf{x}_i,\mathbf{x}_o)} \left[ H( p(\mathbf{z}|\mathbf{x}_{\phi},\mathbf{x}_i,\mathbf{x}_o)) \right] - \mathbb{E}_{\mathbf{x}_{\phi} \sim p(\mathbf{x}_\phi|\mathbf{x}_o)} \mathbb{E}_{\mathbf{x}_i \sim p(\mathbf{x}_i|\mathbf{x}_\phi, \mathbf{x}_o)}\left[ H(p( \mathbf{z}|\mathbf{x}_{\phi}, \mathbf{x}_o)) \right] \\
=& \mathbb{E}_{\mathbf{x}_i \sim p(\mathbf{x}_i|\mathbf{x}_o)}\mathbb{E}_{\mathbf{x}_{\phi} \sim p(\mathbf{x}_\phi|\mathbf{x}_i,\mathbf{x}_o)} \left[ H( p(\mathbf{z}|\mathbf{x}_{\phi},\mathbf{x}_i,\mathbf{x}_o)) \right] - \mathbb{E}_{\mathbf{x}_{\phi} \sim p(\mathbf{x}_\phi|\mathbf{x}_o)} \left[ H(p( \mathbf{z}|\mathbf{x}_{\phi}, \mathbf{x}_o)) \right], 
\end{align*}
where MacKay's result is applied to $\mathbb{E}_{\mathbf{x}_i \sim p(\mathbf{x}_i|\mathbf{x}_\phi, \mathbf{x}_o)}\left[ D_{KL}( p(\mathbf{z}|\mathbf{x}_{\phi},\mathbf{x}_i,\mathbf{x}_o)||p( \mathbf{z}|\mathbf{x}_{\phi}, \mathbf{x}_o)) \right] $. Putting everything together, we have 

\begin{align*}
R(i,\mathbf{x}_o) & = \mathbb{E}_{\mathbf{x}_i \sim p(\mathbf{x}_i|\mathbf{x}_o)} \left[ H( p(\mathbf{z}|\mathbf{x}_i,\mathbf{x}_o))-H(p( \mathbf{z}|\mathbf{x}_o))] \right] \\
& -\mathbb{E}_{\mathbf{x}_i \sim p(\mathbf{x}_i|\mathbf{x}_o)}\mathbb{E}_{\mathbf{x}_{\phi} \sim p(\mathbf{x}_\phi|\mathbf{x}_i,\mathbf{x}_o)} \left[ H( p(\mathbf{z}|\mathbf{x}_{\phi},\mathbf{x}_i,\mathbf{x}_o)) \right] + \mathbb{E}_{\mathbf{x}_{\phi} \sim p(\mathbf{x}_\phi|\mathbf{x}_o)} \left[ H(p( \mathbf{z}|\mathbf{x}_{\phi}, \mathbf{x}_o)) \right] \\
& = \left\lbrace \mathbb{E}_{\mathbf{x}_i \sim p(\mathbf{x}_i|\mathbf{x}_o)} \left[ H( p(\mathbf{z}|\mathbf{x}_i,\mathbf{x}_o))\right] - \mathbb{E}_{\mathbf{x}_i \sim p(\mathbf{x}_i|\mathbf{x}_o)}\mathbb{E}_{\mathbf{x}_{\phi} \sim p(\mathbf{x}_\phi|\mathbf{x}_i,\mathbf{x}_o)} \left[ H( p(\mathbf{z}|\mathbf{x}_{\phi},\mathbf{x}_i,\mathbf{x}_o)) \right] \right\rbrace \\
&- \left\lbrace \mathbb{E}_{\mathbf{x}_i \sim p(\mathbf{x}_i|\mathbf{x}_o)} \left[ H( p(\mathbf{z}|\mathbf{x}_o))\right] - \mathbb{E}_{\mathbf{x}_{\phi} \sim p(\mathbf{x}_\phi|\mathbf{x}_o)} \left[ H( p(\mathbf{z}|\mathbf{x}_{\phi},\mathbf{x}_o)) \right] \right\rbrace.
\end{align*}

We can show that
\begin{align*}
 &H( p(\mathbf{z}|\mathbf{x}_i,\mathbf{x}_o)) - \mathbb{E}_{\mathbf{x}_{\phi} \sim p(\mathbf{x}_\phi|\mathbf{x}_i,\mathbf{x}_o)} \left[ H( p(\mathbf{z}|\mathbf{x}_{\phi},\mathbf{x}_i,\mathbf{x}_o)) \right] \\
 =& -\int_{\mathbf{z}} p(\mathbf{z}|\mathbf{x}_i,\mathbf{x}_o) \log p(\mathbf{z}|\mathbf{x}_i,\mathbf{x}_o) d\mathbf{z} + \int_{\mathbf{z},\mathbf{x}_{\phi}} p(\mathbf{z}, \mathbf{x}_{\phi}|\mathbf{x}_i,\mathbf{x}_o) \log p(\mathbf{z}|\mathbf{x}_{\phi},\mathbf{x}_i,\mathbf{x}_o) \\
 = & \int_{\mathbf{z},\mathbf{x}_{\phi}}  p(\mathbf{z}, \mathbf{x}_{\phi}|\mathbf{x}_i,\mathbf{x}_o) \log \frac{p(\mathbf{z}, \mathbf{x}_{\phi}|\mathbf{x}_i,\mathbf{x}_o)}{p(\mathbf{z}|\mathbf{x}_i,\mathbf{x}_o) p(\mathbf{x}_{\phi}|\mathbf{x}_i,\mathbf{x}_o)} \\
 = &\mathcal{I}\left[ \mathbf{z}, \mathbf{x}_{\phi}|\mathbf{x}_i,\mathbf{x}_o \right],
\end{align*}
which is exactly the conditional mutual information $\mathcal{I}\left[ \mathbf{z}, \mathbf{x}_{\phi}|\mathbf{x}_i,\mathbf{x}_o \right]$ used in BALD. Therefore, our chain rule representation of reward function leads us to
\begin{align*}
R(i,\mathbf{x}_o) = \mathbb{E}_{\mathbf{x}_i \sim p(\mathbf{x}_i|\mathbf{x}_o)}\mathcal{I}\left[ \mathbf{z}, \mathbf{x}_{\phi}|\mathbf{x}_i,\mathbf{x}_o \right]- \mathbb{E}_{\mathbf{x}_i \sim p(\mathbf{x}_i|\mathbf{x}_o)}\mathcal{I}\left[ \mathbf{z}, \mathbf{x}_{\phi}|\mathbf{x}_o \right].
\end{align*}

\end{document}